\documentclass[runningheads]{llncs}
\usepackage[T1]{fontenc}
\usepackage{graphicx}
\usepackage{algorithm}
\usepackage{algorithmicx}
\usepackage[noend]{algpseudocode}

\usepackage{booktabs}
\usepackage{xspace}
\usepackage{amsmath}
\usepackage{amsfonts}
\usepackage{xcolor}
\usepackage{enumitem} 
\usepackage{subcaption}
\usepackage{array}
\usepackage{multirow} 
\usepackage{placeins} 
\usepackage{hyperref}
\usepackage{xurl}
\usepackage{wrapfig}

\newcommand{\ruc}{CountTRuCoLa\xspace}

\def\pa#1{\textbf{\underline{#1}}}
\def\pb#1{\textbf{#1}}
\def\pc#1{\underline{#1}}

\newcolumntype{P}[1]{>{\raggedleft\arraybackslash}p{#1}}

\DeclareMathOperator*{\argmin}{arg\,min}

\begin{document}
\title{CountTRuCoLa: Rule Learning for Interpretable Temporal Knowledge Graph Forecasting}

%
\titlerunning{Rule Learning for Interpretable Temporal Knowledge Graph Forecasting}
%
\author{Julia Gastinger\orcidID{0000-0003-1914-6723} \and
Christian Meilicke\orcidID{0000-0002-0198-5396} \and 
Heiner Stuckenschmidt\orcidID{0000-0002-0209-3859}} 
\authorrunning{J. Gastinger et al.}
\institute{Data and Web Science Group, University of Mannheim, Germany 
\email{first.last@uni-mannheim.de}}
\maketitle              
\begin{abstract}
We address the task of temporal knowledge graph forecasting with an inherently interpretable method based on symbolic rules. Motivated by recent work proposing a strong baseline based on recurrent facts, our approach learns four simple rule types, including temporal rules with confidence functions that combine both recency and frequency.
Evaluated on nine datasets, our method achieves performance that is competitive with state-of-the-art models and outperforms the majority of them, while each prediction remains directly traceable to the rules and observations that produced it. Moreover, our approach remains functional on very large datasets, where other methods encounter runtime or memory failures.

\end{abstract}

\section{Introduction}
\label{sec:intro}
Temporal knowledge graphs (TKG) extend static knowledge graphs (KG) with temporal information by using timestamps that indicate when a triple is valid~\cite{Han2021xerte}. 
In recent years, the task of TKG forecasting, i.e. predicting future links in TKG, has attracted significant interest, leading to the development of diverse approaches~\cite{Jin2020renet,Li2022TiRGN,chen2025cogntke}. Domain-specific use cases can be found in finance~\cite{li2024findkg}, career trajectory prediction~\cite{Lee2025Caper}, and clinical prediction~\cite{Sun2025Clinical}. 

This paper is motivated by a TKG forecasting baseline recently introduced by~\cite{gastinger2024baselines}, which predicts future links purely based on the recurrence of facts. The authors demonstrate that despite its inability to capture complex dependencies, this baseline offers a competitive or even superior alternative to existing models on three out of five common datasets.
These results challenge the capabilities of current models, which are often built on neural networks or use reinforcement learning within a complex end-to-end architecture.
Motivated by these considerations, we propose a deliberately simple and inherently {interpretable} method for TKG forecasting, which picks up some ideas from the baseline proposed in~\cite{gastinger2024baselines} and extends it to a rule-based TKG forecasting model.

Our model comprises four rule types. Two of them capture simple frequency distributions over entities and relations, while the other two model temporal regularities expressed as symbolic temporal rules. 
Each temporal rule is equipped with a confidence function that assigns a score to a prediction based on two features: the temporal distance to the most recent relevant observation and the frequency of relevant observations. 
The temporal rules contain exactly one body atom, meaning that each prediction is triggered by a single observed fact and its occurrences over time. 

Within this paper we present a comprehensive experimental study comparing our method to eleven 
existing approaches and two baselines on nine datasets. Our results show that the proposed model achieves competitive performance across diverse datasets, outperforming the majority of existing approaches, while being inherently interpretable and computationally efficient. It is the only TKG method that remains functional on all tested datasets without encountering out-of-time or out-of-memory errors, demonstrating that an interpretable model can handle large-scale TKG without sacrificing accuracy.

With interpretability, we mean that each prediction can be directly linked to the rules that fired, their confidence scores, and the particular observations that support them. This interpretability is inherent to the model itself; no post-hoc auxiliary explanation method is required. We note that this reflects the traceability of predictions to model internals: we do not empirically validate human understanding of these explanations through a user study.

\section{Related Work}
\label{sec:relwork}
\paragraph{Rule Learning for Knowledge Graph Completion}
Knowledge graph completion (KGC) usually refers to the non-temporal case. It addresses the task of predicting candidates in an incomplete triple, given a knowledge graph.
It has been shown in~\cite{meilicke2019anytime} and~\cite{rossi2021knowledge} that symbolic methods, which rely on rules, achieve competitive performance compared to KG embedding models, while being inherently interpretable.
Representative symbolic KGC approaches include AMIE~\cite{galarraga2013amie} and its successors~\cite{galarraga2015fast,lajus2020fast}, as well as AnyBURL~\cite{meilicke2019anytime,meilicke2024anytime}. There are also approaches that combine KG embedding models and rules, such as SPaRKLE~\cite{purohit2023sparkle}. 
Static KG are based on the assumption that relations between entities are time-invariant~\cite{han2022TKG}. 
Symbolic rule learning approaches for KGC do not account for temporal information when estimating rule confidence and thus cannot capture temporal dynamics.
Extending these static confidence measures to TKG therefore requires explicitly modeling the temporal dimension; we explain how our approach does so in Section~\ref{sec:approach}.

In contrast to TKG forecasting, TKG completion aims to infer missing facts within the temporal scope of an existing TKG, rather than predicting future facts.
Representative rule-based approaches to TKG completion learn temporal logic rules using differentiable random walks. As an example, TILP~\cite{xiong2023tilp} estimates rule confidences using shared attention parameters together with temporal feature distributions for entity prediction on interval-based TKG. TEILP~\cite{xiong2024teilp} extends this to time prediction by attaching a conditional probability density function to each rule.
Since TKG completion and TKG forecasting address different tasks, methods are not directly comparable. We therefore restrict our experimental evaluation to forecasting methods.

\paragraph{Approaches for TKG forecasting}
Temporal knowledge graph forecasting has been addressed using a variety of approaches. 
Methods based on {deep graph networks} combine message-passing with sequential models to capture both structural and temporal patterns; notable examples include RE-Net~\cite{Jin2020renet}, RE-GCN~\cite{Li2021regcn}, xERTE~\cite{Han2021xerte}, TANGO~\cite{Han2021tango}, RETIA~\cite{Liu2023retia}, CEN~\cite{Li2022cen}, and DiMNet~\cite{dong2025dimnet}.
Graph Hawkes Neural Network (GHNN) \cite{han2020ghnn} utilizes temporal point processes to estimate conditional probabilities of future events in continuous time. 
The intensity for a candidate fact is conditioned on a continuous-time recurrent state that decays as time passes since the last relevant past event, and is updated by each new occurrence, so that both, recency and frequency of a fact occurence, shape the prediction. 
Extending this line of work, HERLN~\cite{du2025hawkes} incorporates community-aware embeddings with a Hawkes-process-based relational graph convolutional network. 
Another line of work leverages {reinforcement learning} agents to discover temporal paths, such as CluSTeR~\cite{Li2020cluster} and TimeTraveler~\cite{Sun21TimeTraveler}. 
Other approaches include history-based prediction in CyGNet~\cite{Zhu2021cygnet}, the use of local and global encoders in TiRGN~\cite{Li2022TiRGN}, candidate-specific temporal context modeling in CRAFT~\cite{zhang2024modeling}, contrastive learning in CENET~\cite{Xu2023CENET}, latent relation mining in L2TKG~\cite{Zhang2023L2TKG}, temporal path encoding in TPAR~\cite{Chen2024TPAR}, and tensor factorization with timestamp encoding in an extended TNTComplEx~\cite{dileo2026tensor}. 
More recently, {LLM-based models} such as zrLLM~\cite{ding2024zrllm}, CoH~\cite{xia2024chain}, and GenTKG~\cite{liao2024gentkg} have been proposed to incorporate semantic information into TKG forecasting.

In addition, {rule-based approaches} aim to learn temporal logic rules. 
StreamLearner~\cite{omran2019streamlearner} was the first approach for learning temporal rules on KG. It addresses a related but distinct problem, namely learning rules continuously from an evolving KG stream. In this context, it first mines static rules and then instantiates separate rules for different time distances between rule body and head. Rule confidences are updated online by aggregating them with a recency-weighted strategy across successive timestamps, restricted to a fixed-size sliding window.
TLogic learns rules via temporal random walks~\cite{Liu2021tlogic}, while TRKG extends this by introducing acyclic and relaxed-time-constraint rules~\cite{Kiran2023TRKG}. TR-Rules also supports acyclic rules and introduces a window confidence measure to address temporal redundancy~\cite{Li2023trrules}. 
TempValid~\cite{huang2024confidence} builds on TLogic rules but explicitly models temporal validity by treating confidence and decay coefficients as learnable parameters within a machine learning framework, applying an exponential decay transformation to temporal features and aggregating them linearly. It relies on a neural model, which, together with advanced negative sampling strategies, improves predictive performance at the expense of interpretability.
Further, several approaches combine rules with embedding-based methods, such as ALRE-IR~\cite{Mei2022alreir}, LogE-Net~\cite{Liu2023logenet}, TECHS~\cite{Lin2023TECHS}, INFER~\cite{li2025iinfer}, and CognTKE~\cite{chen2025cogntke}. These hybrid methods generally sacrifice interpretability due to their reliance on learned embeddings and neural components.

Lastly, two deterministic heuristic baselines have been introduced: EdgeBank~\cite{poursafaei_towards_2022}, developed for single-relational graphs, predicts the recurrence of the same subjects and objects. In contrast, the Recurrency Baseline~\cite{gastinger2024baselines} predicts fact recurrence by incorporating temporal distance and frequency. 

\paragraph{Positioning of our Work}
While StreamLearner was the first temporal rule-learning approach, its primary focus is online rule adaptation for evolving KG streams. 
It instantiates a separate rule per observed time distance, and updates each independently with a weighted moving average. 
Temporal behavior is represented by multiple independent rules, rather than modeling the underlying time-distance and frequency distribution, which is the focus of our approach.

The works closest to ours are TLogic and TempValid. 
However, both approaches have a different language bias that allows them to learn 
rules with multiple atoms in the body.
While this increases expressiveness, it also increases the risk of overfitting and reduces interpretability. Moreover, in an ablation study that we report in the Supplementary Material (Section I)\footnote{Supplementary Material: \url{https://github.com/JuliaGast/counttrucola_submission/blob/main/Supplementary_Material_CountTRuCoLa.pdf}}, we find that restricting TLogic to length-1 rule bodies has little impact on performance on most datasets.
In contrast, our approach intentionally focuses on rules with at most one single body atom.
However, we do not simply reduce complexity.
We design these short rules to be richer and more nuanced.
Specifically, we introduce additional rule types, 
namely rules with constants, and rules reflecting general distributions inherent in the dataset, thus
supporting a broader set of temporal patterns. 
We also propose an improved temporal scoring function with learned parameters, enabling more accurate modeling of temporal dynamics. In this context, our method not only takes into account the time distance to the most recent occurrence of the rule body but, in contrast to TempValid and TLogic, also the frequency of occurrence.

We note that conditioning predictions on recency and frequency is not itself novel; 
in a sense, this is already reflected in neural Hawkes-process approaches such as GHNN.
Our contribution is not this combination itself, but a transparent, symbolic realization of it. 
Instead of relying on neural representations, our confidence function is a per-rule parametric function, 
where each prediction's score can be attributed to specific, human-readable rules and temporal triples.

\section{Background and Notation}
\label{sec:background}
A temporal knowledge graph~$G$ is a set of quadruples $(s,p,o,t)$.  We refer to 
$C(G) = \{ s \mid (s,p,o,t) \in G \} \cup \{ o \mid (s,p,o,t) \in G \}$ as entities (or constants) of $G$, $P(G) = \{ p \mid (s,p,o,t) \in G \}$ as relations (or predicates) of $G$, $T(G) = \{ t  \mid (s,p,o,t) \in G \} \subset \mathbb{N}^+$ as timestamps of $G$. Timestamps may represent hours, years, or any other temporal granularity, depending on dataset and use case.
The semantic meaning of a quadruple $(s,p,o,t)$ is that $s$ is in relation $p$ to $o$ at time $t$. Thus, a quadruple can also be understood as a triple $(s,p,o)$ that is additionally annotated with a timestamp $t$, which tells that the statement made by the triple is true at (or during) $t$. 

Given a TKG $G$, TKG forecasting or extrapolation is the task of predicting quadruples $(s,p,o,t^\star)$ for future timestamps $t^\star > \max(T(G))$ with $p \in P(G)$ and $s,o \in C(G)$. In this work we focus on the task of {entity forecasting}, that is, predicting object or subject entities for queries $(s,p,?,t^\star)$ or $(?,p,o,t^\star)$. This task has become a common choice for measuring the predictive quality of TKG forecasting methods~\cite{gastinger2023eval}. Akin to static knowledge graph completion, TKG forecasting is approached as a ranking task~\cite{han2022TKG}. For a given query, which is usually derived from a quadruple in a test or validation set, a model needs to rank entities in $C(G)$ using a scoring function that assigns plausibility scores. Due to that specific setting, for each query, there is always a correct entity to predict.

To simplify the evaluation protocol, one can extend a TKG $G$ by adding for each $(s,p,o,t) \in G$ the inverse quadruple $(o,p^{-1},s,t)$ where $p^{-1} \not\in P(G)$ denotes a fresh relation used as inverse relation of $p$. This doubles the size of $G$ and the number of relations. It allows to focus on object queries only by converting a subject query as $(?,p,o,t^\star)$ into the equivalent object query $(o,p^{-1},?,t^\star)$. We will see in the following section that it also reduces the number of required rules. Thus, we always extend a given TKG by its inverse quadruples. Note that most evaluation datasets are already available in that extended form.  

We propose a model that learns logical rules to capture the temporal regularities inherent in $G$. In this context we use a prefix notation and understand relation $p$ as a ternary predicate. We use a literal $p(s,o,t)$ as logical representation of a quadruple $(s,p,o,t)$. A logical rule is a disjunction of literals with at most one unnegated literal, for example, $l_0 \vee \neg l_1 \vee \ldots \vee \neg l_n$. Rules are usually written in the equivalent implicative form $l_0 \leftarrow l_1 \wedge \ldots \wedge l_n$. We call $l_0$ the head of the rule and $l_1 \wedge \ldots \wedge l_n$ the body of the rule. Given a rule $r$, we use $\mathfrak{h}(r)$ to refer to its head and $\mathfrak{b}(r)$ to refer to its body.

As explained in Section~\ref{sec:relwork}, rule-based methods have been successfully applied to static KGC~\cite{rossi2021knowledge}. In the non-temporal setting the confidence value of a rule determines (or is aggregated into) the plausibility score which defines the position of a candidate within a ranking for a given query. A rule's confidence value is usually defined as the number of correct predictions divided by the number of all predictions made by that rule~\cite{galarraga2013amie}. In~\cite{meilicke2024anytime} the authors slightly modified this definition  by adding a small positive number $\mathcal{P}$ to the denominator as a smoothing parameter that pushes the confidence score towards $0$ if the number of predictions is small. We include in our approach a similar smoothing factor.

While static KGC works with static confidences, computing confidence values in the temporal case is less straightforward and requires to consider the distances between the timestamps that are associated to the quadruples that made the rule fire and the timestamp of the predicted quadruple. Thus, it is with respect to TKG forecasting not sufficient to learn a single value for each rule. Instead, we have to learn a confidence function that uses temporal distances as input.

\section{Approach}
\label{sec:approach}
\begin{figure*}
    \centering
    \begin{subfigure}[t]{0.49\linewidth}
         \includegraphics[trim={.3cm .3cm 0.2cm .2cm},clip,width=\linewidth]{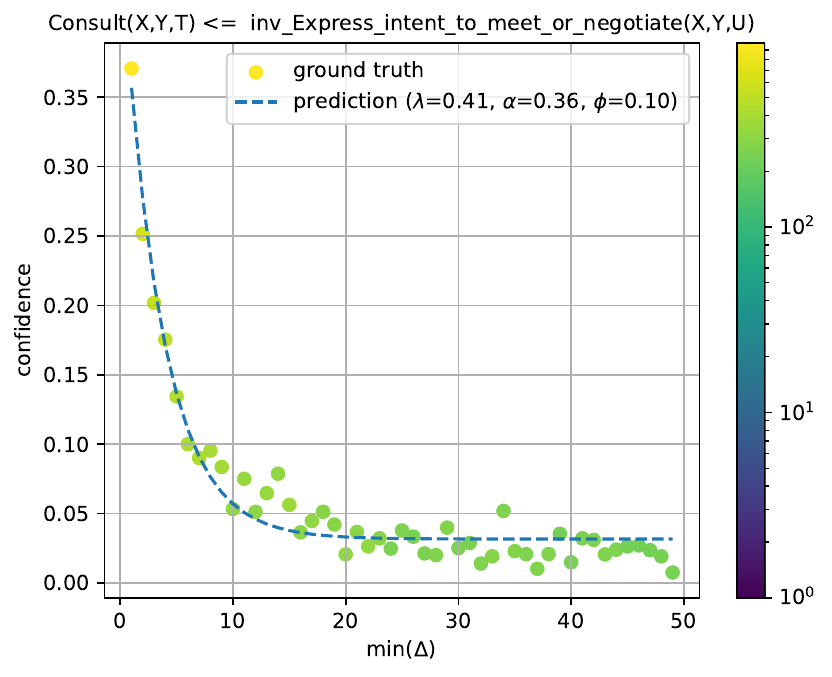}
    \end{subfigure}
    \begin{subfigure}[t]{0.49\linewidth}
        \includegraphics[trim={.3cm .3cm 0.2cm .2cm},clip,width=\linewidth]{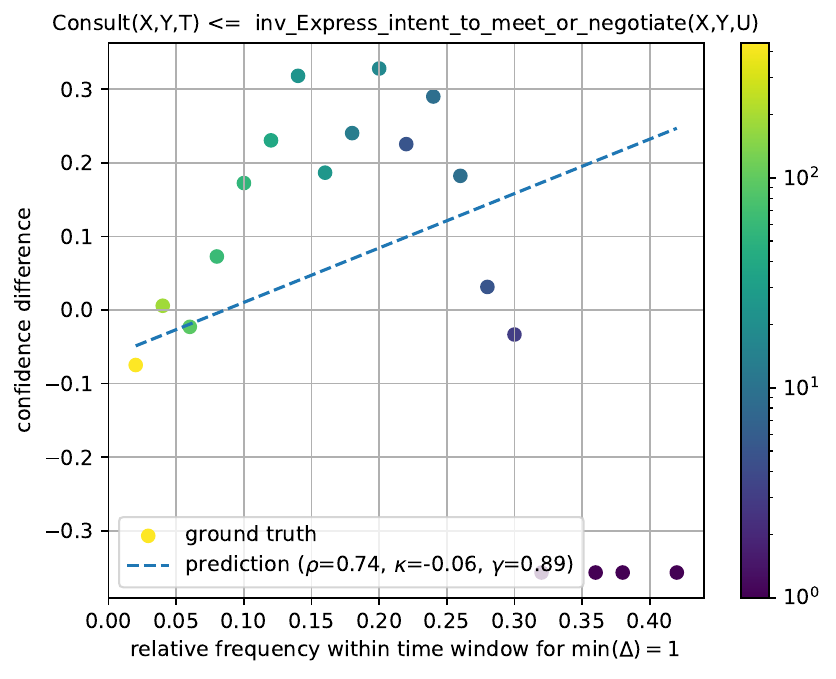}
    \end{subfigure}    
    \vspace{-7pt}
    \caption{Aggregated view of the Examples $E_r$ (points) and predicted curves (blue lines) for one rule $r$. Colors indicate the number of samples in $E_r$ that led to the aggregated value. Left: Learning $f_r$, confidence vs. $\min(\Delta)$. Right: Learning $g_r$, residuals $\tilde{y}-f(\Delta)$ vs.\ frequency $\frac{|\Delta_{\mathcal{W}}|}{\mathcal{W}}$ for $\min(\Delta)=1$.
    }
    \label{fig:curve_combined}
    \vspace{-16pt}
\end{figure*}

In Section~\ref{sub:rules} we explain the rule types supported by our approach and how the rules are used to make a prediction. We explain in Section~\ref{sub:examples} how to collect the positive and negative examples associated to a rule, before we discuss in Section~\ref{sub:tfunction}, how to count these example sets to compute the confidence functions associated to these rules. An example is shown in Figure \ref{fig:curve_combined}. We refer to our approach as \ruc (Temporal Rule Confidence Learning).

\subsection{Rules}
\label{sub:rules} 
Our approach supports four types of rules. Keep in mind that these rules are not used within a complete and general reasoning framework. Instead, they are used for the specific purpose to predict candidates for a query $h(c,?,t^\star)$.
As an example, Figure~\ref{fig:example-explain} shows different rules that predict candidates for the query \emph{(Alexis\_T., Consult, ?, 334)}.
We assume in the following that $t^\star$ refers to a future timestamp and for any timestamp $t  \in T(G)$ we have $t < t^\star$. 

The most important type of rule, called $xy$-rule, is shown in (\ref{rule:xy-rule}). It is called an $xy$-rule because it uses two variables $x$ and $y$ in its head and body.
\begin{align}
\label{rule:xy-rule}  h(x, y, t^\star) \leftarrow b(x,y,t)
\end{align}
We call an $xy$-rule where $h = b$ a recurrent $xy$-rule. 
Recurrent $xy$-rules correspond to the core element in the baseline from~\cite{gastinger2024baselines}.
Rules~(\ref{rule:dated1}) and (\ref{rule:dated2}) are examples for recurrent and non-recurrent $xy$-rules. 
\begin{align}
\label{rule:dated1} \textit{dated}(x, y, t^\star) \leftarrow  &  \textit{dated}(x,y,t)  \\
\label{rule:dated2} \textit{engaged}(x, y, t^\star) \leftarrow  &  \textit{dated}(x,y,t) 
\end{align}
Note that our approach does not support rules where the order of variables is flipped. We mentioned above that we extend each TKG by its inverse quadruples. Thus, a rule as $h(x, y, t^\star) \leftarrow b(y,x,t)$ is equivalent to $h(x, y, t^\star) \leftarrow b^{-1}(x,y,t)$.

Suppose we are concerned with a query $h(c,?,t^\star)$. Relevant $xy$-rules are 
rules that use $h$ in their head, for example, a rule $r = h(x,y,t^\star) \leftarrow b(x,y,t)$. 
Rule $r$ predicts a candidate $d$, i.e., $r$ predicts the quadruple $h(c,d,t^\star)$, if there exists a $t \in T(G)$ with $t^\star\!>\!t$ such that $b(c,d,t) \in G$. We can check this for a given $d$ via a look-up in constant time by using appropriate index structures and we can retrieve all predicted candidates in linear time with respect to the number of candidates. 
The existence of one such timestamp $t$ is sufficient for the rule to fire and produce a prediction. However, to compute the confidence value associated to such a prediction, we must consider all timestamps that satisfy this condition.

The second type of rule that we support, is a rule type that uses constants $d$ and $d'$ in head and body of the rule. We call the following rule a $c$-rule.
\begin{align}
\label{rule:crule} h(x, d, t^\star) \leftarrow  &   b(x, d' ,t) 
\end{align}
The constants that appear in c-rules are usually frequent constants. The following example says that a person born in Amsterdam will (probably) study at University of Amsterdam.   
\begin{align}
\textit{studied}(x, \textit{uva}, t^\star) \leftarrow  &   \textit{born}(x, \textit{amsterdam},t) 
\end{align}
Suppose again that we are concerned with a query $h(c,?,t^\star)$. Similar to the $xy$-rule, a $c$-rule (\ref{rule:crule}) predicts a candidate
$d$, i.e., predicts the quadruple $h(c, d, t^\star)$, if there exists $t \in T (G)$
such that $b(c, d', t) \in G$.

We include two further types of rules that use a static confidence, similar to the  confidence value that is used in static KGC. These rules can be understood as a means to capture very basic frequency distributions within the dataset. We call these rules $z$-rules (\ref{rule:zrule}) and $f$-rules (\ref{rule:frule}). Note that $c$ and $d$ refer to constants, while $x$ refers to a variable.
\begin{align}
\label{rule:zrule} h(x, d, t^\star) \leftarrow  &    \\
\label{rule:frule} h(c, d, t^\star) \leftarrow  &     
\end{align}

Both types of rules have an empty body. This means that these rules always yield a prediction regardless of $G$ whenever we ask a query for which these rules are relevant.  A $z$-rule is relevant for a given query if it uses the relation of that query, an $f$-rule is relevant if is uses the relation of that query and the query subject in its subject position.

The following two rule examples may help to clarify the difference between these rule types. 
\begin{align}
\label{rule:ex-zrule} eats(x, pizza, t^\star) \leftarrow  &     \\
\label{rule:ex-frule} eats(kim, pizza, t^\star) \leftarrow  &   
\end{align}
The first rule predicts pizza based on the eating behavior of all people, whereas the second rule captures the specific eating behavior of Kim. 
For the query $eats(kim,?,t^\star)$ both rules would be relevant and both would predict $pizza$ regardless of $G$. In contrast, for a query $eats(joe,?,t^\star)$ only Rule (\ref{rule:ex-zrule}) would be relevant, and Rule (\ref{rule:ex-frule}) related to Kim's eating behavior would be irrelevant.

While we explain how to compute the confidence functions associated to $xy$ and $c$-rules in Section~\ref{sub:tfunction}, we explain how to compute the static confidence value associated to $z$ and $f$-rules already here. The static confidence value associated to a $z$-rule is the number of all $h$-quadruples where $d$ is on object position, divided by the number of all $h$-quadruples. The static confidence value associated to an $f$-rule is the number of all $h$-quadruples where $c$ is on subject position and $d$ is on object position, divided by the number of all $h$-quadruples where $c$ is on subject position. These formulas are slightly modified by two hyperparameters listed in the Supplementary Material (Section D). One of them is the smoothing constant that we introduced in Section~{\ref{sec:background}}.

We construct each possible $xy$-rule ($|P(G)|^2$ combinations), each $z$-rule \linebreak ($|P(G)| \times |C(G)|$ combinations) and each $f$-rule ($|P(G)| \times |C(G)|^2$ combinations). The number of possible combinations are an upper bound. The actual numbers are much lower, as certain combinations of constants and relations do not appear together. However, the combinatorial space is too large for mining all $c$-rules. For that reason we collect a subset of all $c$-rules that make use of frequent constants only, controlled by the hyperparameter $\mathcal{C}_{\textit{x-count}}$, find more details in the Supplementary Material (Section D). 
The Supplementary Material reports the distribution of rule types in Section F.3 and the number of rules for each dataset in Section F.5, Table 7.

\subsection{Collecting Examples}
\label{sub:examples} 
Learning a confidence function for the $xy$ and $c$-rules requires to collect 
positive and negative examples for each of the constructed rules. 
Here, an example consists of a specific instance of the rule, i.e., a concrete 
prediction, obtained by substituting TKG entities for the rule's variables,
which is marked as correct or incorrect, and a set of feature values.  
The feature values we are interested in are rather specific. For a given grounded prediction, they are the distances from the timestamp of the head atom to the timestamps, for which the corresponding body atom is true, i.e., for which the quadruple corresponding to that atom is an element of $G$. 
With respect to Rule~(\ref{rule:dated1}), it makes, for example, a difference if $x$ dated $y$ two weeks ago, or two years ago. Moreover, it might also be important to know how often $x$ and $y$ have been on dates within a certain time span. Given a rule $r = h(x,y, t^\star) \leftarrow b(x,y,t)$, the confidence function that we introduce makes use of all time stamps $t'$ with $b(c,d,t') \in G$ to estimate the confidence of a prediction $h(c,d,t^\star)$. More precisely, it uses the individual distances between each of these timestamps and $t^\star$. 

Given a logical atom $\mathfrak{a}$ that uses variable $x$, we introduce a substitution function $\theta_{x\!/\!c}$, which maps $\mathfrak{a}$ to an atom where all occurrences of $x$ are replaced by $c$, for example,  $\theta_{x\!/\!c}(p(x,y,t)) = p(c,y,t)$. We use $\theta_{x\!/\!c,y\!/\!d}(\mathfrak{a})$ as an abbreviation for $\theta_{x\!/\!c}(\theta_{y\!/\!d}(\mathfrak{a}))$. 

We further introduce $\Delta_{\theta_{x\!/\!c,y\!/\!d}}^{r,G,t^\star}$ to refer to a set of distances between timestamps: 
\begin{align*}
\Delta_{\theta_{x\!/\!c,y\!/\!d}}^{r,G,t^\star} = \{  t^\star - t' \mid \theta_{x\!/\!c,y\!/\!d,t\!/\!t'}(\mathfrak{b}(r)) \in G, t' \in T(G)  \}
\end{align*}
Subscript $\theta_{x\!/\!c,y\!/\!d}$ expresses that we are concerned with the body groundings and the prediction that results from substituting $x$ by $c$ and $y$ by $d$. 

Compared to the $xy$-rule, the set of distances relevant for the confidence computation of a $c$-rule depends only on the $x$-substitution. The object position in the prediction is already fixed to a constant. Thus, $\Delta_{\theta_{x\!/\!c}}^{r,G,t^\star}$ is the relevant set given a $c$-rule $r$. 
For $c$-rules, we have to distinguish between two application scenarios with two different example sets associated to each $c$-rule. Here we explain only one scenario. The Supplementary Material (Section A) contains a complete description.

So far, we understood that the feature values of an example are distances between the timestamp of a prediction and the timestamps for which the rule body, which yields this prediction, is true according to $G$. However, we have not yet explained how to collect these examples. 

For each $xy$ and $c$-rule $r$ we store examples in a collection $E_r$. Such examples are illustrated in an aggregated form as colored points in Figure~\ref{fig:curve_combined}. The collection procedure is sketched in Algorithm~\ref{alg:examples-collect}. Note that all steps rely exclusively on information from the training set.

\vspace{-10pt}
\begin{algorithm}
  \caption{Collecting Examples}
  \label{alg:examples-collect}
\begin{algorithmic}[1]
     \State  \textbf{Input:} TKG $G$, $xy$-rule $r = h(x,y, t^\star) \leftarrow b(x,y,t)$
     \State  \textbf{Output:} collection of examples $E_r$
     \State  $E_r = [ \ ]$
    \For{$t^\star \in T(G)$}
        \For{$c,d \in C(G)$}
            \If{$\exists z \ h(c,z,t^\star) \in G$}
                \State $\Delta = \Delta_{\theta_{x\!/\!c,y\!/\!d}}^{r,G^{t^ \star},t^\star} $
                \If{$\Delta \neq \emptyset $}
                    \If{$h(c,d,t^\star) \in G$}{ append($\Delta$, 1) to $E_r$}
                    \Else{ append($\Delta$, 0) to $E_r$}
                    \EndIf
                \EndIf
            \EndIf
        \EndFor
    \EndFor
\end{algorithmic}
  
\end{algorithm}
\vspace{-15pt}

The notation $G^{t} = \{ (s,p,o,t') | (s,p,o,t') \in G \wedge t' < t \}$ allows us to refer to the subset of $G$ that contains all quadruples before 
a certain timestamp $t$ (exclusive). The procedure for collecting examples $E_r$ for an $xy$-rule $r = h(x,y,t^\star) \leftarrow b(x,y,t)$ is organized in two nested loops. The outer loop iterates over all timestamps in $t^\star \in T(G)$, treating $G^{t^\star}$ as the given observartions. 
From the perspective of $G^{t^\star}$, timestamp $t^\star$ represents a future point in time for which predictions can be made.
Within the outer loop, we have a nested inner loop that iterates over all pairs $(c,d)$ with $c,d \in C(G)$. For each combination we check if there exists some $z$ with $h(c,z,t^\star) \in G$, ensuring that only entities that might actually appear in a query are included in the example set. This condition enforces that the collected example set adheres to the intended application scenario. 
Then we collect the time distances from the $\theta_{x\!/\!c,y\!/\!d}$-substitution if applied to $r$ for making a prediction that targets the timestamp $t^\star$. We collect only distances within a time window $\mathcal{W}$, where $\mathcal{W}$ is a hyperparameter. If the set is not empty we proceed as follows: If the prediction is correct, i.e., $h(c,d,t^\star) \in G$, we store the set of distances as a positive example. If the prediction is incorrect,  we store it as negative example. Then we continue with the next $(c,d)$ pair.

For a $c$-rule, the procedure is the same with two minor modifications. We can omit to loop over $d$ and, as there is no variable $y$ in the rule body, we have to drop $y\!/\!d$ from the substitution $\theta_{x\!/\!c,y\!/\!d}$. 

\subsection{Temporal Confidence Functions}
\label{sub:tfunction} 
Each $xy$-rule (and each $c$-rule) $r$ is associated with a parameterized temporal confidence function. This function is designed to map a set of distances, which represents the feature description of a prediction made by $r$, to a confidence value. We already introduced this set of distances for a prediction $h(c,d,t^\star)$ as $\Delta_{\theta_{x\!/\!c,y\!/\!d}}^{r,G,t^\star}$ ($\Delta_{\theta_{x\!/\!c}}^{r,G,t^\star}$ for a $c$-rule) and we denote it in the following as $\Delta$ for simplicity. Our confidence function for a rule $r$ is defined by:
\begin{equation}
\textit{conf}_r(\Delta) = f_{r}(\Delta) + g_{r}(\Delta)     
\label{eq:conf_delta}
\end{equation}

The first summand, $f_{r}$, describes the effect of the most recent occurrence, i.e., the smallest time distance $\min(\Delta)$, of the body grounding:
\begin{equation}
    f_{r}(\Delta) =\alpha_r \cdot \frac{1}{1+\phi_r} \cdot ( 2^{-\lambda_r \cdot (\min(\Delta) -1)} + \phi_r)
    \label{eq:score_single}
\end{equation}
where $\alpha_r, \lambda_r,\phi_r \in \mathbb{R}_0^+$ are learnable parameters. 

The second summand, $g_{r}$, captures the effect of the frequency, i.e., occurrence number $|\Delta|$, of the body grounding:
\begin{equation}
g_{r}(\Delta)  =    \min(\max(\rho_r\dfrac{|\Delta_{\mathcal{W}}|}{\mathcal{W}} + \dfrac{\kappa_r}{\min(\Delta)}, -\gamma_r), \gamma_r)
 \label{eq:score_multi}
\end{equation}
where $\Delta_{\mathcal{W}}$ denotes the set of body groundings whose time difference is at most $\mathcal{W}$, i.e., $\Delta_{\mathcal{W}} =  \{  \delta \in \Delta: \delta \leq \mathcal{W}\}$, with $\mathcal{W}$ being a hyperparameter. Further, $\rho_r, \kappa_r, \gamma_r \in \mathbb{R}$ are learnable parameters. 
Note that the additive offset in $g_r$ depends inversely on $\min(\Delta)$.
Intuitively, when $\min(\Delta)$ is low, a high proportion of the window $\mathcal{W}$ can contain body groundings. In contrast, when $\min(\Delta)$ is high, the maximum possible number of additional groundings in $\Delta_{\mathcal{W}}$ is reduced, since the upper limit for $|\Delta_{\mathcal{W}}|$ becomes $\mathcal{W} - min(\Delta) +1$. 

Each confidence function parameter has a specific and interpretable role:
\setlength{\itemindent}{0pt}
\setlength{\leftmargini}{0.8em} 
\begin{itemize}
\setlength{\itemsep}{0em}  
\item $\alpha_r$: scales the $f_r$ score and determines its value when $\min(\Delta) = 1$.
\item $\lambda_r$: specifies the decay rate, determining how quickly $f_r$ decreases as $\min(\Delta)$ increases. When $\lambda_r = 0$, the time distance between body and head has no effect, while higher values result in faster decay.
\item $\phi_r$: determines the asymptotic lower bound of $f_r$ as $\min(\Delta) \to \infty$.
\item $\rho_r$: defines the slope of the frequency term $g_{r}$. When $\rho_r = 0$, frequency has no impact; for larger $\rho_r$, the score increases proportionally with observed frequency.
\item $\kappa_r$: acts as an additive offset modulated by $\min(\Delta)$.
\item $\gamma_r$: bounds the frequency score via clipping, ensuring the impact of $g_r$ remains within $[-\gamma_r, \gamma_r]$.
\end{itemize}
Finally, $\textit{conf}_r(\Delta)$ is bounded via clipping to remain within $[0,1]$.

Using an exponential function to model temporal decay was inspired by TLogic~\cite{Liu2021tlogic}.
While TLogic applies a fixed decay factor to all rules, we learn rule-specific parameters to represent diverse temporal behaviors. In addition, we introduce $\phi_r$ to control the asymptotic lower bound of $f_r$

Recall that $E_r$ for a rule $r$ consists of $N$ examples $(\Delta_i, y_i)$, where $i = 1, \ldots, N$. Here, $y_i$ is the truth value indicating whether a body and head grounding appeared together ($true=1$) or not ($false=0$). We also call $y_i$ the observed confidence value. 
Our approach learns a confidence function for each rule by optimizing the parameters \((\alpha_r, \lambda_r, \phi_r, \rho_r, \kappa_r, \gamma_r)\) aiming to minimize the sum of squared errors between the predicted confidence \(\textit{conf}_r(\Delta)\)~(Equation~\ref{eq:conf_delta}) and the observed confidence 
values from the examples $E_r$.
For simplicity, we omit the subscript~$_r$ in the following.

Following the approach of~\cite{meilicke2024anytime} for confidence computation in static KGC, we adjust the observed confidence values by adding a constant $\mathcal{P}$ to the denominator. 
Here, $\mathcal{P}$ is a hyperparameter. 
Unlike the static setting, we apply this adjustment separately for each value of $\min(\Delta)$. 
We denote this resulting transformation by $s$, yielding modified observed confidence values $\tilde{y_i} = s(\Delta_i, y_i)$. 
For each group of examples in $E_r$ that share the same minimal time distance $\min(\Delta_i)$, we compute a scaling factor $k(\min(\Delta_i))$ rescale the observed confidence value $y_i$ as: 
\begin{equation}
\tilde{y}_i = s(\Delta_i, y_i) = k(\min(\Delta_i)) \cdot y_i.
\end{equation}
Algorithm~\ref{alg:scale} outlines the computation of $k$.

\vspace{-10pt}
\begin{algorithm}
  \caption{Computing the scaling factor $k$}
  \footnotesize
  \label{alg:scale}
  
\begin{algorithmic}[1]
    \State \textbf{Input:} Examples $E_r$ for rule $r$, Parameter $\mathcal{P}$
    \State \textbf{Output:} Mapping $k$ from minimal distance $d$ to scaling factor
    \State Initialize empty set \(X = \emptyset\) and empty maps \(a, k\)
    \For{each $(\Delta_i, y_i)$ in $E_r$}
        \State $d = \min(\Delta_i)$ \hfill // minimal distance 
        \If{$d \notin X$}
            \State Add $d$ to $X$, $a(d) = 0$
        \EndIf
        \State $a(d) = a(d) + 1$  \hfill // count all examples for $d$
    \EndFor
    \For{each $d$ in $X$}
        \State $k(d) = \frac{a(d)}{a(d) + \mathcal{P}}$ \hfill // compute scaling factor for $d$
    \EndFor
    \Return $k$
\end{algorithmic}
\end{algorithm}
\vspace{-10pt}

This approach ensures that the observed confidence values are adjusted according to the number of examples for each smallest time distance, controlled by the hyperparameter $\mathcal{P}$.

To learn the confidence function, we minimize the sum of squared errors (SSE) between the predictions $f+g$ and the transformed observed confidence values:
\begin{equation}
\argmin_{\alpha, \lambda, \phi, \rho, \kappa, \gamma} \sum_{(\Delta_i, y_i) \in E} \big(f(\Delta_i) + g(\Delta_i) - \tilde{y}_i\big)^2.
\label{eq:mse_multi}
\end{equation}
Supplementary Material (Section H) shows learned recency function examples.

\subsection{Aggregation}
\label{sub:aggregation}
A candidate $c$ is typically predicted by several rules $r_0, ..., r_n$ with different confidence scores $s_0, ..., s_n$. We show an example in Figure~\ref{fig:example-explain}. Combining these scores is known as rule aggregation problem~\cite{betz2024rule}. A common strategy is the \textit{noisy-or} strategy, which computes $1 - \prod_{i=0}^{n} (1 - s_i)$, as used also by TLogic~\cite{Liu2021tlogic} and StreamLearner~\cite{omran2019streamlearner}. Because noisy-or assumes independence among rules, we adopt the version proposed by~\cite{betz2024rule} that applies noisy-or to the top $\mathcal{H}$ confidences (e.g., $\mathcal{H} = 5$). We treat $\mathcal{H}$ as a hyperparameter.
Additionally, we introduce a decay factor $\mathcal{D} \in [0,1]$ as hyperparameter. {Let $s_0, ..., s_n$ be sorted in descending order}; we apply a modified noisy-or where each $s_i$ is replaced by $s_i \cdot \mathcal{D}^i$, giving less weight to lower-ranked scores.
The Supplementary Material (Section~D) summarizes all hyperparameters, and Section~E and~F in the Supplementary Material contain studies on hyperparameter sensitivity and effects of rule aggregation.

\section{Experiments}
\label{sec:experiments}
\subsection{Experimental Setup}
\label{sec:experiment-setup}

For our experiments\footnote{\url{https://github.com/JuliaGast/counttrucola_submission} contains our code.}, we use the {datasets} from TGB 2.0 \texttt{smallpedia}, \texttt{polecat}, \texttt{icews}, \texttt{wikidata}~\cite{gastinger2024tgb}, and the datasets \texttt{ICEWS14/18}, \texttt{GDELT}, \texttt{YAGO}, \texttt{WIKI}, in the versions as used by \cite{Li2021regcn} and \cite{gastinger2023eval}. More information on the datasets and dataset statistics are in the Supplementary Material (Section~B). 

We follow the {evaluation protocol} from~\cite{gastinger2023eval}, and use the TGB 2.0 evaluation framework~\cite{gastinger2024tgb}. We report time-aware filtered mean reciprocal rank (MRR) and Hits@10\footnote{More results in the Supplementary Material: Hits@1 and Hits@3 in Section~F.5, variance across repetitions in Section~F.6, and a significance test in Section~F.7.}. 
The time-aware filter setting excludes from the candidate set, for a given query, temporally conflicting candidates that are known to be true at the query timestamp~\cite{gastinger2023eval}.
All experiments are conducted for single-step prediction.

We compare our model to 11 of the 31 methods described in Section~\ref{sec:relwork}, as well as the baselines EdgeBank~\cite{poursafaei_towards_2022} and Recurrency Baseline (Rec.B)~\cite{gastinger2024baselines}. 
Unless otherwise stated, we report results for these 13 methods based on the evaluations in~\cite{gastinger2024tgb} (TGB 2.0 datasets) and~\cite{gastinger2023eval} (all other datasets). 
For TiRGN and CognTKE, we use the reported results and performed a sanity check of the released code.
Since CognTKE achieves the best performance among related methods, we additionally ran the authors’ code on datasets not included in their study (\texttt{GDELT}, \texttt{YAGO}, and TGB~2.0 datasets). We also verified the reproducibility of the reported scores by comparing our and the originally reported scores on \texttt{ICEWS14}, which were the same. Since for TempValid and DiMNet we could not reproduce the scores (Supplementary Material, Section~C.2), we rerun TempValid and DiMNet on all datasets.
We excluded 20 prior methods from comparison for reasons including unavailable code, incompatible evaluation 
protocols, or missing reproducibility information; details are in the Supplementary Material (Section~C.1).

\subsubsection{Hardware}
We tested \ruc on an AMD EPYC 9474F 48-core processor running AlmaLinux 9.5. Parallelization was restricted to a maximum of 20 parallel threads, and RAM was limited to 500 GB. Since \ruc is CPU-based and does not require GPU acceleration, no GPUs were used in these experiments.
CognTKE, TempValid, and DiMNet require GPU acceleration; therefore, these experiments were conducted on an NVIDIA RTX A6000 GPU with 48~GB VRAM, using up to 16 CPU cores and up to 500~GB of RAM.

\subsection{Results}
\label{sec:results}
Table~\ref{tab:results_extended} reports test scores on all nine datasets. \ruc achieves the highest MRR on four datasets and the second or third highest on the remaining. While Hits@10 performance is slightly lower, this is likely because hyperparameter tuning was performed based on MRR.
\ruc outperforms the Recurrency Baseline on seven out of nine datasets with substantial improvements on \texttt{ICEWS14/18}, \texttt{smallpedia}, and \texttt{polecat}, suggesting that rules beyond recurrency, combined with learned confidence functions, improve performance.

Among nine methods evaluated on the large-scale TGB 2.0 datasets, \ruc is the only approach, apart from the simple baseline heuristic EdgeBank, that remains functional on all datasets without encountering out-of-time (OT) or out-of-memory (OM) errors. A runtime study (Supplementary Material Section~F.4) shows that \ruc attains consistently low runtimes relative to prior methods under their respective hardware configurations. Overall, these results indicate that our approach can reliably handle graphs of all tested sizes while delivering predictive accuracy competitive with, and in many cases surpassing, more complex neural architectures.

\setlength{\tabcolsep}{1mm}
\begin{table*}
\footnotesize
\centering
\caption{Experiment results. OM means Out of Memory (40~GB GPU or 500~GB RAM), OT means Out of Time (7 days), - means no results were reported. Best results are bold \& underlined, second-best bold, third-best underlined. }
\resizebox{\textwidth}{!}{
\begin{tabular}{l|rr|rr|rr|rr|rr|rr|rr|rr |rr}
\toprule

& \multicolumn{2}{c|}{\texttt{GDELT}} & \multicolumn{2}{c|}{\texttt{YAGO}} & \multicolumn{2}{c|}{\texttt{WIKI}} & \multicolumn{2}{c|}{\texttt{ICEWS14}} & \multicolumn{2}{c|}{\texttt{ICEWS18}} & \multicolumn{2}{c|}{\texttt{smallp.}} & \multicolumn{2}{c|}{\texttt{polecat}} & \multicolumn{2}{c|}{\texttt{icews}} & \multicolumn{2}{c}{\texttt{wikidata}} \\
\midrule
                          & MRR  & H10 & MRR  & H10 & MRR  & H10 & MRR & H10 & MRR & H10  &  MRR  & H10     &    MRR  & H10       &   MRR  & H10     &    MRR  & H10      \\
\midrule
TRKG                             & 21.5      & 37.3       & 71.5      & 79.2      & 73.4      & 76.2      & 27.3      & 50.8      & 16.7       & 35.4      &    -      &    -      &    -      &    -      &    -      &    -      &    -      &    -      \\
xERTE                            & 18.9      & 32.0       & 87.3      & 91.2      & 74.5      & 80.1      & 40.9      & 57.1      & 29.2       & 46.3      &    -      &    -      &    -      &    -      &    -      &    -      &    -      &    -      \\
TANGO                            & 19.2      & 32.8       & 62.4      & 67.8      & 50.1      & 52.8      & 36.8      & 55.1      & 28.4       & 46.3      &    -      &    -      &    -      &    -      &    -      &    -      &    -      &    -      \\
Timetraveler                     & 20.2      & 31.2       & 87.7      & 91.2      & 78.7      & 83.1      & 40.8      & 57.6      & 29.1       & 43.9      &    -      &    -      &    -      &    -      &    -      &    -      &    -      &    -      \\
TiRGN                            & \pc{21.7} & \pc{37.6}  & 88.0      & 92.9      & 81.7      & \pb{87.1} & {44.0} & \pc{63.8} & \pb{33.7}  & \pb{54.2} &    -      &    -      &    -      &    -      &    -      &    -      &    -      &    -      \\
TempValid & OM & OM & 46.7 & 50.4 & OT & OT & \pb{45.5} & \pa{64.5} & OT & OT&OM & OM & error & error& error& error& error & error \\
DiMNet & 21.1 & 36.5 & 72.4 & 82.7 & 71.2 & 79.7 & 41.4 & 60.8 & 33.9 & 55.4 & 54.3 & 64.7 & OM & OM & OM & OM & OM & OM\\
CognTKE                          & OM       & OM        & \pc{90.6} & \pa{93.2} & \pa{83.2} & \pa{87.3} & \pa{46.1} & \pa{64.5} & \pa{35.2}  & \pa{54.7} & 53.4      & 70.1      & \pa{28.7} & \pa{45.5} & OM       & OM       & OM       & OM       \\
TLogic                           & 19.8      & 35.6       & 76.5      & 79.2      & \pc{82.3} & 87.0      & 42.5      & 60.3      & 29.6       & 48.1      & 59.5      & \pc{70.7} & \pc{22.8} & \pc{37.8} & 18.6      & 30.1      & OT       & OT       \\
RE-GCN                           & 19.8      & 33.9       & 82.2      & 88.5      & 78.7      & 84.7      & 42.1      & {62.7} & 32.6       & \pc{52.6} & 59.4      & 68.7      & 17.5      & 29.2      & 18.2      & \pb{33.1} & OM       & OM       \\
CEN                              & 20.4      & 35.0       & 82.7      & 89.4      & 79.3      & 84.9      & 41.8      & 60.9      & 31.5       & 50.7      & \pb{61.2} & 70.5      & 18.4      & 32.3      & \pc{18.7} & \pa{33.4} & OM       & OM       \\
$\text{EdgeBank}_{\text{tw}}$    & 1.9       & 3.5        & 61.7      & 61.7      & 58.5      & 84.4      & 13.5      & 34.2      & 7.2        & 17.9      & 35.3      & 56.6      & 5.6       & 11.9      & 2.0       & 5.8       & \pb{53.5} & \pb{59.6} \\
Rec.B ($\psi_\Delta\xi$)         & \pa{24.5} & \pb{39.8}  & \pa{90.9} & \pc{93.0} & 81.4      & \pb{87.1} & 37.4      & 51.5      & 28.7       & 43.6      & \pc{60.5} & \pb{71.6} & 19.8      & 31.7      & \pb{21.1} & \pc{32.4} & OT       & OT       \\
\midrule
\ruc                             & \pb{23.8} & \pa{40.3}  & \pa{90.9} & \pa{93.2} & \pb{82.7} & 86.6      & \pc{45.0} & 62.0      & \pc{32.8}  & 51.0      & \pa{64.4} & \pa{71.7} & \pb{25.6} & \pb{40.8} & \pa{21.4} & 32.1      & \pa{60.9} & \pa{62.8} \\
\bottomrule
\end{tabular}
}
\label{tab:results_extended}
\vspace{-10pt}
\end{table*}

\paragraph{Comparison to \texttt{CognTKE}} 
CognTKE exceeds \ruc on four datasets. One likely reason are CognTKE’s local multi-hop reasoning and its temporal relation attention layers, which model reasoning paths of up to four hops and allow to adaptively combine multiple paths and features~\cite{chen2025cogntke}.
These capabilities, however, come at the cost of higher total runtime and reduced transparency. 
As shown in the Supplementary Material (Section~F.4), \ruc is over five times faster, even though it was run solely on CPU while CognTKE was executed on GPU. In addition, running CognTKE resulted in memory errors on the three largest datasets. 
Moreover, although CognTKE highlights relational subgraphs and attention-weighted paths, key steps remain black-box operations due to neural components, and lack the full transparency of our rule-based approach.

\paragraph{A note on \texttt{GDELT}}
On \texttt{GDELT}, \ruc underperforms the Recurrency Baseline despite outperforming all other models.
\texttt{GDELT}, based on a platform that sources global news events in 15 minute intervals, 
features two temporal patterns: (i) short-term repetition driven by media coverage, and (ii) long-term dependencies, such as a country criticizing another in response to a military action.
The difficulty of disentangling these temporal patterns likely explains the relative performance gap.

\paragraph{Ablation Studies}
\label{sec:ablation}
\begin{wraptable}{r}{0.5\textwidth} \vspace{-14pt}
\vskip -0.15in
\caption{Ablation studies showing test scores in different settings.}
\footnotesize
\setlength{\tabcolsep}{1mm}
\centering
\resizebox{.5\textwidth}{!}{
    \begin{tabular}{p{0.1cm}p{0.3cm}l|rr|rr} 
    \toprule
    &&& \multicolumn{2}{c}{\texttt{WIKI}} & \multicolumn{2}{|c}{\texttt{ICEWS14}} \\
    \midrule
     &&                     & MMR  & H10 & MRR  & H10  \\
                   \midrule 
    \multirow{9}{*}{\rotatebox[origin=c]{90}{(a) Rule Types}} &
    & rec-rules     & 82.6          & 86.3                       & 35.6          & 47.3                 \\
    &&$xy$-rules (including rec.) & 82.5          & 86.3               & 42.8          & 60.4                    \\
    &&$c$-rules        & 29.6          & 30.0                    & 25.2          & 34.6                 \\
    &&$z$-rules         & 14.1          & 24.9                & 14.1          & 27.9               \\
    &&$f$-rules         & 67.5          & 82.8                  & 34.0          & 46.5                   \\
    &&all - $z$-rules          & 82.6 & 86.4 & 44.7 & 61.6 \\
    &&all - $f$-rules           & 82.6 & 86.5 & 43.5 & 61.3 \\
    &&all - $c$-rules          & 82.6 & 86.5 & 44.8 & 61.7 \\
    &&all - ($xy$-rules (including rec.))         & 68.5 & 83.2 & 39.0 & 53.1 \\
    && all            & 82.6          & 86.6                & 45.0          & 62.0         \\
    \midrule 
    \multirow{4}{*}{\rotatebox[origin=c]{90}{(b) Conf.}} & \multirow{4}{*}{\rotatebox[origin=c]{90}{Functions}} &static     & 68.0        & 82.7      &   39.9         & 58.5         \\
    &&$g_r$         & 68.6        & 83.2    & 43.4 & 59.5 \\
    &&$f_r$         & 81.1        & 86.5      &  44.7         & 61.8         \\
    && $f_r+g_r$    & 82.6        & 86.6      &  45.0         & 62.0         \\
      \midrule
    \multirow{3}{*}{\rotatebox[origin=c]{90}{{(c) }}} & \multirow{3}{*}{\rotatebox[origin=c]{90}{{Params}}}
    & fix \((\alpha, \lambda, \phi, \rho, \kappa, \gamma)\)  &    77.9     &   86.6    &    38.0      &    56.8      \\  
        && fix \((\lambda, \phi, \rho, \kappa, \gamma)\)   &   80.6      & 86.6      &    42.8     &  59.4        \\  
    && learn all & 82.6        & 86.6      &  45.0         & 62.0         \\    
        \bottomrule
    \end{tabular}
}
\label{tab:ablation}
\vskip -0.18in
\end{wraptable}

Table~\ref{tab:ablation} shows ablation studies on \texttt{ICEWS14} and \texttt{WIKI}.
First, we analyze the contributions of different rule types, see part (a). 
For \texttt{ICEWS14}, $xy$-rules yield the highest MRR, which is expected since they form the core of our approach. 
Combining all rule types (row \textit{all}) leads to the highest scores.
However, the MRR scores do not simply add up due to overlapping predictions among rule types.
We further evaluate the importance of each rule type by removing them individually. This leads to a performance decrease, with the highest drop observed when $xy$-rules are excluded. This highlights the benefit of incorporating multiple rule types.
For \texttt{WIKI}, the study reveals a different behavior. Here, recurrency rules alone already achieve the same results as all rule types combined, suggesting two things: First, capturing only very simple dependencies leads to strong predictive performance;
and second, \ruc is robust against potential noise introduced by additional rules. 

Second, we analyze the benefit of our scoring functions (Table~\ref{tab:ablation}, part~(b)) by comparing four variants: 
(i) static confidence; (ii) recency only, using $f$ (Equation~\ref{eq:score_single}) with $g=0$; 
(iii) frequency only, using $g$ (Equation~\ref{eq:score_multi}) with $f=0$; 
and (iv) the default setting combining $f$ and $g$ (Equation~\ref{eq:conf_delta}). 
On both datasets, each temporal scoring function improves over static confidence. 
Recency yields the largest gain, while incorporating frequency provides a further improvement. 
Since the frequency score is computed over a sliding window of size $\mathcal{W}$ (Equation~\ref{eq:score_multi}), it 
measures how frequently a grounding occurs within the recent past, rather than over the entire history.
Consequently, the $g$ also incorporates a notion of recency. 
Overall, the results indicate that recency contributes most to the observed improvements, with frequency providing a complementary benefit.

{Third, we evaluate whether learning the confidence-function parameters is necessary, see part (c) of Table~\ref{tab:ablation}.
In variant (i), parameters \((\alpha, \lambda, \phi, \rho, \kappa, \gamma)\) are fixed to $(0.5, 0.1, 0, 0.01, 0, 1)$.
In variant (ii), only $\alpha$ is adapted to the static confidence of each rule, while the remaining parameters are fixed.
These settings result in an exponential curve with mild decay for $f_r$ and a small slope for $g_r$. Variant (iii) corresponds to the full \ruc approach.
Both fixed-parameter settings reduce performance:
Rule-specific parameters are important, as different rules require different confidence-function shapes. }

\section{Explanations}
\label{sec:explain}
\ruc provides a tool that generates, for given queries, an html-page displaying candidate rankings and the rules contributing to each prediction. 
It supports several use cases: 
First, it assists model development by uncovering open problems. Second, it can enable comparison with other models, thereby identifying differences in predictions and the data patterns underlying them.
And third, it provides explanations in real-world applications, showing reasons behind predictions and highlighting temporal patterns in a TKG. We illustrate the second use case with an example comparison between \ruc and RE-GCN.
Find details in the Supplementary Material (Section~G).

Figure~\ref{fig:example-explain} illustrates an explanation for the query \emph{(Alexis\_T., Consult, ?, 334)} 
with ground truth \emph{Evangelos\_V.} (\texttt{ICEWS14}).
In this case, \ruc ranks the correct entity at rank~1, whereas RE-GCN ranks it at position~8. 
The tool reveals that the top candidate’s score ($\approx 0.22$) is largely due to a single rule, which states that individuals who expressed intent to meet earlier are likely to consult later.
This rule has a confidence of 0.17 (recency score 0.171, frequency score~$-0.001$), and was triggered by a triple occurring once, four timestamps prior, with a confidence function depicted in Figure~\ref{fig:curve_combined}. By clicking on the PLOT link, this figure is shown to users. Further, the top candidate's score it substantially higher than subsequent candidates.
This explanation suggests a possible limitation of RE-GCN, that it may struggle to capture longer temporal dependencies (here, four timesteps) due to its restricted window size (three timesteps). 
\begin{figure*}
    \centering
    \includegraphics[width=1.0\linewidth]{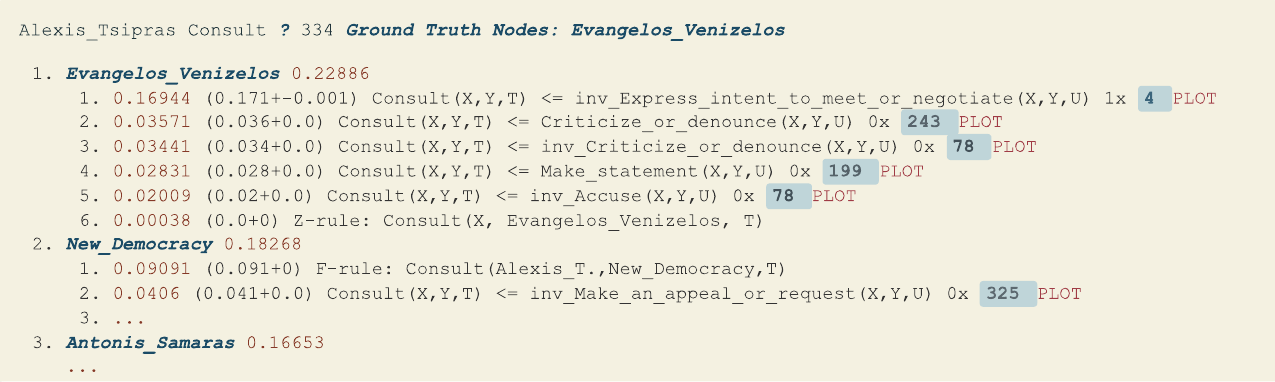}
    \vspace{-16pt}
    \caption{Explanation for the query \emph{(Alexis\_T., Consult, ?, 334)}}
    \label{fig:example-explain}
    \vspace{-10pt}
\end{figure*}

\section{Discussion}
\label{sec:conc}
We introduced an inherently interpretable rule-based approach for TKG forecasting. 
Each prediction can be traced back to the regularity and the specific observation that triggered it. Despite its simplicity, our method achieves competitive performance across diverse datasets, outperforming most existing approaches and reaching state-of-the-art on some datasets. 
Further, our method is the only one that can handle TKG of all sizes without OT and OM failures.

The strong performance of our simple rule-based approach suggests that, at least for the datasets and methods considered here, much of what is gained from complexity is in fact explainable by simple dependencies, and that complex end-to-end architectures are not necessary to achieve competitive predictive results.

Beyond strong empirical results, \ruc provides a transparent view into why predictions are made, allowing 
to trace outcomes back to concrete patterns in the data. 
This makes \ruc not only useful as a forecasting method, but also as a tool for understanding the behavior of existing methods and the properties of datasets.

\emph{Future Work}
Future work could analyze existing benchmarks to identify which queries genuinely demand complex reasoning beyond simple rules, and if the datasets provide the information such reasoning would need. Based on the insights, one could construct or identify datasets where simple rules fall short.
For \ruc specifically, a next step consists of a user study or quantitative interpretability evaluation. On the modeling side, incorporating multiple time granularities into a single confidence function, combining recent occurrences with summarized longer-term history, may benefit predictions for datasets like GDELT. Further, exploring alternative aggregation approaches for TKG forecasting, e.g., extending the ideas for static KGC from~\cite{betz2022supervised} to the temporal setting, is a promising direction for further research.

\newpage
\section*{Supplemental Material Statement} 
Source code, including code for the evaluation, is available in our GitHub repository \url{https://github.com/JuliaGast/counttrucola_submission}.
The hyperparameter ranges and values are discussed in the Supplementary Material. The Supplementary Material can also be found on our GitHub repository \url{https://github.com/JuliaGast/counttrucola_submission/blob/main/Supplementary_Material_CountTRuCoLa.pdf}.  

\section*{Declaration of use of Generative AI}
The authors used generative AI tools, specifically Lumo (Proton), Claude (Anthropic), and ChatGPT (OpenAI), during the preparation of this manuscript. These tools were used exclusively to assist with rephrasing and improving the clarity of text drafted by the authors. All scientific content, ideas, and conclusions are solely the work of the authors. The authors take full responsibility for the integrity of the submitted work.
%
%
\bibliographystyle{splncs04}
\bibliography{iswc26}

\newpage
\appendix
\section*{Supplementary Material}
\section{Special remark on $c$-rules}
\label{ap:crules}
In this section we provide a more detailed consideration of $c$-rules with the help of an example.
Consider a dataset where we observed several times that persons, who eats pizza at timestamp $t$, will drink espresso at a subsequent timestamp.
This regularity can be represented by the following $c$-rule:
\begin{align}
\label{r:cpizza}drinks(x,espresso, t^\star) \leftarrow eats(x,pizza, t)
\end{align}
This $c$-rule can be applied to answer a query (Q1) $drinks(sally, ?, t^\star)$.
Let us assume that Q1 has been derived from the triple $drinks(sally, espresso, t^\star)$. If we replace the subject instead of the object by a ?, this yields query (Q2') $drinks(?, espresso, t^\star)$. However, within our approach we do not support queries that ask for a subject directly. Instead, we translate them into equivalent object queries that use the inverse relation. We get (Q2) $drinks^{-1}(espresso, ?, t^\star)$.

Can we use the regularity expressed in Rule~\ref{r:cpizza} to make a prediction for (Q2)? Obviously not directly, because this rule uses a different head relation. If write down Rule~\ref{r:cpizza} with an inverse head, we will get the following rule: 
\begin{align}
	\label{r:cpizza-inv}drinks^{-1}(espresso, y, t^\star) \leftarrow eats^{-1}(pizza, y, t)
\end{align}
This rule is not supported by our language bias, because we do not support rules where the constant appears in subject position. This means that we are forced to resort to Rule~\ref{r:cpizza} for answering (Q2') directly. The candidates which are predicted as answer to (Q2') are then combined with the answers for Q2 that we might get from other rules.  

However, when selecting the training examples in Line 6 of Algorithm 1 we assumed that we have to deal with queries like (Q1) only. When asking a query as $drinks(sally, ?, t^\star)$ we can be sure that there is a correct candidate for the question mark and that is why we selected only those  examples where, with respect to our example regularity, something has been drunk. If we focus on queries as (Q2)  $drinks^{-1}(?, espresso, t^\star)$, we are in a different situation. We know already that Espresso has been drunk, and we ask who was drinking it. This means that for such a query we can limit our attention to training examples that fulfill the condition  $\exists z \ drinks(z,espresso,t^\star) \in G$.

Thus, we have to distinguish between two application scenarios for $c$-rules and for these two application scenarios we have to collect different example sets associated to a $c$-rule. If we are concerned with object-queries, queries that have the question mark in the object position, we use Algorithm 1 and especially Line 6 as depicted in the main paper. This means that we check the condition $\exists z \ h(c,z,t^\star) \in G$. If, on the contrary, we are concerned with subject queries, queries that have the question mark in the subject position, we replace the condition in Line 6 in Algorithm 1 by $\exists z \ h(z,d,t^\star) \in G$ where $d$ is the constant used in the head of the rule. As we distinguish between two examples sets associated to a $c$-rule, we also learn two confidence functions associated to each $c$-rule. Depending on the type of query, we apply one or the other confidence function to compute the confidence for a prediction. Within the code and the rule files we refer to these two variants as forward (F) and backward (B) rules respectively.

\section{Additional Information on Datasets}
\begin{table*}
\footnotesize
\centering
\caption{Dataset statistics. \# Quads refers to the number of quadruples without inverse quadruples.}
\resizebox{\textwidth}{!}{
\begin{tabular}{l|rrrrrrrrr}
\toprule
Dataset & \texttt{smallpedia} & \texttt{polecat} & \texttt{icews} & \texttt{wikidata} & \texttt{ICEWS14} & \texttt{ICEWS18} & \texttt{GDELT} & \texttt{YAGO} & \texttt{WIKI} \\ 
\midrule
\# Quads Train & 387,757 & 1,246,556 & 10,861,600 & 6,982,503 & 74,845 & 373,018 & 1,734,399 & 161,540 & 539,286 \\ 
\# Quads Valid & 81,033 & 266,736 & 2,326,157 & 1,434,950 & 8,514 & 45,995 & 238,765 & 19,523 & 67,538 \\ 
\# Quads Test & 81,586 & 266,318 & 2,325,689 & 1,438,750 & 7,371 & 49,545 & 305,241 & 20,026 & 63,110 \\ 
\# Nodes & 47,433 & 150,931 & 87,856 & 1,226,440 & 7,128 & 23,033 & 7,691 & 10,623 & 12,554 \\ 
\# Relations & 283 & 16 & 391 & 596 & 230 & 256 & 240 & 10 & 24\\ 
\# Timestamps & 125 & 1,826 & 10,224 & 2,025 & 365 & 303 & 2,975 & 188 & 231 \\ 
Granularity & year & day & day & year & day & day & 15 min & year & year \\ 
\bottomrule
\end{tabular}
}
\label{tab:datasetstats}
\end{table*}

\label{ap:datasets}
We provide additional details on the datasets used in our experiments. For each dataset, we use the version provided by~\cite{Li2021regcn} and~\cite{gastinger2023eval}, or, if lowercase letters only, the datasets introduced by~\cite{gastinger2024tgb}. An overview on the statistics of the datasets is in Table~\ref{tab:datasetstats}.

\begin{description}   
\item[ICEWS Datasets:]  
\texttt{ICEWS14}~\cite{GarciaDuran2018ICEWS14}, \texttt{ICEWS18}~\cite{Jin2019oldrenet}, and \texttt{icews}~\cite{gastinger2024tgb} are derived from the Integrated Crisis Early Warning System (ICEWS)~\cite{Boschee2015,shilliday2012data}. These datasets span different periods (2014, 2018, and 1995--2022) and contain event data on global political activities such as conflicts, protests, and diplomatic interactions. Events are categorized according to the CAMEO taxonomy~\cite{gerner2002conflict}.

\item[GDELT:]  
The Global Database of Events, Language, and Tone (\texttt{GDELT})~\cite{Leetaru2013GDELT} contains large-scale event data extracted from global news sources. It encompasses a wide range of political, societal, and cultural events across various countries and timeframes.

\item[{polecat}:]  
Based on the POLECAT (POLitical Event Classification, Attributes, and Types) dataset~\cite{Scar2023polecat}, this dataset records cooperative and hostile interactions between socio-political actors. POLECAT uses the PLOVER ontology~\cite{halterman2023plover} and automated NLP pipelines to classify and extract time-stamped, geolocated events from multilingual news sources. The dataset used in this work covers the period from January 2018 to December 2022.

\item[YAGO and WIKI:]  
\texttt{YAGO}~\cite{Mahdisoltani2015YAGO} and \texttt{WIKI}~\cite{Leblay2018WIKI} provide structured knowledge graph data with temporal relations. \texttt{WIKI} is extracted from Wikidata~\cite{vrandevcic2014wikidata} and both datasets have been further processed by~\cite{Jin2019oldrenet} to represent temporal facts as quadruples. Events before 1786 (\texttt{WIKI}) and 1830 (\texttt{YAGO}) are excluded.

\item[{smallpedia} and {wikidata}:]  
These datasets are constructed from Wikidata~\cite{vrandevcic2014wikidata} and processed by~\cite{gastinger2024tgb}. \texttt{smallpedia} includes entities with IDs below 1 million, while \texttt{wikidata} extends the scope to entities with IDs up to 32 million. Both datasets contain event-based (point-in-time) and fact-based (duration) temporal relations between entities.
\end{description}

\section{Details on Experimental Setup}
\subsection{Reasons for Excluding Prior Methods from Comparison}\label{ap:exp_relwork}
As noted in Section~5 in the main paper, 
we exclude 20 prior methods from direct comparison due to various limitations. In this section, we provide detailed justifications for each exclusion.

INFER reports result on the “best” ranking protocol in the presence of ties. 
As a result, the evaluation always assigns the best possible rank to the ground-truth entity in the case of ties, rather than using an average or random tie-breaking strategy.
This protocol is known to inflate evaluation metrics unfairly, since it does not reflect the true ranking uncertainty in the presence of score ties~\cite{Sun2020reevaluation}. Moreover, the provided code cannot be executed as it lacks the necessary specifications for pretraining the required Complex model.
L2TKG, StreamLearner, TPAR, Logenet, CRAFT, CoH, ALREIR and TECHS do not provide code to reproduce results.
For HERLN, it is unclear under which filtering setting (raw, static, or time-aware) the reported results were obtained. The released code computes raw and time-aware scores, and the paper does not specify which setting is used in the main experiments. In addition, the official implementation cannot be executed due to missing files and unspecified configurations, resulting in runtime errors. We opened GitHub issues and contacted the authors for clarification, but did not receive a response. 
CENET, RETIA, and CluSTER do not report results in time-aware filter setting.
CyGNet, RE-Net, and the approach for tensor factorization with timestamp encoding~\cite{dileo2026tensor} run only in multi-step setting, not in single-step setting.
TR-Rules uses different dataset versions for ICEWS14 and does not report results on WIKI, YAGO, GDELT, or the TGB 2.0 datasets. 
Similarly, GHNN uses different dataset versions for ICEWS14 and GDELT, and does not report results on the other datasets.
GenTKG evaluates on different versions of GDELT and YAGO, does not report results on WIKI or the TGB 2.0 datasets, and does not provide MRR scores for any dataset.
Finally, we do not compare to zrLLM because it uses a different evaluation setup, focusing on zero-shot relations with different datasets and on predicting previously unseen relations.

\subsection{Notes on Experiments for TempValid and Dimnet}
For both TempValid and DiMNet, we report results from our own experiments, conducted using the official public GitHub repositories\footnote{\url{https://github.com/Kaka1aaaaa/TempValid} for TempValid and \url{https://github.com/hhdo/DiMNet} for DiMNet}. In both cases, we carefully verified the evaluation protocols before running the experiments. 

Despite following the provided code bases and reported settings, we encountered difficulties in reproducing the scores presented in the original papers. Table~\ref{tab:results_tempvalid} reports our results and the reported ones. We could not determine the cause of these differences. One possible explanation are hyperparameter values that were not documented in the papers and repositories.
Whenever hyperparameters were specified in the original papers, we used those values. Otherwise, we relied on the default settings from the released code. For datasets that were not included in the original works, we selected hyperparameters based on the datasets that were most similar in size and temporal granularity (\texttt{GDELT} for \texttt{polecat} and \texttt{icews}; \texttt{WIKI} for \texttt{smallpedia} and \texttt{wikidata}), since no hyperparameter-tuning scripts were provided.

Additionally, for TempValid, we encountered technical issues when running experiments on \texttt{polecat}, \texttt{icews}, and \texttt{wikidata}. Execution terminated with a runtime exception (“ValueError: Values are too large to be losslessly converted to uint16.”). We suspect this is due to the larger scale of these datasets. The GitHub repository did not allow opening new issues.  
Furthermore, we observed out-of-memory (OM) errors in TempValid even when restricting execution to a single process as well as out-of-time (OT) errors.

\setlength{\tabcolsep}{1mm}
\begin{table*}
\caption{Results for TempValid and DiMNet on five benchmark datasets. Both papers did not report results on the datasest from TGB 2.0. OM means Out Of Memory (40~GB GPU or 500~GB RAM), - means that no results have been reported. The term orig refers to the results reported in the original papers, the term ours to the results from our experiments.}
\footnotesize
\centering
\resizebox{.8\textwidth}{!}{
\begin{tabular}{l|rr|rr|rr|rr|rr}
\toprule

& \multicolumn{2}{c|}{\texttt{GDELT}} & \multicolumn{2}{c|}{\texttt{YAGO}} & \multicolumn{2}{c|}{\texttt{WIKI}} & \multicolumn{2}{c|}{\texttt{ICEWS14}} & \multicolumn{2}{c}{\texttt{ICEWS18}}  \\
\midrule
                          & MRR  & H10 & MRR  & H10 & MRR  & H10 & MRR & H10 & MRR & H10      \\
\midrule
TempValid orig & 21.9  & 37.0  & 79.7    &  85.7 & 83.2  & 97.5 & 45.8 & 65.1 & 33.5 & 52.3 \\
TempValid ours & OM & OM & 46.7 & 50.4 & OT & OT & {45.5} & {64.5} & OT & OT \\
\midrule
DiMNet orig   & 21.9  & 37.5 & -  & -   & -  & - & 45.7 & 68.0 & 34.1  & 55.8 \\
DiMNet ours & 21.1 & 36.5 & 72.4 & 82.7 & 71.2 & 79.7 & 41.4 & 60.8 & 33.9 & 55.4\\
\bottomrule
\end{tabular}
}
\label{tab:results_tempvalid}
\end{table*}

\subsection{Negative Samples}
Following the TGB 2.0 evaluation framework, we use the provided negative samples for the large dataset \texttt{tkgl-wikidata}. Specifically, TGB 2.0 includes 1,000 negative samples per query, sampled based on the query relation type. As a result, evaluation on \texttt{tkgl-wikidata} is performed by ranking the correct entity against these 1,000 candidates rather than against all entities. For all other datasets, we compute scores by ranking against the full set of entities.

\section{Hyperparameters}
\label{ap:hyperparams}
\ruc comes with nine hyperparameters:
\begin{itemize}
    \item $\mathcal{P}$: A constant added to the denominator when computing confidences for $xy$- and $c$-rules.
    \item $\mathcal{P}_{f\text{-rules}}$: The corresponding constant used for $f$-rules.
    \item $\mathcal{C}$: A boolean flag indicating whether $c$-rules are used for a given dataset.
    \item $\mathcal{W}$: The window size defining the length of the time window for the examples $E$, i.e., the maximum value in $\Delta$.
    \item $\mathcal{M}$: A threshold defining the minimum number of data points required before learning the parameters of the frequency-based scoring function $g_r$. If a rule has fewer than $\mathcal{M}$ examples, all parameters of $g_r$ are set to zero to reduce noise.
    \item $\mathcal{Z}$: A scaling factor $\mathcal{Z} \in [0, 1]$ applied to the score predicted by the $z$-rules.
    \item $\mathcal{H}$ and $\mathcal{D}$: Two hyperparameters for rule aggregation. $\mathcal{H}$ specifies how many of the top-confidence rules to aggregate, and $\mathcal{D} \in [0, 1]$ is a decay factor used in our modified noisy-or model, where the $i$-th confidence score $s_i$ is weighted by $s_i \cdot \mathcal{D}^i$.
    \item $\mathcal{C}_{\textit{x-count}}$: A threshold controlling the construction of $c$-rules. A rule candidate is retained only if its head and body relation–object pairs occur at least \(\left\lfloor \frac{\mathcal{C}_{\textit{x-count}}}{5} \right\rfloor\) times in the training data. We further evaluate rule support over all head quadruples matching the rule by sampling five past timestamps per quadruple. A rule is kept if the resulting groundings cover more than $\mathcal{C}_{\textit{x-count}}$ distinct $x$-values in total. 
\end{itemize}

We perform grid search on the validation MRR to select the values of these hyperparameters for each dataset. For smaller datasets, we explore a broader range of values, while for larger datasets, we reduce the search space to limit runtime and energy consumption.
Table~\ref{tab:hyperparam_range} lists the tested ranges for each dataset. Note that the range for $\mathcal{W}$ varies depending on the total number of timesteps available in each dataset.
The value of $\mathcal{C}_{\textit{x-count}}$ is set depending on dataset size: If a dataset has more than $10^6$ quadruples, $\mathcal{C}_{\textit{x-count}}=20$, otherwise $\mathcal{C}_{\textit{x-count}}=3$. This provides a practical trade-off that enables scalable c-rule mining on large datasets, where exhaustive enumeration of all candidates is infeasible.

Table~\ref{tab:hyperparams} reports the hyperparameter values selected for each dataset.

\begin{table*}
\footnotesize
\centering
\caption{Hyperparameter ranges. We allow fewer values for larger datasets to reduce computation costs.}
\resizebox{\textwidth}{!}{
\begin{tabular}{p{4.35cm}|p{3.8cm}|p{2.25cm}|p{3.45cm}|p{1.8cm}}
\toprule
 & small \texttt{WIKI},~\texttt{ICEWS14},~\texttt{YAGO}, \texttt{smallpedia}  & medium 
\texttt{ICEWS18} & large \texttt{polecat},~\texttt{wikidata}, \texttt{GDELT}   & very large \texttt{icews}\\
\midrule
$\mathcal{P}$ \scriptsize{\texttt{(RULE\_UNSEEN\_NEGATIVES)}} & $\{0,1,2,3,5,10,20,30,100\}$ & $\{1,5,10,30,100\}$ & $\{1, 10, 30, 100\}$ & $\{1, 10, 30, 100\}$ \\
$\mathcal{P}_{f\text{-rules}}$ \scriptsize{\texttt{(F\_UNSEEN\_NEGATIVES)}} & $\{0, 1, 5, 10, 20, 30\}$ & $\{0,10,30\}$ & $\{10\}$ & $\{10\}$ \\
$\mathcal{C}$ \scriptsize{\texttt{(RULE\_TYPE\_C)}} & $\{{\text{True}, \text{False}}\}$ & $\{{\text{True}, \text{False}}\}$ & $\{{\text{True}, \text{False}}\}$ & $\{{\text{False}}\}$ \\
$\mathcal{W}$ \scriptsize{\texttt{(LEARN\_WINDOW\_SIZE)}} & $\{50, 100, 150\}$ \texttt{ICEWS14}, $\{2, 3, 5, 10, 30, 50, 100\}$ \texttt{WIKI},  $\{2, 3, 5, 10, 30, 50, 100\}$ \texttt{smallpedia}, $\{2, 3, 5, 10, 30, 30, 50\}$ \texttt{YAGO} & $\{50, 100, 150\}$ & $\{50, 100, 150\}$ \texttt{polecat}, $\{2, 3, 5, 10, 30, 50\}$ \texttt{wikidata},  $\{10, 30, 50, 100, 150, 200\}$ \texttt{GDELT} & $\{50, 100\}$  \\
$\mathcal{M}$ \scriptsize{\texttt{(DATAPOINT\_THRESHOLD\_MULTI)}} & $\{0, 10, 50\}$ & $\{0, 50\}$ & $\{0, 50\}$ & $\{50\}$ \\
$\mathcal{Z}$ \scriptsize{\texttt{(Z\_RULES\_FACTOR)}}  & $\{0, 0.1, 0.2, 0.3, 0.4, 0.5, 1.0\}$ & $\{0, 0.1, 0.5\}$ & $\{0, 0.1, 0.5\}$ & $\{0, 0.1, 0.5\}$ \\
$\mathcal{H}$ \scriptsize{\texttt{(NUM\_TOP\_RULES)}} & $\{5, 10, 50\}$ & $\{10, 50\}$ & $\{10\}$ & $\{10\}$ \\
$\mathcal{D}$ \scriptsize{\texttt{(AGGREGATION\_DECAY)}} & $\{1, 0.9, 0.8, 0.7, 0.6, 0.5, 0.4\}$ & $\{1, 0.8, 0.6\}$ & $\{0.8\}$ & $\{0.8\}$ \\
\bottomrule
\end{tabular}
}
\label{tab:hyperparam_range}
\end{table*}

\setlength{\tabcolsep}{1mm}
\begin{table*}
\caption{Hyperparameter values for each dataset, selected based on the validation MRR. The column \texttt{default} represents default hyperpararmeters that we represent, when no hyperparameter tuning can or should be conducted.}
\footnotesize
\centering
\resizebox{\textwidth}{!}{
\begin{tabular}{l|r|r|r|r|r|r|r|r|r|r}
\toprule
& {\texttt{GDELT}} & {\texttt{YAGO}} & {\texttt{WIKI}} & {\texttt{ICEWS14}} & {\texttt{ICEWS18}} & {\texttt{smallp.}} & {\texttt{polecat}} & {\texttt{icews}} & {\texttt{wikidata}} & \texttt{default} \\
\midrule
$\mathcal{P}$ \scriptsize{\texttt{(RULE\_UNSEEN\_NEGATIVES)}} & 30    & 30   & 1       & 30      & 100      & 3          & 30      & 100   & 30     &  30 \\
$\mathcal{P}_{f\text{-rules}}$ \scriptsize{\texttt{(F\_UNSEEN\_NEGATIVES)}} & 10    & 30   & 10      & 10      & 10       & 30         & 10      & 10    & 10    &  10 \\
$\mathcal{C}$ \scriptsize{\texttt{(RULE\_TYPE\_C)}} & False & True & True    & True    & True   & True       & False   & False & True &   True \\
$\mathcal{W}$ \scriptsize{\texttt{(LEARN\_WINDOW\_SIZE)}}  & 200   & 30   & 3       & 50      & 50     & 3          & 150     & 50    & 10     & 10  \\
$\mathcal{M}$ \scriptsize{\texttt{(DATAPOINT\_THRESHOLD\_MULTI)}} & 0     & 50   & 50      & 0       & 50       & 10         & 0       & 50    & 0     &  0 \\
$\mathcal{Z}$ \scriptsize{\texttt{(Z\_RULES\_FACTOR)}} & 0.1   & 0.1  & 0.1     & 0.1     & 0.1     & 0          & 0.1     & 0.1   & 0.1     &  0.1 \\
$\mathcal{H}$ \scriptsize{\texttt{(NUM\_TOP\_RULES)}} & 10    & 5    & 10      & 10      & 10      & 5          & 10      & 10    & 10   &   10  \\
$\mathcal{D}$ \scriptsize{\texttt{(AGGREGATION\_DECAY)}}  & 0.8   & 0.7  & 0.8     & 0.8     & 0.8     & 0.4        & 0.8     & 0.8   & 0.8  &  0.8 \\
$\mathcal{C}_{\textit{x-count}}$ \scriptsize{\texttt{(C\_X\_COUNT)}} & 20 & 3 & 3 &3 &3 & 3& 20 & - & 20 &  3 \\
\bottomrule
\end{tabular}
}
\label{tab:hyperparams}
\end{table*}

\section{{Hyperparameter Sensitivity}}
\begin{figure*}
    \centering
    \includegraphics[width=\linewidth]{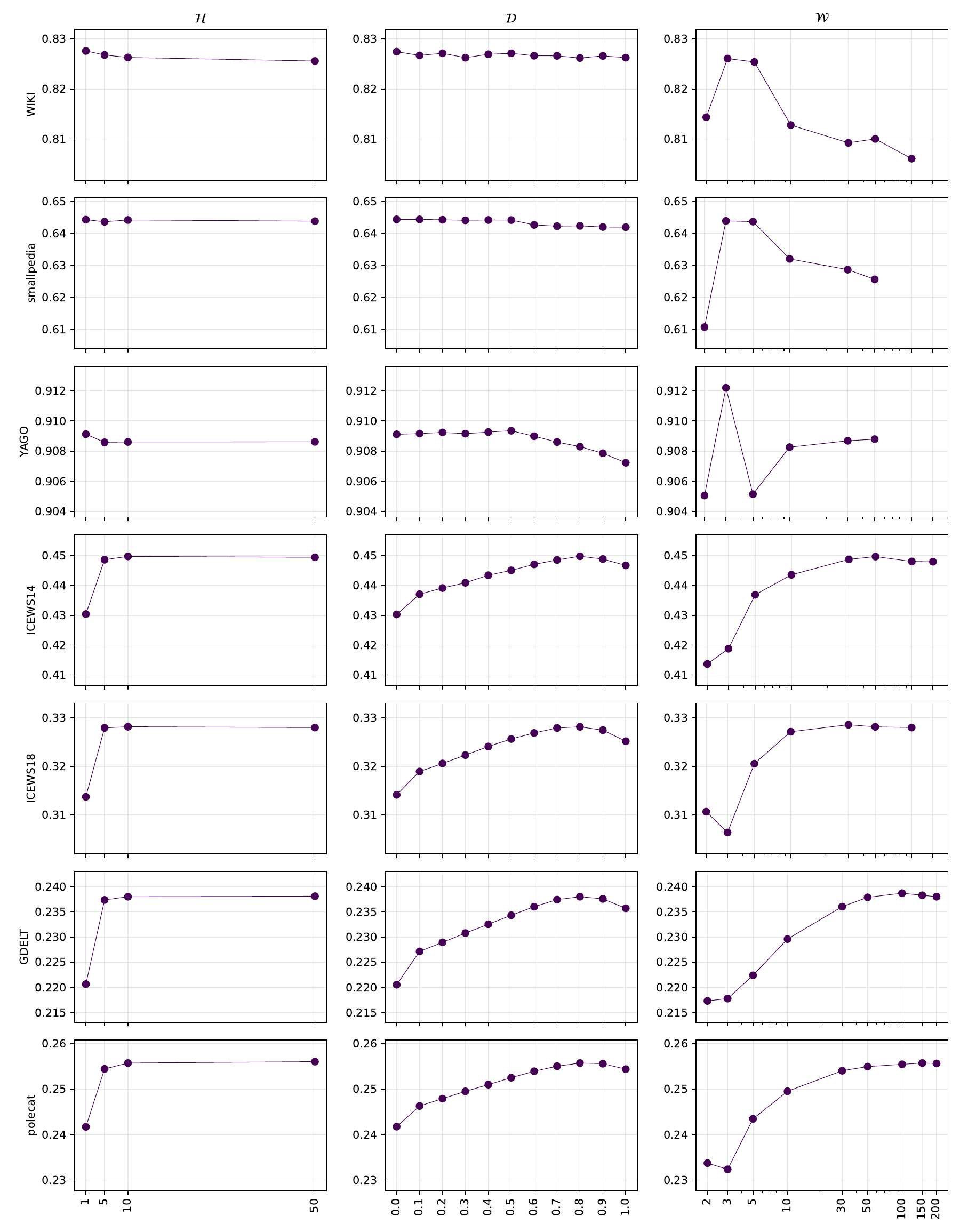}
    \caption{{Effect of the hyperparameters $\mathcal{H}$~{\texttt{(NUM\_TOP\_RULES)}}, $\mathcal{D}$~{\texttt{(AGGREGATION\_DECAY)}}, and $\mathcal{W}$~{\texttt{(LEARN\_WINDOW\_SIZE)}} on the test MRR. Columns correspond to hyperparameters (left to right), and rows correspond to datasets.}}
    \label{fig:hyperparams}
\end{figure*}

{Figure~\ref{fig:hyperparams} shows the effect of selected hyperparameters on the test MRR across six datasets. We analyze the impact of $\mathcal{H}$~{\texttt{(NUM\_TOP\_RULES)}}, $\mathcal{D}$~{\texttt{(AGGREGATION\_DECAY)}}, and $\mathcal{W}$~{\texttt{(LEARN\_WINDOW\_SIZE)}}. 
Note that the hyperparameters selected for \ruc (Table~\ref{tab:hyperparams}) are based on validation performance and do not necessarily result in the highest test MRR.}

\paragraph{{Learn Window Size $\mathcal{W}$:}}
{Across all datasets, $\mathcal{W}$ has the largest effect on performance. 
For the Wikidata- and Yago-based datasets (\texttt{WIKI}, \texttt{YAGO}, \texttt{smallpedia}), the best results are achieved with a small window of 3, while the other datasets benefit from larger windows. 
This can likely be explained by two factors.
First the Temporal granularity: Yearly datasets (\texttt{WIKI}, \texttt{YAGO}, \texttt{smallpedia}) cover multiple years even with small windows, whereas daily (\texttt{ICEWS14}, \texttt{ICEWS18}, \texttt{polecat}) or 15-min (\texttt{GDELT}) datasets require larger windows to capture sufficient history;
Second, the Recurrency of facts: As observed by~\cite{gastinger2024baselines,gastinger2024tgb}, \texttt{WIKI}, \texttt{YAGO}, and \texttt{smallpedia} exhibit high direct recurrency ($>85\%$), i.e., most temporal triples repeat in the previous timestep. Leveraging only recent history is sufficient, while incorporating older facts may introduce noise.}

\paragraph{{Number of Top Rules $\mathcal{H}$:}}
{For most datasets, performance improves with larger $\mathcal{H}$, with no substantial improvements between $\mathcal{H}=10$ and $\mathcal{H}=50$. 
Wikidata- and Yago-based datasets show a slight performance decrease as $\mathcal{H}$ increases, though the effect is minor. For a more detailed analysis of the effect of rule aggregation we refer to Section~\ref{ap:ruleag}.}

\paragraph{{Aggregation Decay $\mathcal{D}$:}}
{The aggregation decay $\mathcal{D}$ controls the assumed correlation between rules. 
$\mathcal{D}=0$ corresponds to max-aggregation strategy, i.e., equals $\mathcal{H}=1$, assuming highly correlated rules.
In contrast, $\mathcal{D}=1$ corresponds to noisy-or top-$\mathcal{H}$ aggregation, assuming independence between rules.
For most datasets, intermediate values ($0.5\leq \mathcal{D} \leq 0.8$) yield the best performance, reflecting partial correlation between rules. 
This aligns with the intuition that some rules are correlated, but not all. 
Wikidata-based datasets achieve the highest MRR with small $\mathcal{D}$ and $\mathcal{H}$, likely because predictions are dominated by recurrent rules, and additional correlated rules provide limited benefit.}

\paragraph{{Comparison to Default Hyperparameters}}
{Table~\ref{tab:hyperparams_default} compares test MRR when hyperparameters are selected based on validation performance (as reported in Appendix~\ref{ap:hyperparams} and Table~\ref{tab:hyperparams}) versus using default hyperparameter values, which are reported in the last column of Table~\ref{tab:hyperparams} and were chosen based on intuition and general insights.
The results indicate that hyperparameter selection generally improves performance. 
For example, the difference for \texttt{GDELT} ranges from 20.3 to 23.8, whereas other datasets, such as \texttt{polecat} and \texttt{YAGO}, are relatively stable, with differences of 0.6 and 0.1, respectively.}

\setlength{\tabcolsep}{1mm}
\begin{table*}
\caption{{Test MRRs when selecting hyperparameters for each dataset (based on the validation MRR) vs. when using standard hyperparameter values.}}
\footnotesize
\centering
\resizebox{\textwidth}{!}{
\begin{tabular}{l|r|r|r|r|r|r|r|r}
\toprule
& {\texttt{GDELT}} & {\texttt{YAGO}} & {\texttt{WIKI}} & {\texttt{ICEWS14}} & {\texttt{ICEWS18}} & {\texttt{smallp.}} & {\texttt{polecat}} & {\texttt{icews}}  \\
\midrule
selected per dataset &0.238	&0.909	&0.826&	0.450	&0.328	&0.644	&0.256	&0.214 \\
default hyperparams &0.203	&0.908&0.812	&0.444	&0.312	&0.630	&0.250	&0.208 \\
\bottomrule
\end{tabular}
}
\label{tab:hyperparams_default}
\end{table*}

\section{Additional Results}

\subsection{{Rule Aggregation Analysis}}\label{ap:ruleag}

\begin{figure*}
\vspace{-.5em}
    \centering
    \begin{minipage}[t]{0.9\linewidth}
        \centering
        (a) $\mathcal{H}=1$\\ 
        \includegraphics[width=.82\linewidth]{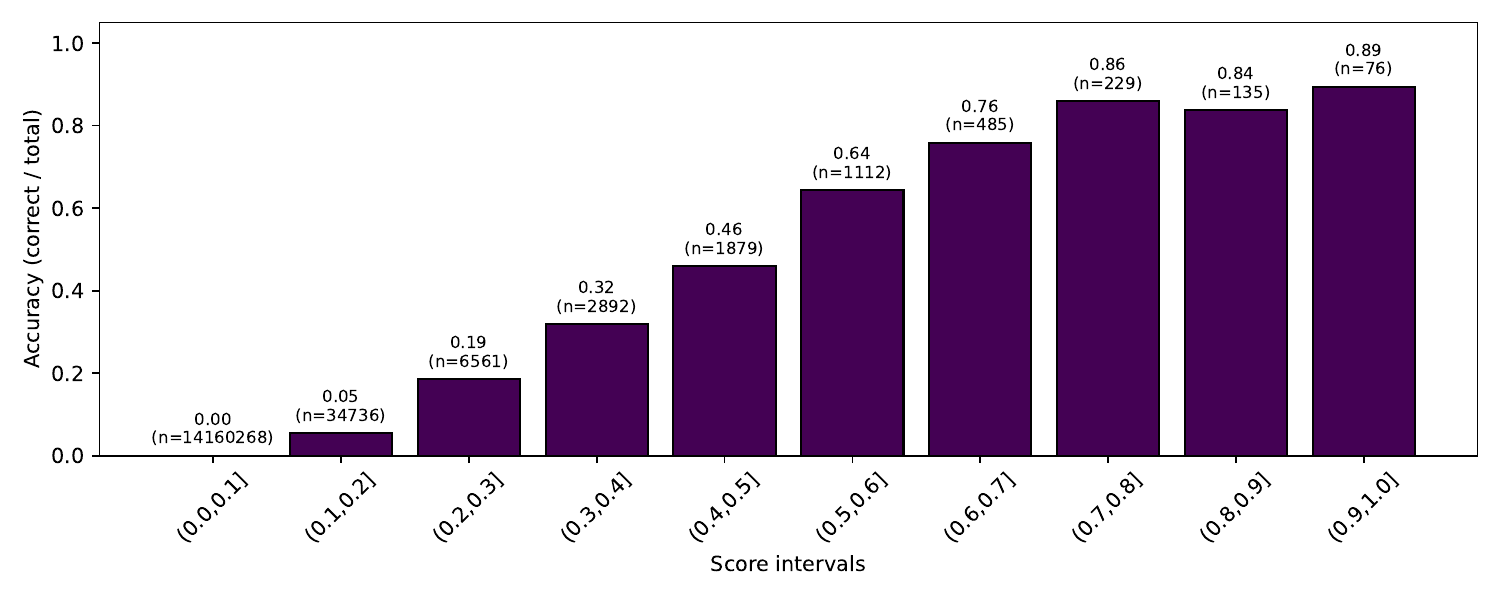} \\
        (b) $\mathcal{H}=100$ and $\mathcal{D}=1$ \\
        \includegraphics[width=.82\linewidth]{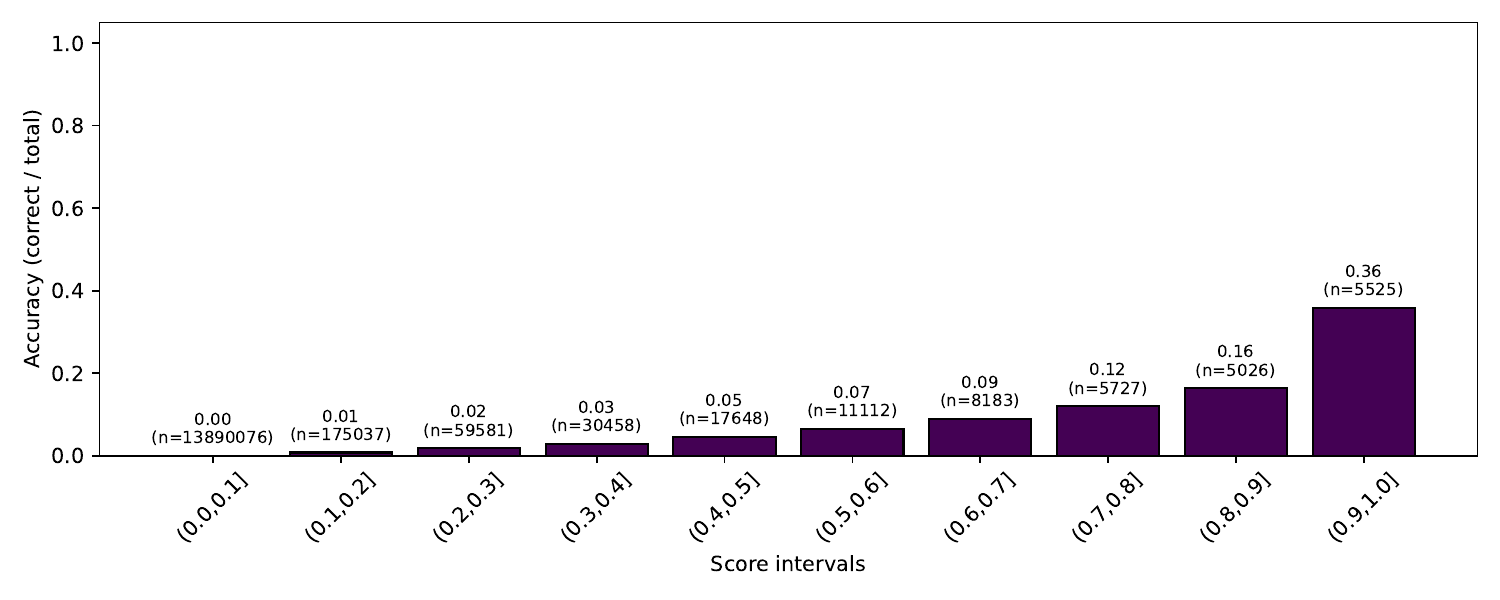} \\
        (c) $\mathcal{H}=10$ and $\mathcal{D}=1$ \\ 
        \includegraphics[width=.82\linewidth]{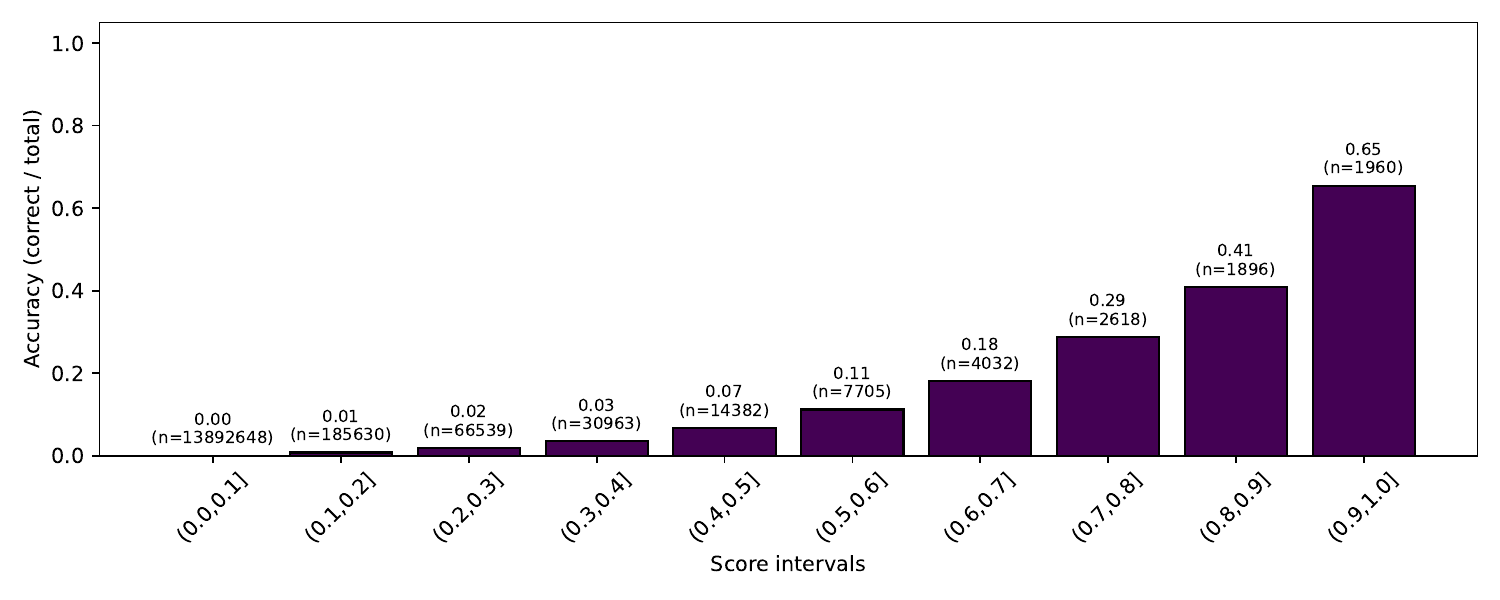} \\
        (d) $\mathcal{H}=10$ and $\mathcal{D}=0.8$\\
        \includegraphics[width=.82\linewidth]{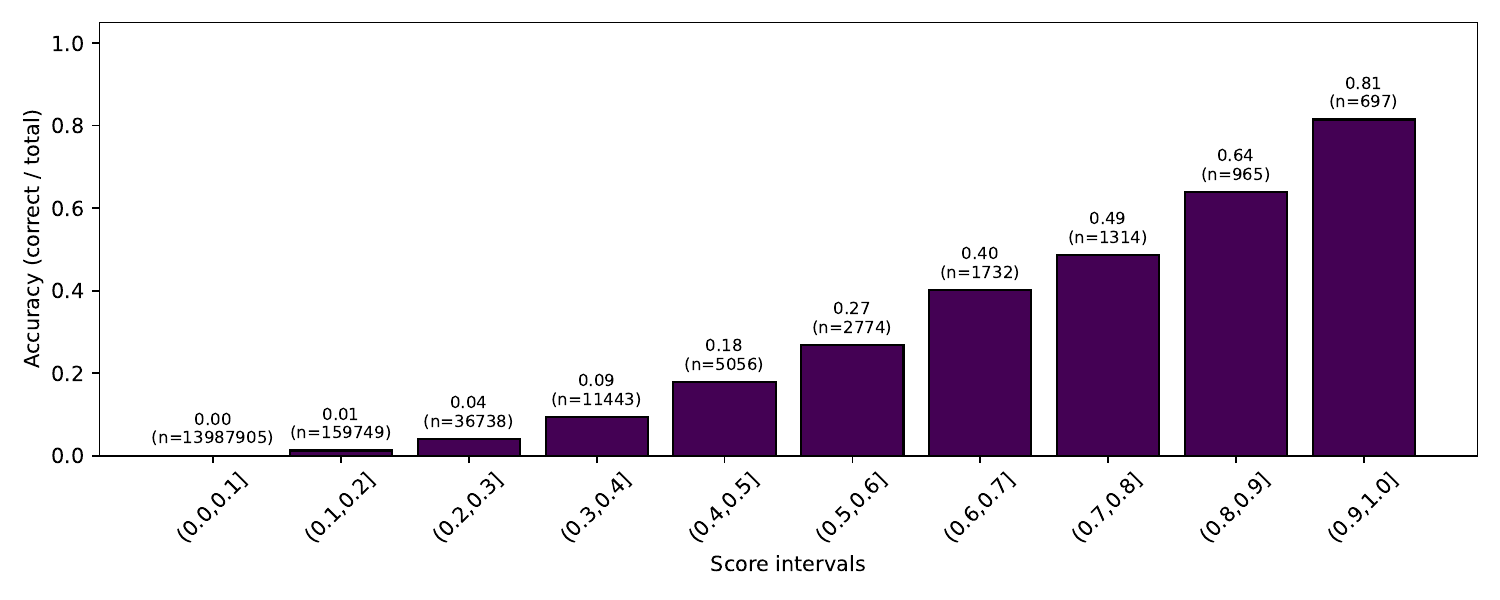}

        \caption{{Analysis of the relationship between aggregated rule confidence and empirical correctness for ICEWS14. Candidates are grouped into ten confidence bins, and the fraction of correct predictions in each bin is reported for four aggregation settings, where $\mathcal{H}$ represents \texttt{(NUM\_TOP\_RULES)}, and $\mathcal{D}$ \texttt{(AGGREGATION\_DECAY)}.}} 
        
        \label{fig:rulecor}
    \end{minipage}
\vspace{-2em}
\end{figure*} 
{To study how aggregation affects the relationship between predicted confidence and empirical correctnes we analyze the correspondence between aggregated confidence scores and ground-truth correctness under different aggregation configurations.
This analysis allows us to study how the aggregation hyperparameter $\mathcal{H}$ \texttt{(NUM\_TOP\_RULES)} and the decay hyperparameter $\mathcal{D}$ \texttt{(AGGREGATION\_DECAY)} influence the aggregated confidence values and the extent to which correlated rules may inflate them.}

\paragraph{{Analysis of Over- and Underestimation in Confidence Bins}}
{In this analysis, for each test query, we collected all predicted candidates together with their aggregated confidence scores and ground-truth labels (true or false). We then group the candidates in ten confidence bins $(i, i+0.1]$ with $i$ in ${0, 0.1, 0.2, …, 0.9}$ based on their assigned confidence score. For each bin, we computed the fraction of correct predictions.}

{The resulting curves indicate whether higher aggregated scores correspond to higher accuracy, thereby providing a diagnostic of whether \ruc produces meaningful confidence values.
We report results for four aggregation strategies: 
(a) max-aggregation ($\mathcal{H}=1$), which uses only the highest-scoring rule and thus assumes strong dependence among rules;
(b) top-100 noisy-or aggregation ($\mathcal{H}=100$ and $\mathcal{D}=1$), which combines the 100 highest-scoring rules under a stronger independence assumption;
(c) top-10 noisy-or aggregation ($\mathcal{H}=10$ and $\mathcal{D}=1$), which combines the ten highest-scoring rules under a slightly weaker independence assumption; and 
(d) decayed noisy-or aggregation ($\mathcal{H}=10$ and $\mathcal{D}=0.8$) which reduces the influence of lower-ranked rules and thus aims to mitigate potential inflation from correlated rule sets.}

{Figure~\ref{fig:rulecor} reports the results for \texttt{ICEWS14}.
The max-aggregation strategy (a) exhibits underestimation: Specifically, bins in the range 0.5–0.8 achieve higher empirical accuracies than their assigned confidence values, and relatively few candidates receive scores above 0.9.
In contrast, the top-100 noisy-or strategy (b) shows substantial overestimation, with predicted scores exceeding accuracy across all bins and a substantially larger fraction of candidates receiving scores above 0.9.
The top-10 noisy-or strategy (c) still overestimates confidence, but the inflation effect is noticeably reduced.
Finally, the decayed noisy-or strategy (d) shows intermediate behavior.
While a slight degree of overestimation remains, its predicted confidence values more closely match the empirical accuracies, and a monotonic relationship between confidence bins and accuracy is preserved.}

\paragraph{{Analysis of Over- and Underestimation in Top-1 Predictions}}
{To complement the analysis above, we further examine the confidence assigned to the top-ranked predictions.  
Table~\ref{tab:h_d_overestimation} reports, for different $(\mathcal{H}, \mathcal{D})$ settings, the achieved Hits@1 and the mean  score assigned to the top-scoring candidate on \texttt{ICEWS14}.  
Ideally, the assigned score should align with the achieved Hits@1. Systematically higher scores indicate overestimation, for example due to overcounting correlated rules, while systematically lower scores indicate underestimation.}

{The results reflect the expected behavior. 
Max-aggregation underestimates confidence: with $\mathcal{H}=1$ and $\mathcal{D}=1$, the model assigns a lower score than the observed Hits@1, reflecting overly strong dependence assumptions. In contrast, a large $\mathcal{H}=100$ leads to clear overestimation, with the predicted score far exceeding the Hits@1. This demonstrates that noisy-or indeed overcounts correlated rules when many rules are aggregated. A moderate setting of $\mathcal{H}=10$ partially alleviates this effect, though still produces overconfident estimates.
Introducing a decay factor ($\mathcal{H}=10$, $\mathcal{D}=0.8$) results in a closer match between predicted scores (42.2) and Hits@1 (35.9), while also producing the highest Hits@1 among the tested configurations.  }

{Together with the trends in Figure~\ref{fig:rulecor}, these results provide empirical evidence that, while not being able to fully hinder them, the decay parameter mitigates overcounting effects.} 

\begin{table*}
\caption{{Hits@1 and the mean score assigned to the top-scoring candidate for different $(\mathcal{H}, \mathcal{D})$ settings on \texttt{ICEWS14}.}}
\footnotesize
\centering
\resizebox{0.3\textwidth}{!}{
\begin{tabular}{l|l|r|r}
\toprule
$\mathcal{H}$ &$\mathcal{D}$	&Hits1	& Mean Score\\
\midrule
1	&1	&32.9	& 27.1 \\
100	&1	&34.0	&56.6\\
10	&1	&35.6	&51.3\\
10	&0.8	&35.9	&42.2\\
\bottomrule
\end{tabular}
}
\label{tab:h_d_overestimation}
\end{table*}

\subsection{{Analysis of Parameter Sensitivity to Learn Window Size}}\label{ap:window_param}
\begin{figure*}[]
    \centering
    
    \begin{subfigure}[t]{0.32\linewidth}
        \caption{\texttt{ICEWS14}}
        \includegraphics[width=\linewidth]{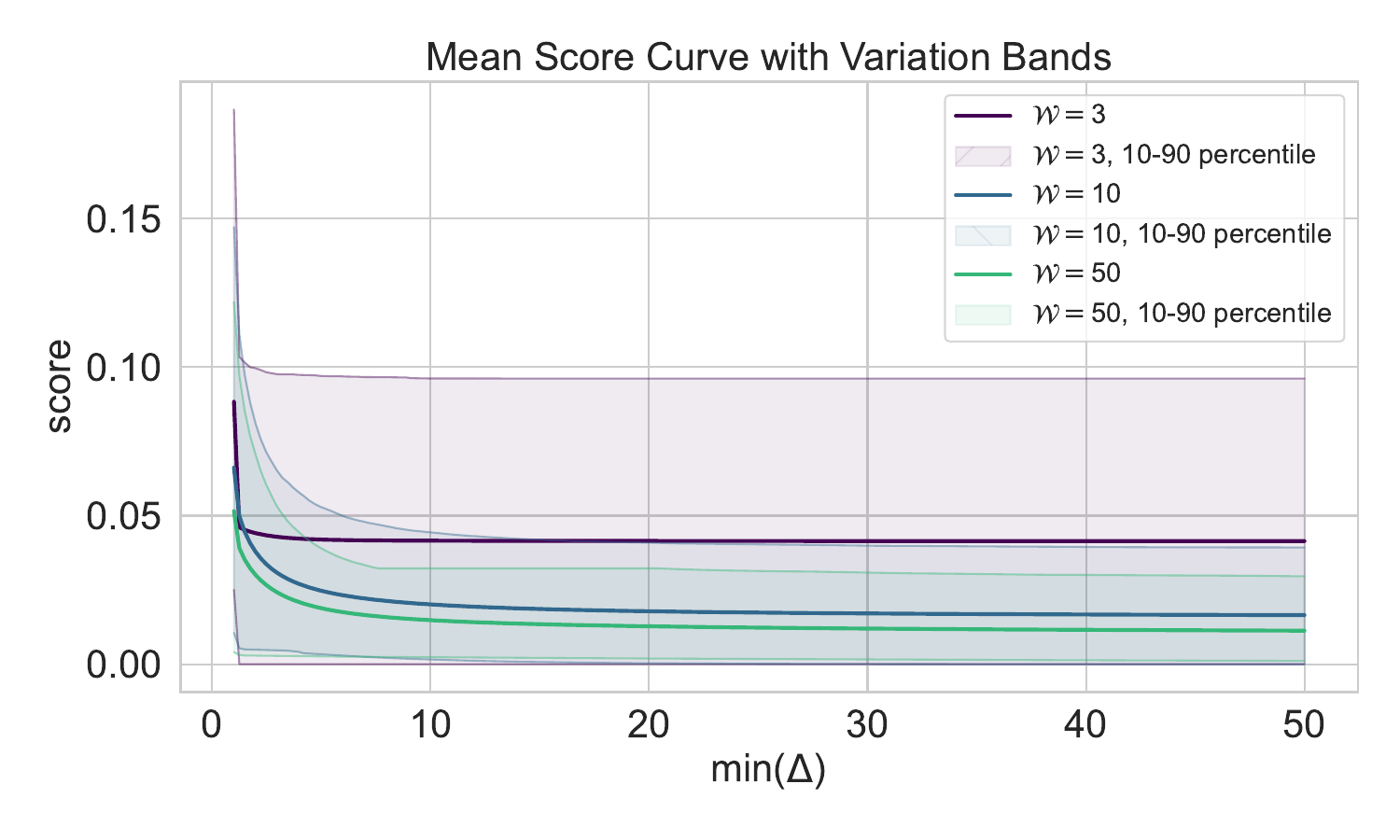}
    \end{subfigure}   
    \hspace{0.1em}
    \begin{subfigure}[t]{0.32\linewidth}
        \caption{\texttt{GDELT}}
        \includegraphics[width=\linewidth]{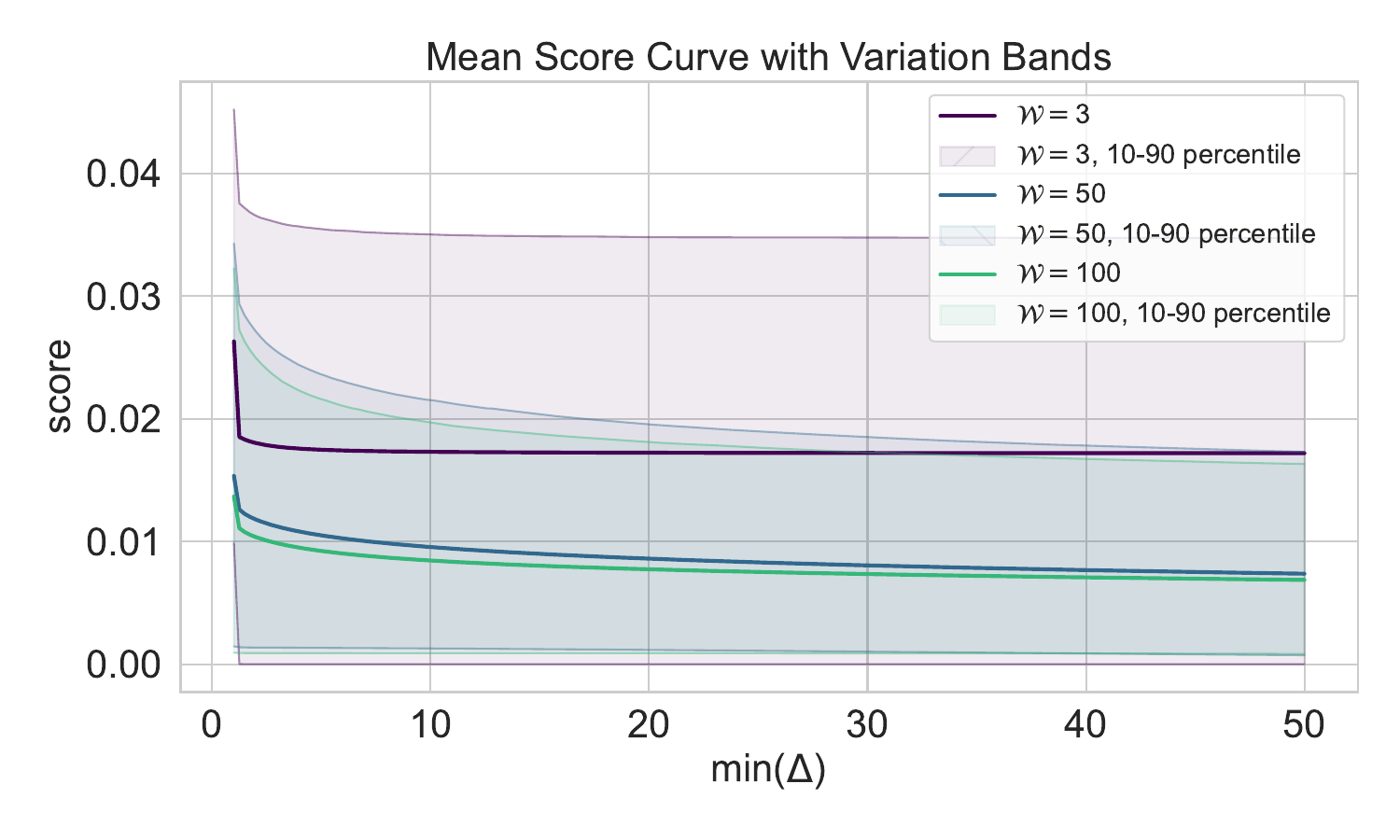}
    \end{subfigure}  
    \hspace{0.1em}
    \begin{subfigure}[t]{0.32\linewidth}
        \caption{\texttt{WIKI} }
        \includegraphics[width=\linewidth]{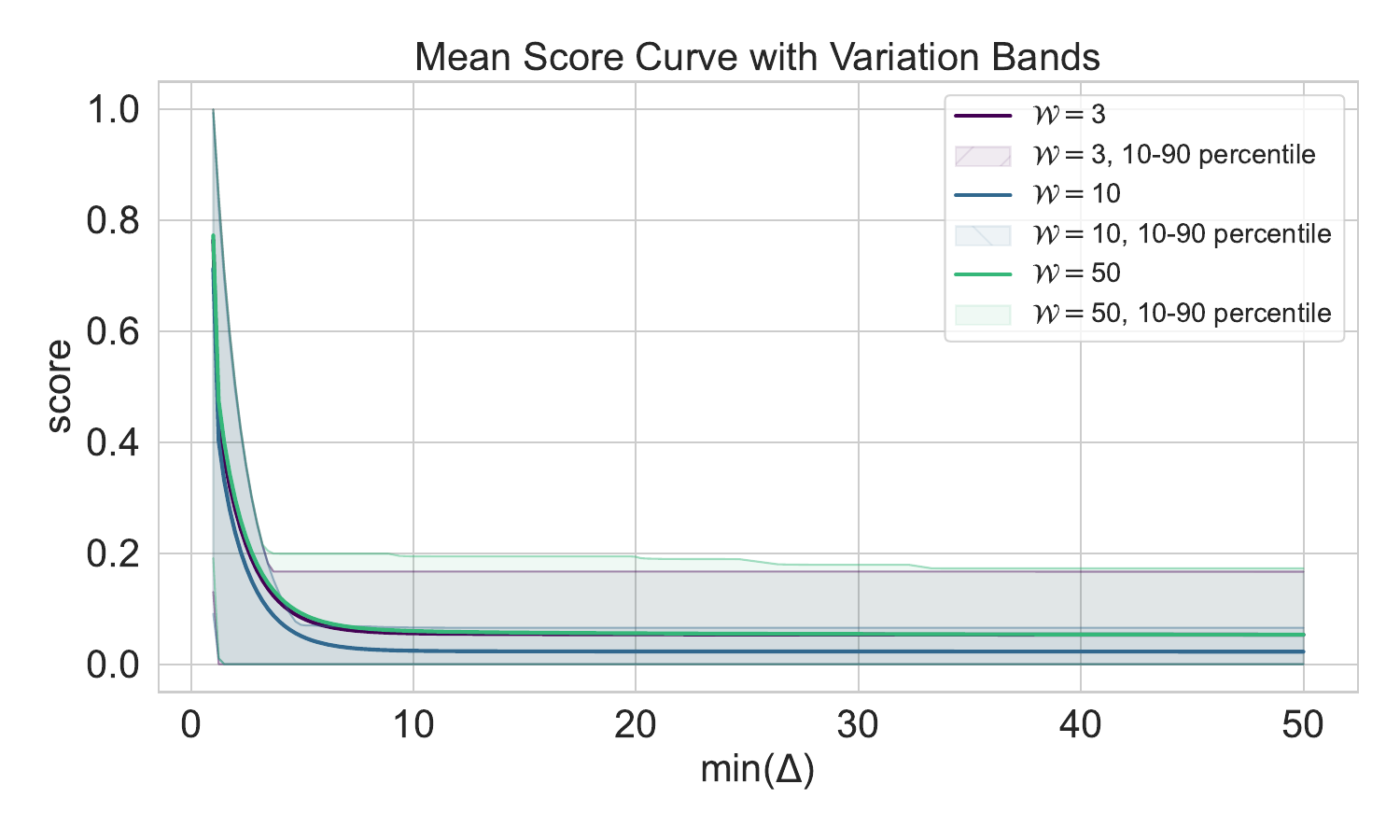}
    \end{subfigure}  
    \hfill
    \begin{subfigure}[t]{0.32\linewidth}
        \includegraphics[width=\linewidth]{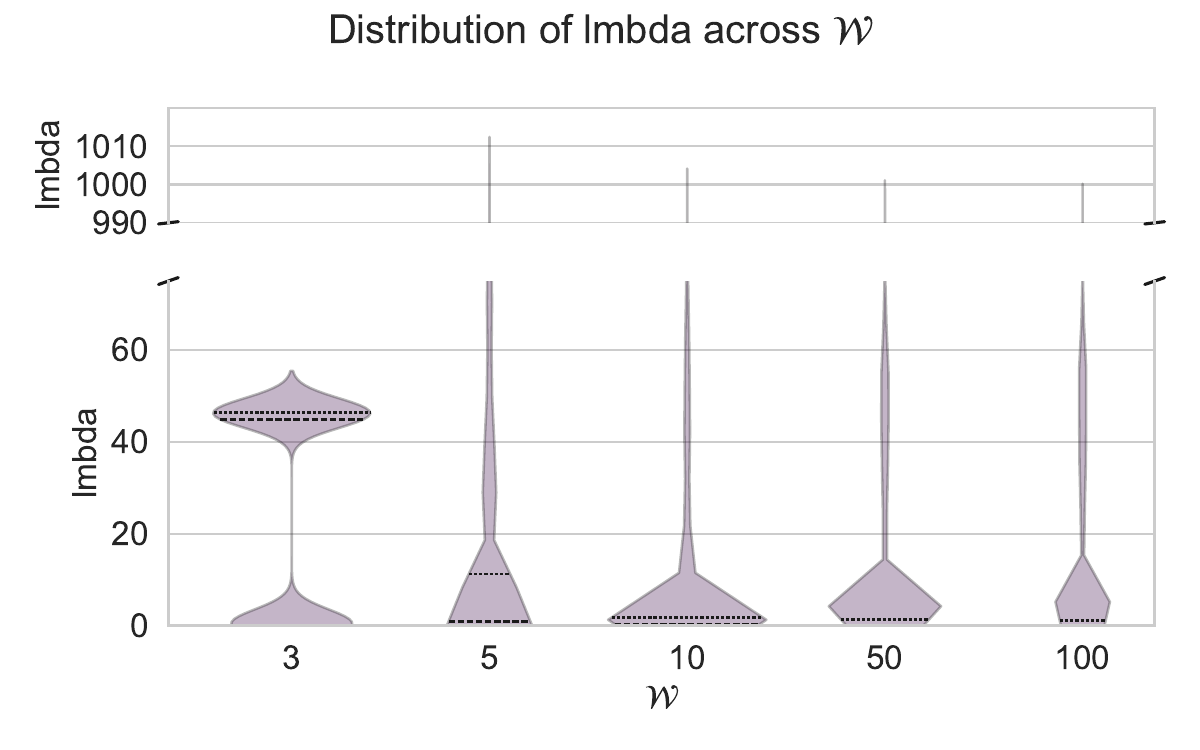}
    \end{subfigure}   
    \hspace{0.1em}
    \begin{subfigure}[t]{0.32\linewidth}
        \includegraphics[width=\linewidth]{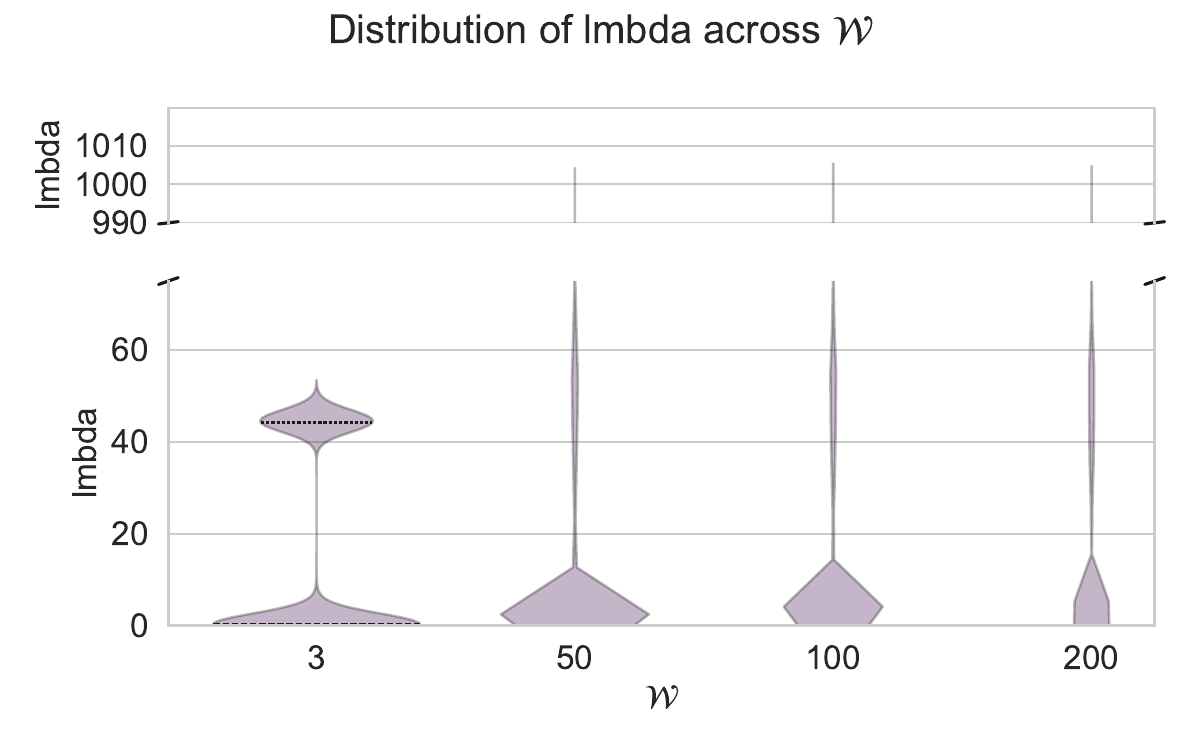}
    \end{subfigure}   
    \hspace{0.1em}
    \begin{subfigure}[t]{0.32\linewidth}
        \includegraphics[width=\linewidth]{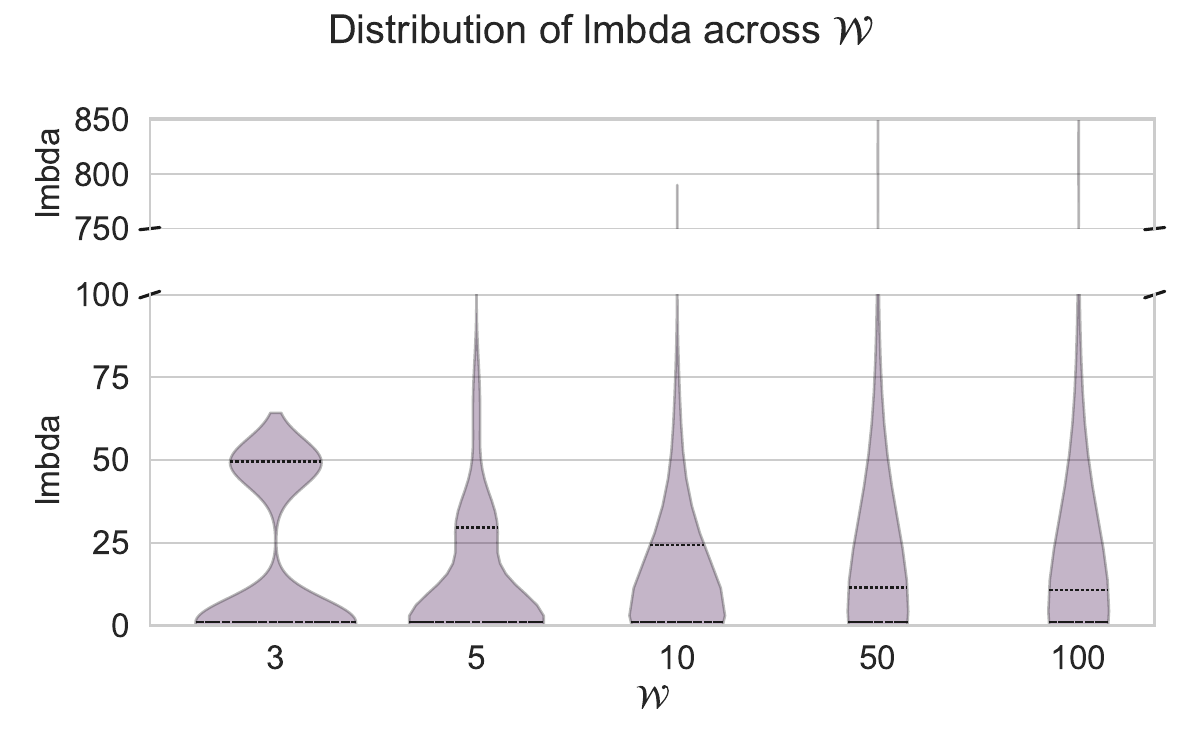}
    \end{subfigure}   
    \hfill
    \begin{subfigure}[t]{0.32\linewidth}
        \includegraphics[width=\linewidth]{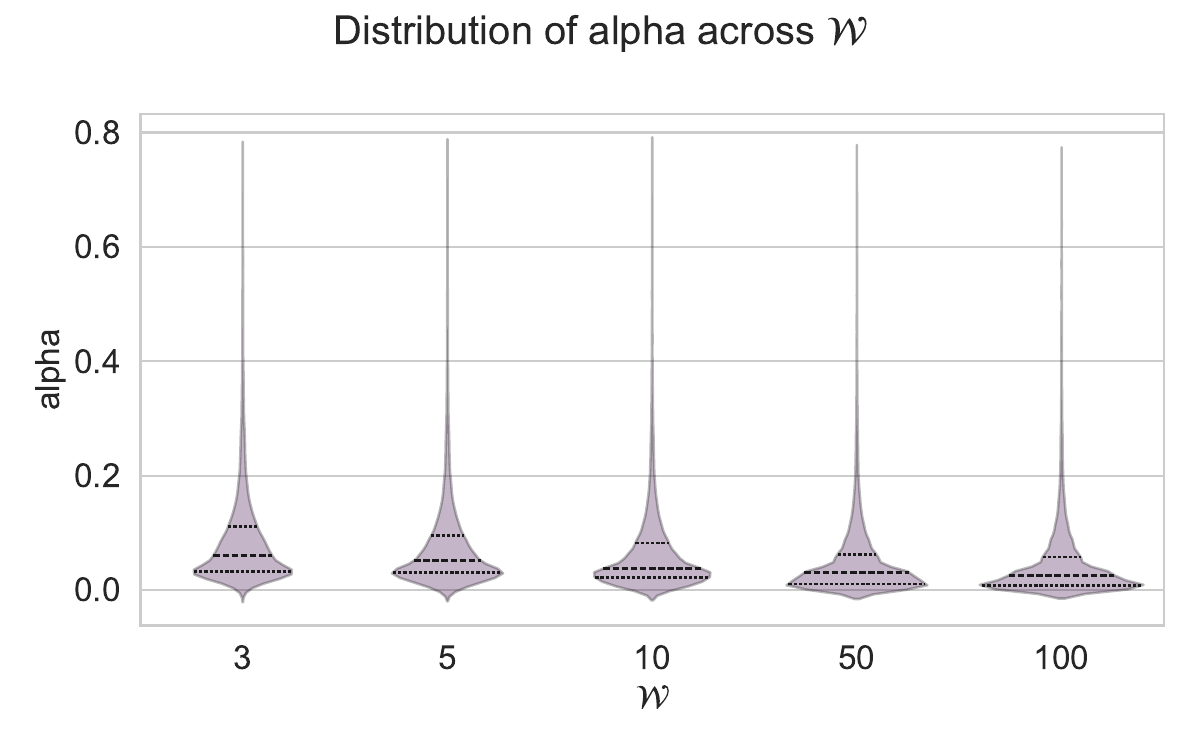}
    \end{subfigure}   
    \hspace{0.1em}
    \begin{subfigure}[t]{0.32\linewidth}
        \includegraphics[width=\linewidth]{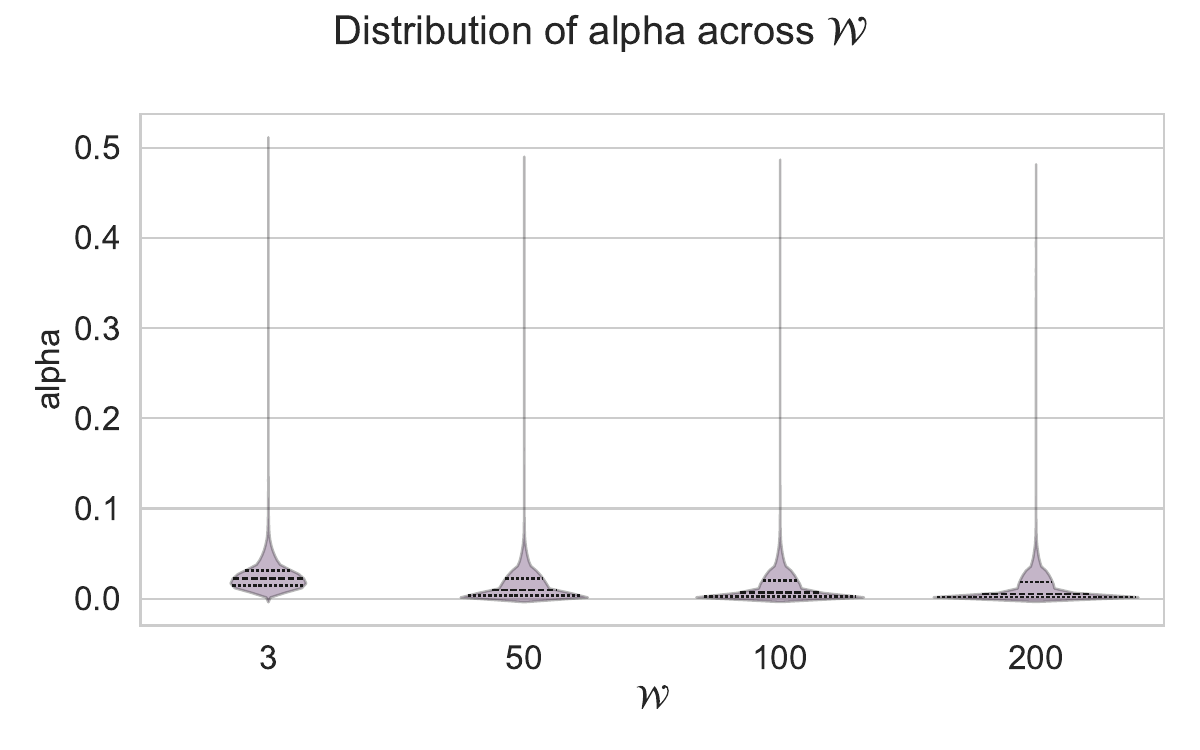}
    \end{subfigure}   
    \hspace{0.1em}
    \begin{subfigure}[t]{0.32\linewidth}
        \includegraphics[width=\linewidth]{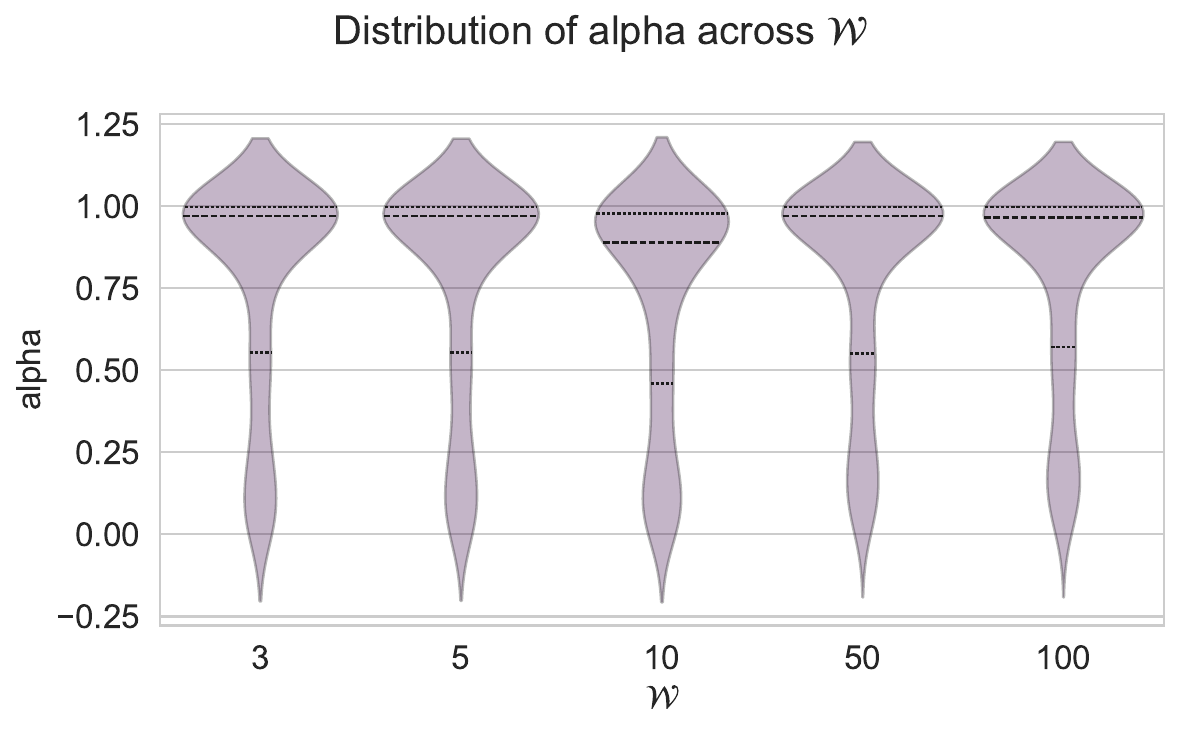}
    \end{subfigure}   
    \hfill 
    \begin{subfigure}[t]{0.32\linewidth}
        \includegraphics[width=\linewidth]{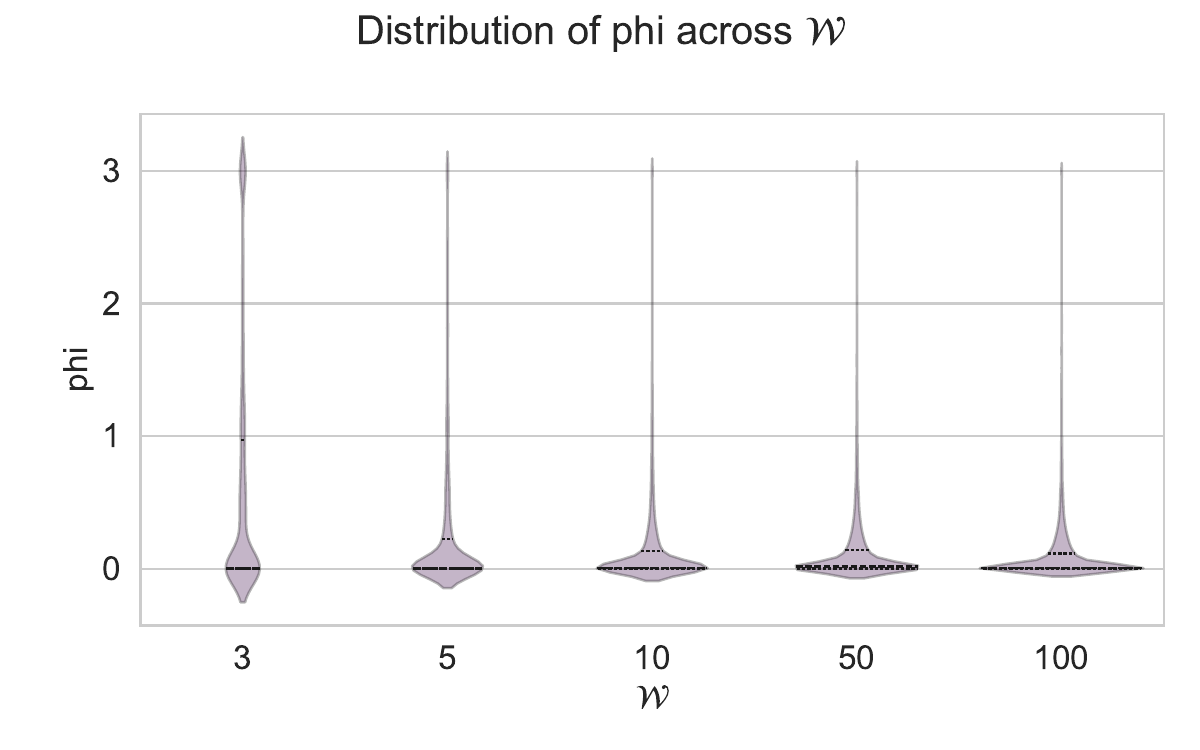}
    \end{subfigure}   
    \hspace{0.1em}
    \begin{subfigure}[t]{0.32\linewidth}
        \includegraphics[width=\linewidth]{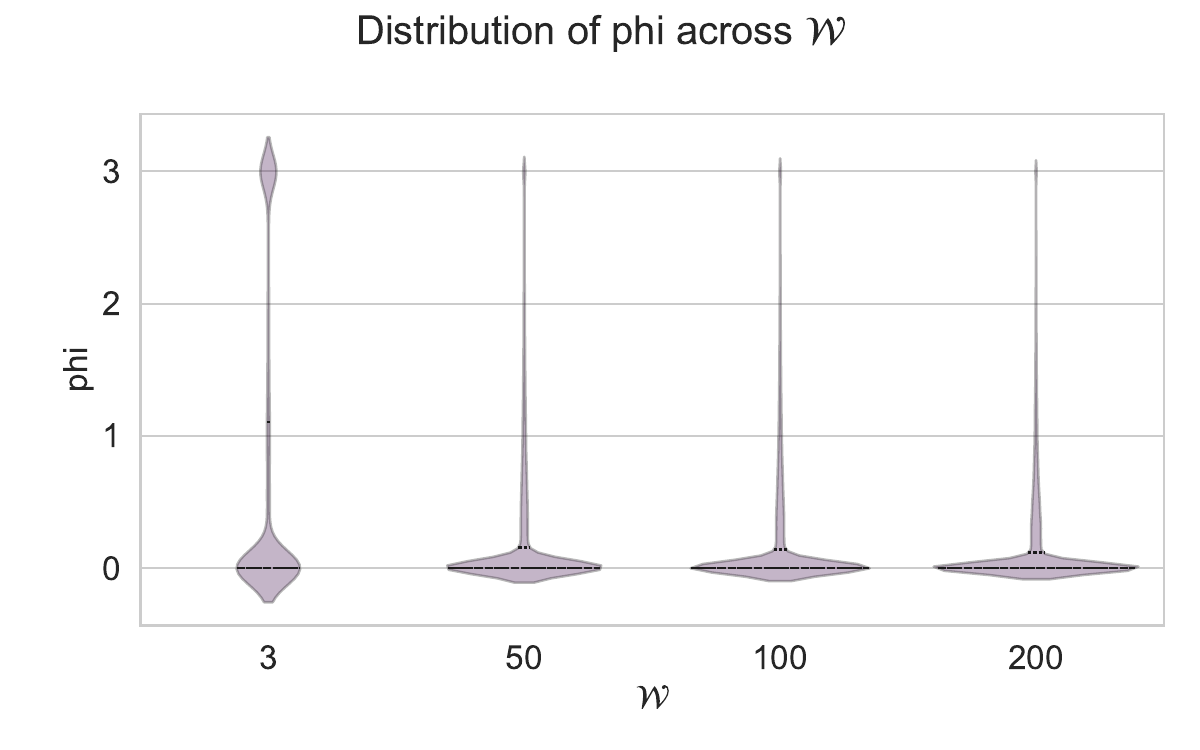}
    \end{subfigure}   
    \hspace{0.1em}
    \begin{subfigure}[t]{0.32\linewidth}
        \includegraphics[width=\linewidth]{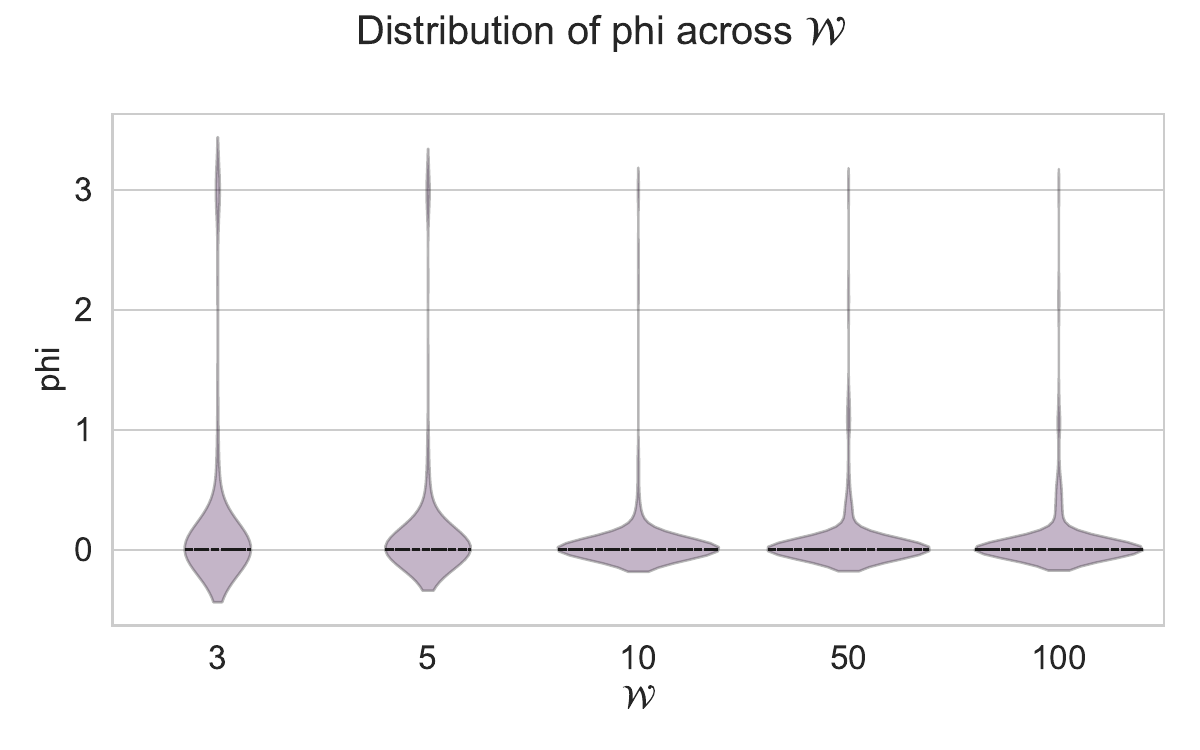}
    \end{subfigure}   
    \caption{
    {Recency function $f$ and the effect of Learn Window Size $\mathcal{W}$ on learned parameters for datasets \texttt{ICEWS14}, \texttt{GDELT}, and \texttt{WIKI} (left to right). Note that the y-axis varies across datasets.}
    }
    \label{fig:window_f}
    \vspace{-16pt}
\end{figure*}

\begin{figure*}[]
    \centering
    
    \begin{subfigure}[t]{0.32\linewidth}
        \includegraphics[width=\linewidth]{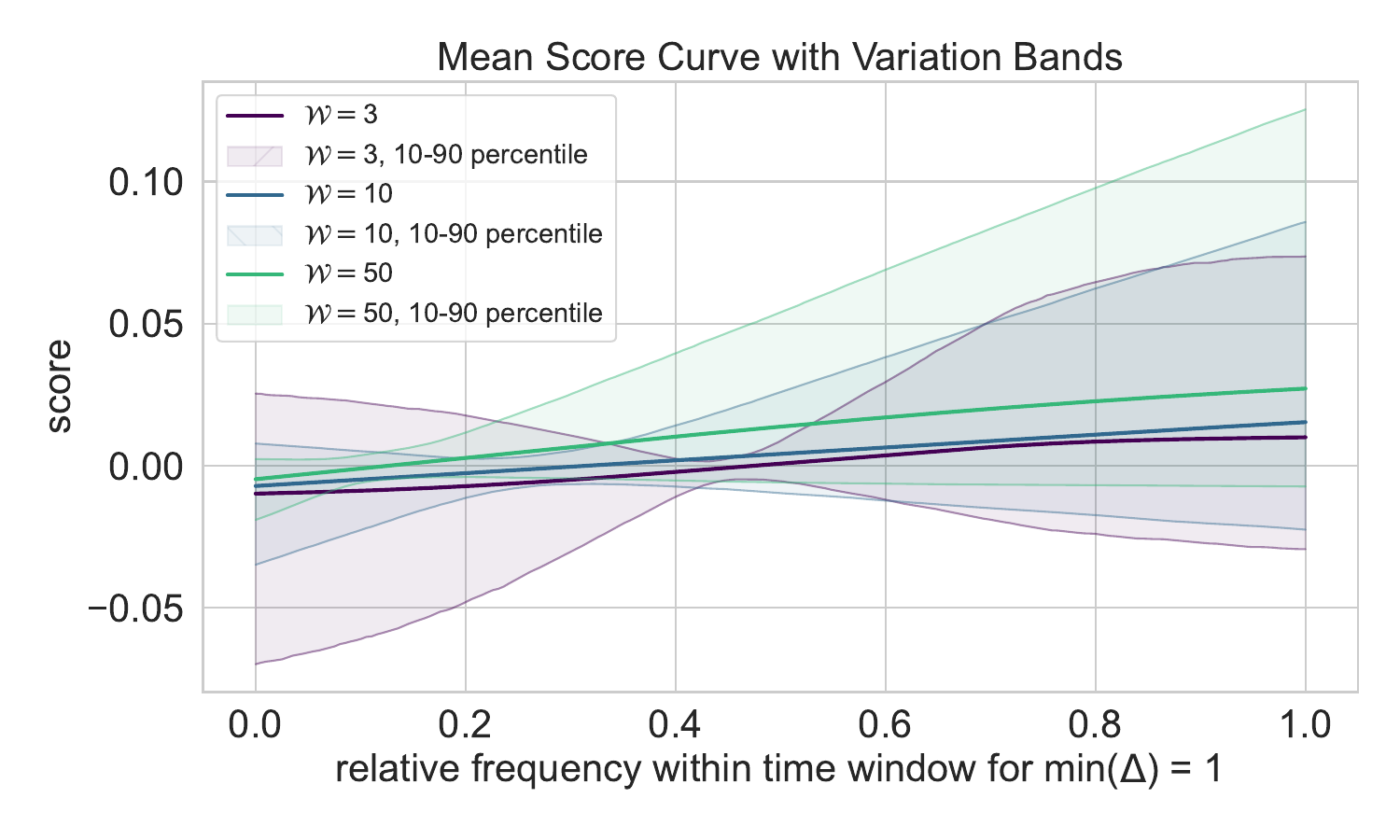}
    \end{subfigure}   
    \hspace{0.1em}
    \begin{subfigure}[t]{0.32\linewidth}
        \includegraphics[width=\linewidth]{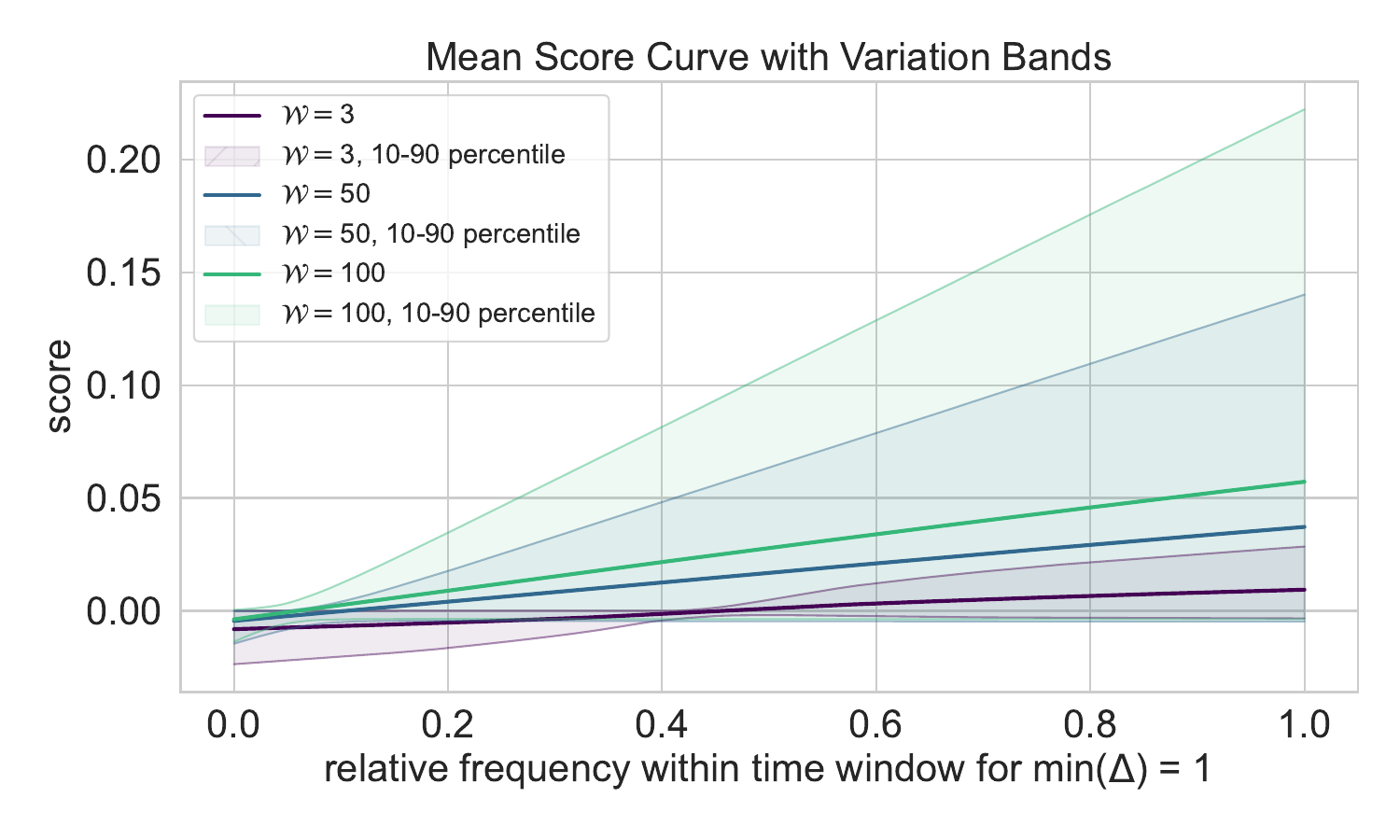}
    \end{subfigure}  
    \hspace{0.1em}
    \begin{subfigure}[t]{0.32\linewidth}
        \includegraphics[width=\linewidth]{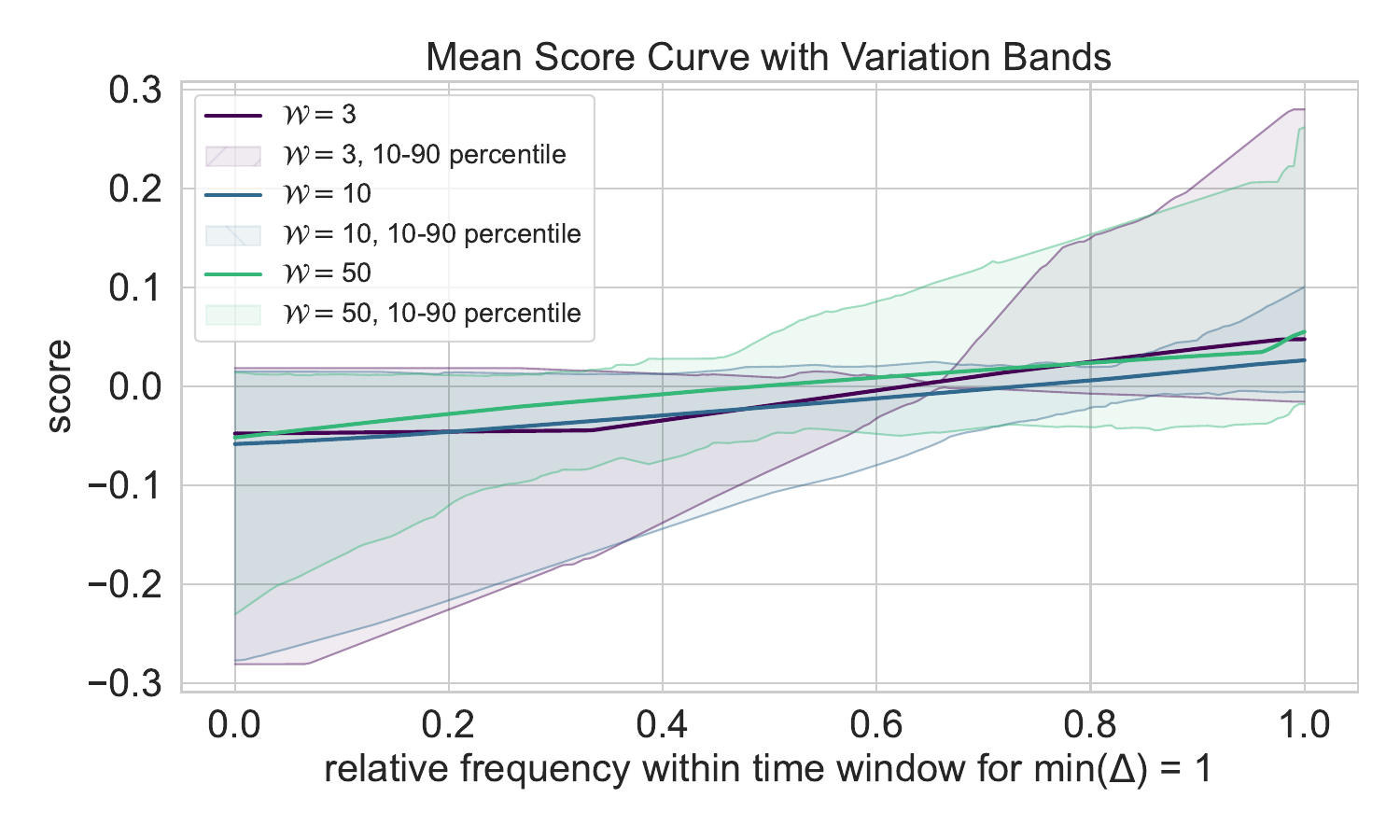}
    \end{subfigure}  
    \hfill
    \begin{subfigure}[t]{0.32\linewidth}
        \includegraphics[width=\linewidth]{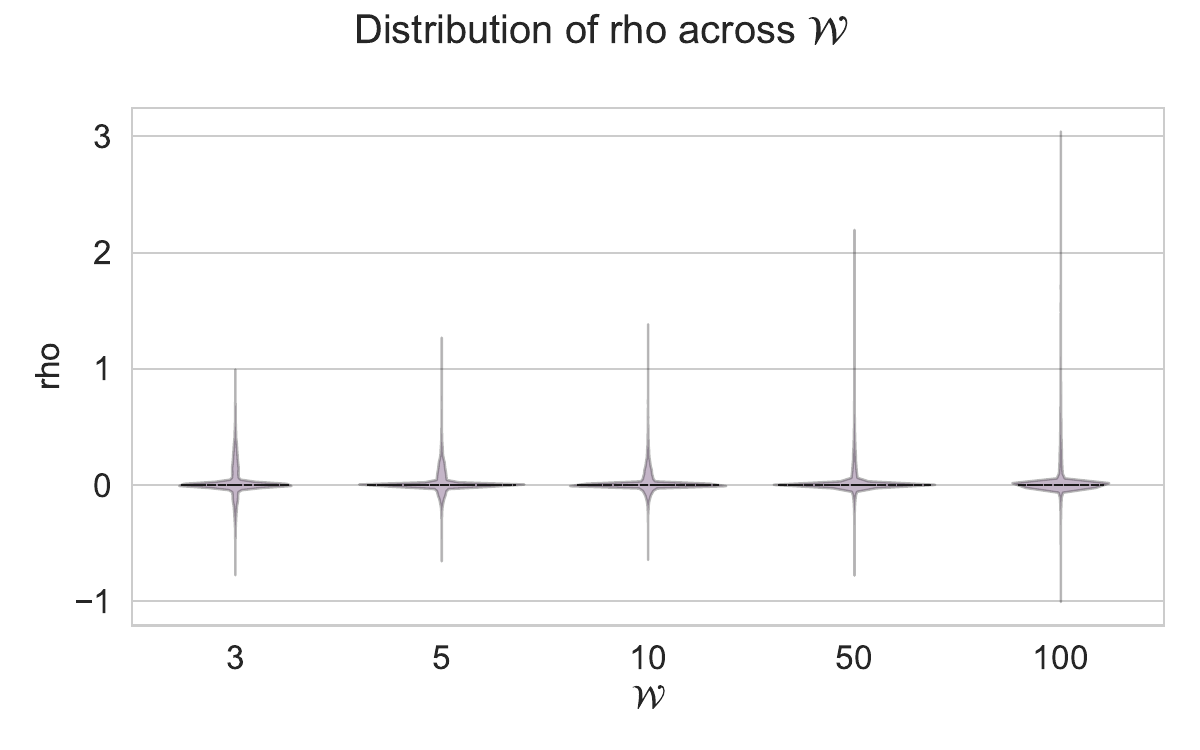}
    \end{subfigure}   
    \hspace{0.1em}
    \begin{subfigure}[t]{0.32\linewidth}
        \includegraphics[width=\linewidth]{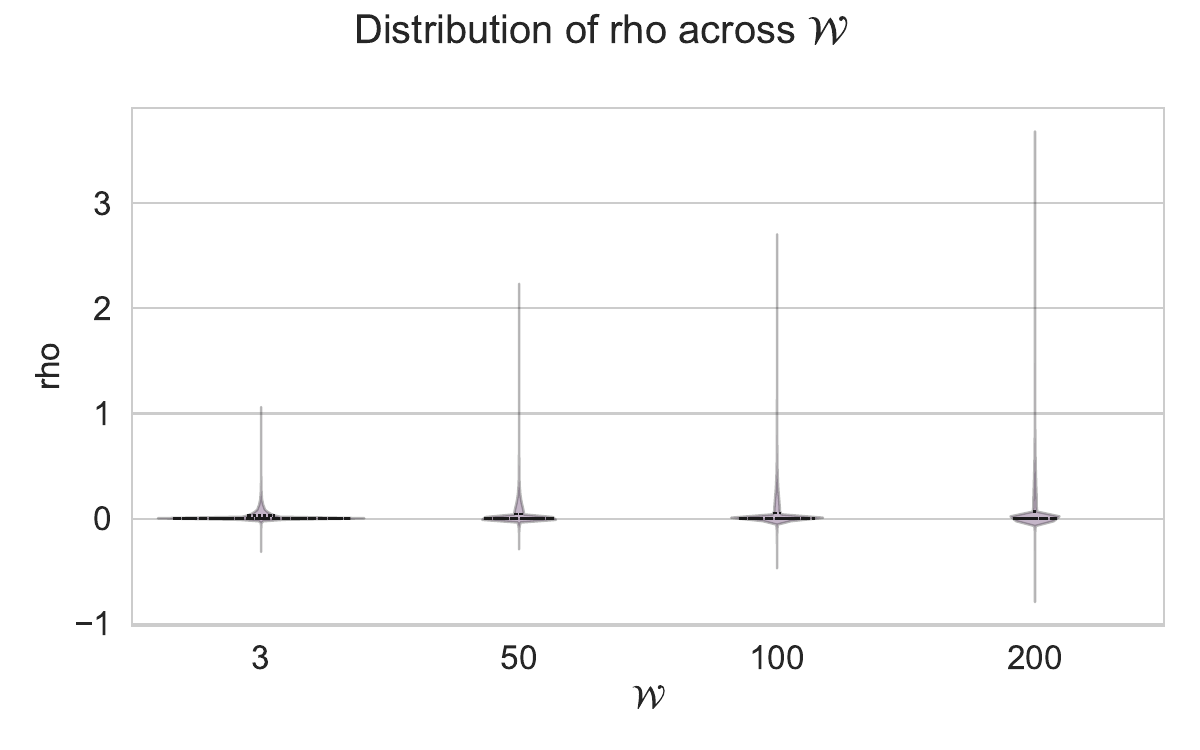}
    \end{subfigure}   
    \hspace{0.1em}
    \begin{subfigure}[t]{0.32\linewidth}
        \includegraphics[width=\linewidth]{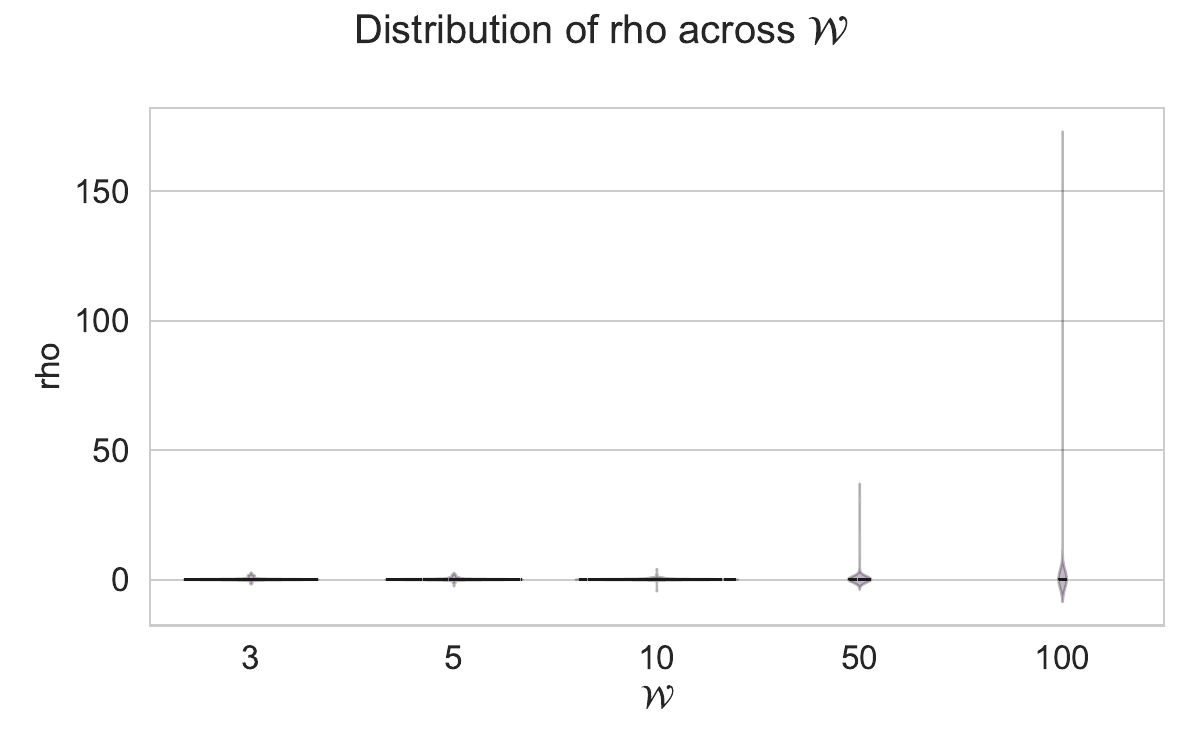}
    \end{subfigure}   
    \hfill
    \begin{subfigure}[t]{0.32\linewidth}
        \includegraphics[width=\linewidth]{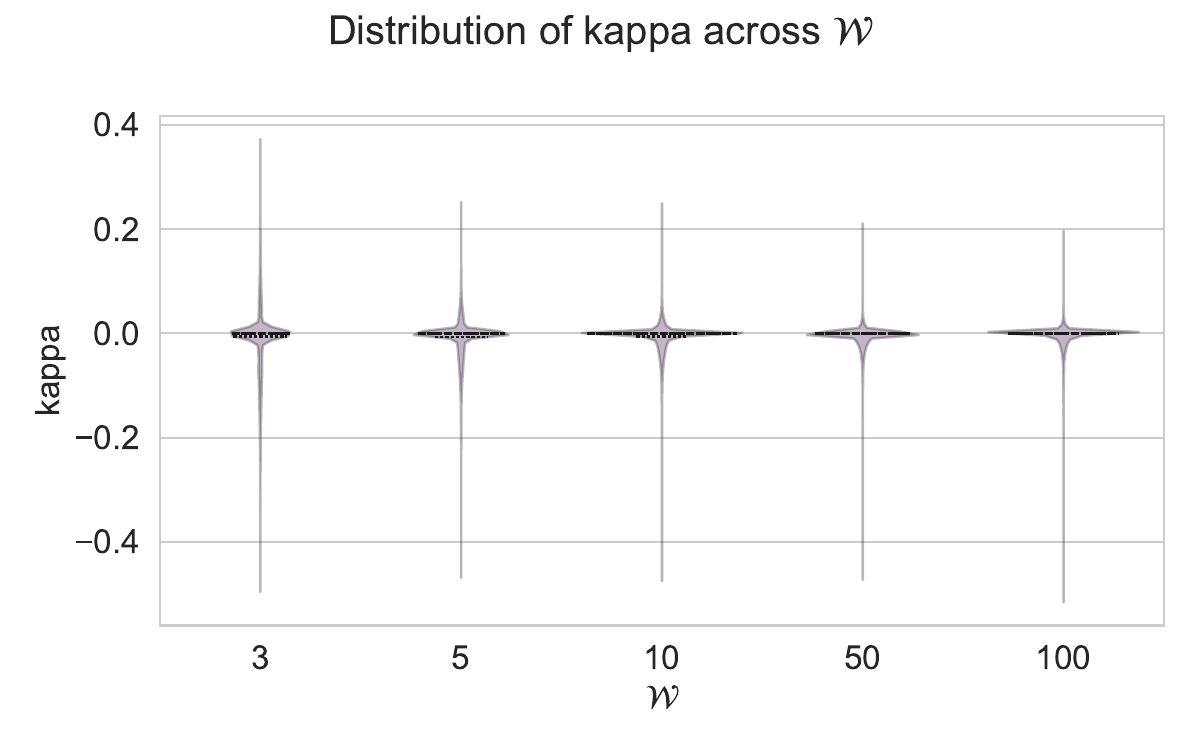}
    \end{subfigure}   
    \hspace{0.1em}
    \begin{subfigure}[t]{0.32\linewidth}
        \includegraphics[width=\linewidth]{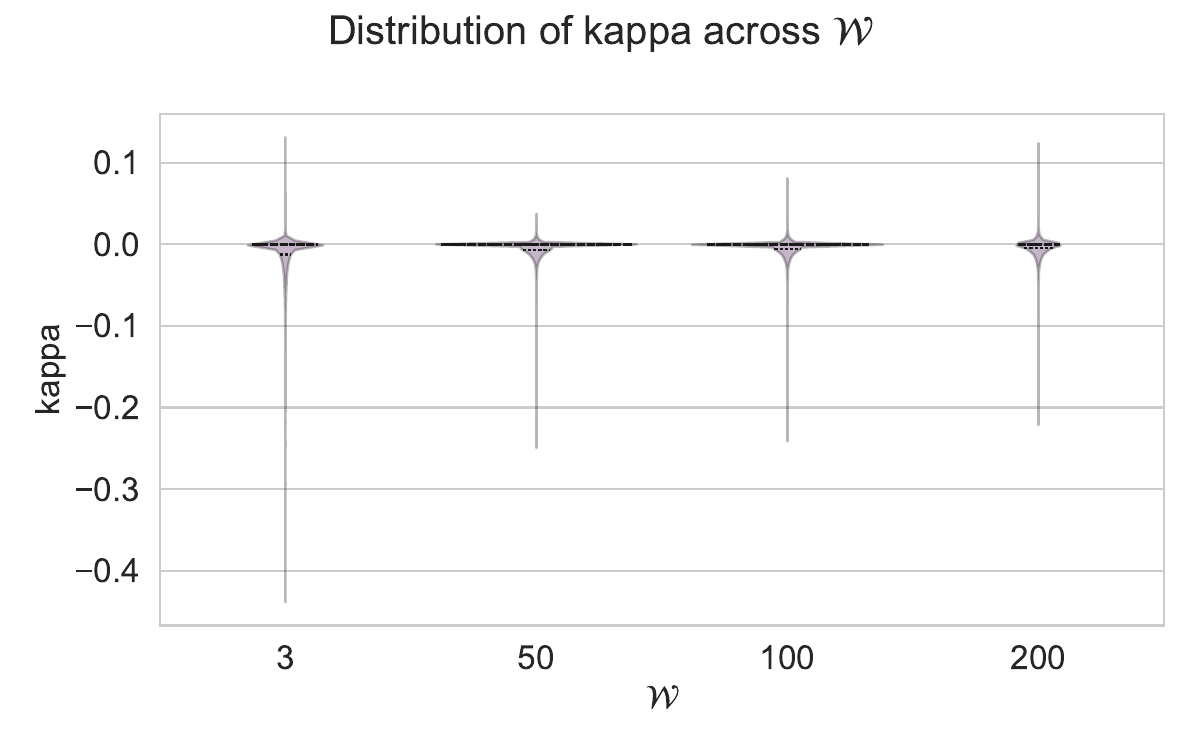}
    \end{subfigure}   
    \hspace{0.1em}
    \begin{subfigure}[t]{0.32\linewidth}
        \includegraphics[width=\linewidth]{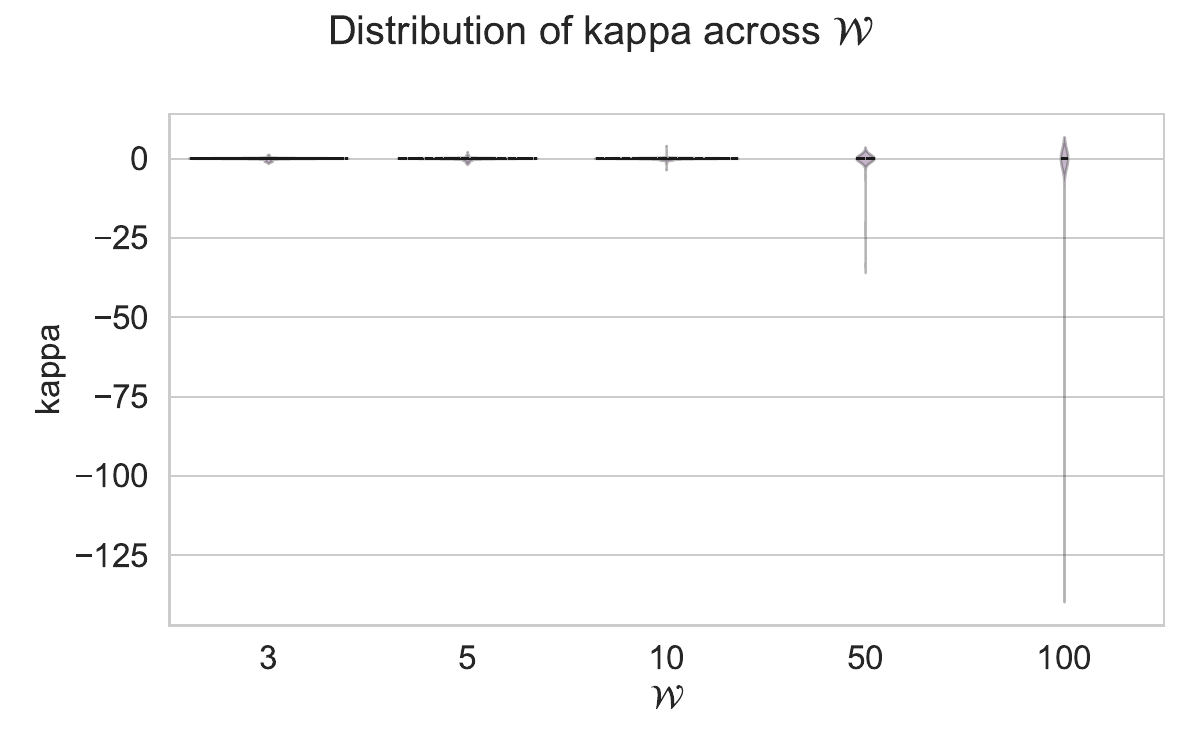}
    \end{subfigure}   
    \hfill 
    \begin{subfigure}[t]{0.32\linewidth}
        \includegraphics[width=\linewidth]{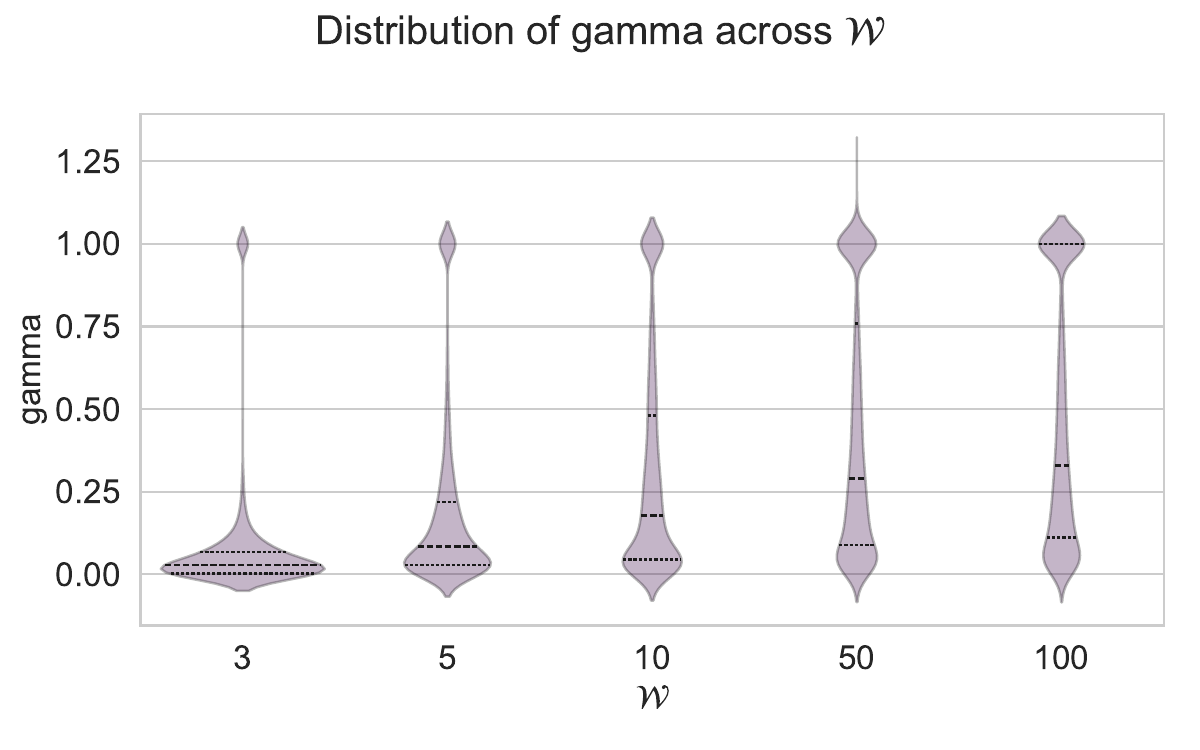}
    \end{subfigure}   
    \hspace{0.1em}
    \begin{subfigure}[t]{0.32\linewidth}
        \includegraphics[width=\linewidth]{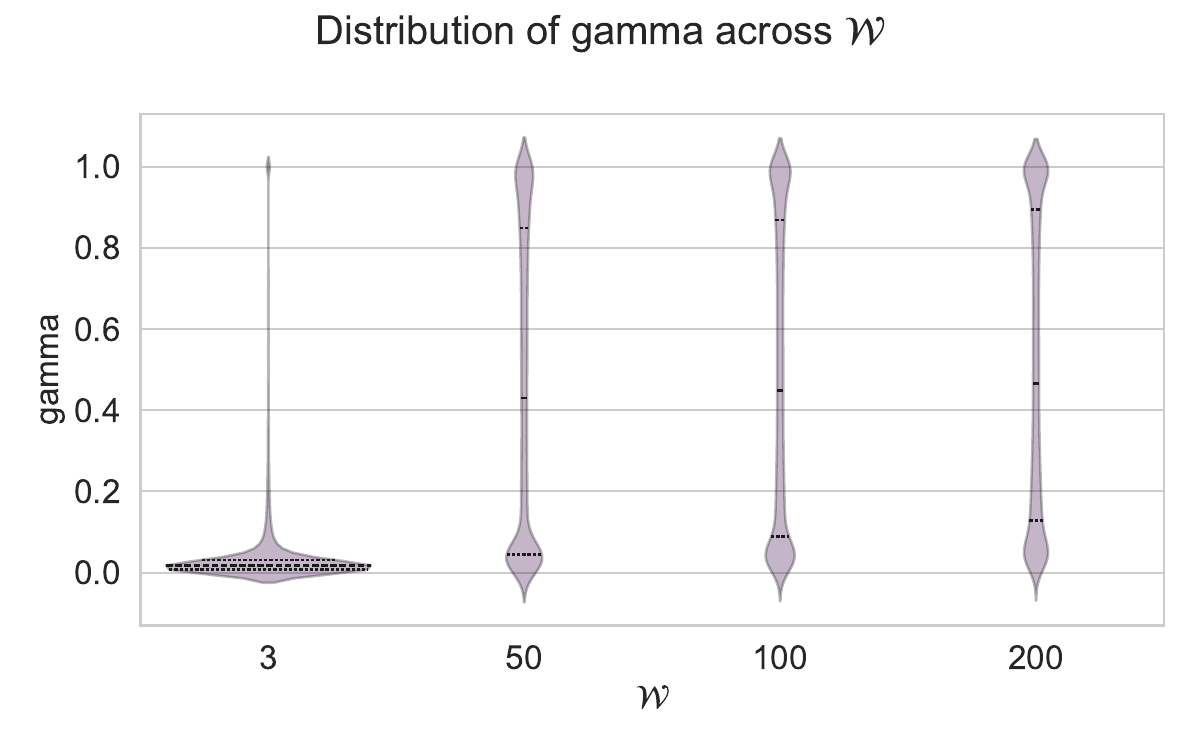}
    \end{subfigure}   
    \hspace{0.1em}
    \begin{subfigure}[t]{0.32\linewidth}
        \includegraphics[width=\linewidth]{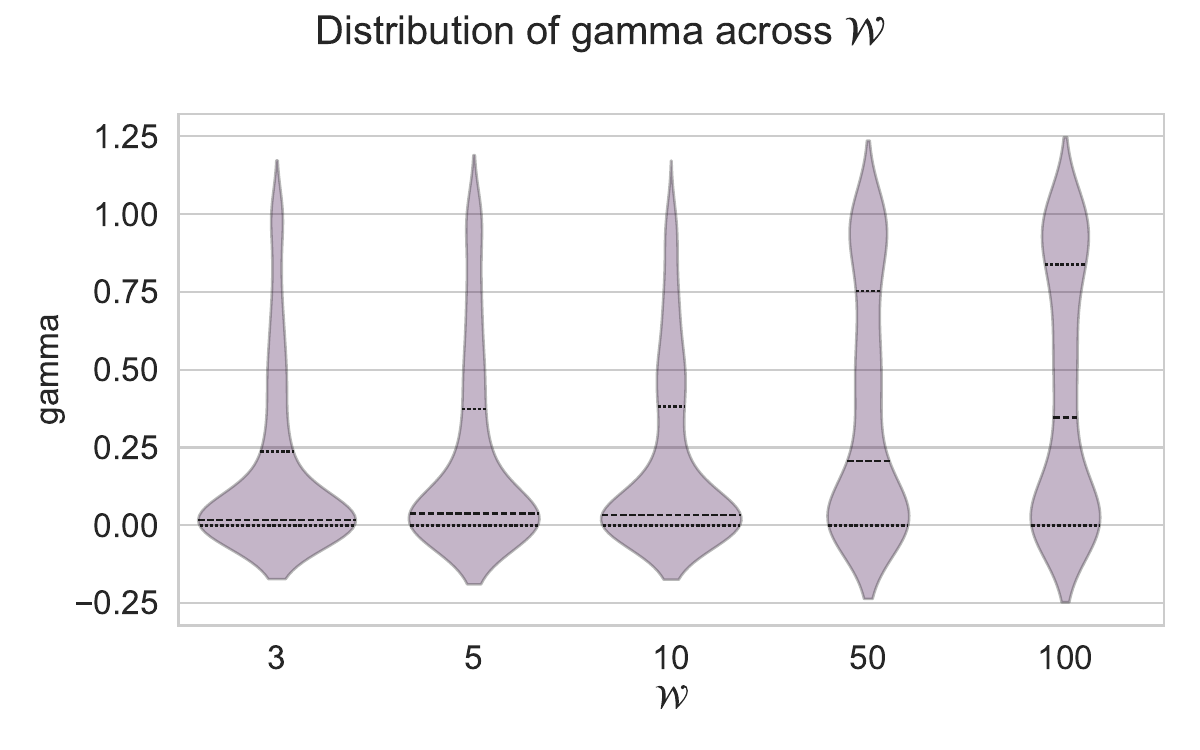}
    \end{subfigure}   
    \caption{
    {Frequency function $g$ and the effect of Learn Window Size $\mathcal{W}$ on learned parameters for datasets \texttt{ICEWS14}, \texttt{GDELT}, and \texttt{WIKI} (left to right). Note that the y-axis varies across datasets.}
    }
    \label{fig:window_g}
    \vspace{-16pt}
\end{figure*}

{Figures~\ref{fig:window_f} and~\ref{fig:window_g} illustrate the impact of the Learn Window Size $\mathcal{W}$ on the distribution of learned parameters across the datasets \texttt{ICEWS14}, \texttt{GDELT}, and \texttt{WIKI}.}

{Recall that the recency function $f$ is parameterized by $\lambda$, $\alpha$, $\phi$, which respectively control the exponential rate of decay, the starting value at $\Delta=1$, and lower bound of the curve, while the frequency function $g$ is parameterized by $\rho$, $\kappa$, and $\gamma$, controlling its slope, shift, and value clipping. Each rule learns its own set of these parameters.}

{Figure~\ref{fig:window_f} summarizes the effect of~$\mathcal{W}$ on the learned recency curves, showing for each dataset the mean curve together with its 10th and 90th percentile variation bands for three representative window sizes.
The figure also includes violin plots depicting, across rules, the distributions of $\lambda$, $\alpha$, and $\phi$ for different $\mathcal{W}$.}

{For the yearly dataset \texttt{WIKI}, the learned $\alpha$ values are consistently larger than in the other datasets across all window sizes, indicating that rules assign comparatively high confidence to events observed at~$\Delta = 1$. The values of~$\lambda$ are also generally higher, leading to steeper decay curves.
This behavior stands in contrast to \texttt{GDELT} and \texttt{ICEWS14}, whose learned $\alpha$ values are substantially smaller, indicating lower confidence in general.
Both of these datasets have much finer temporal granularity (15-minute and daily), so events are observed at a much higher density.
In such settings, a single additional day or timestep carries less significance, and the model correspondingly learns shallower decay.}

{Despite differences in scale, \texttt{GDELT} and \texttt{ICEWS14} exhibit similar trends as $\mathcal{W}$ increases.
Larger windows tend to reduce both $\alpha$ and $\phi$, while the smallest~$\lambda$ values appear at intermediate window sizes.
The variance of~$\lambda$ grows with~$\mathcal{W}$, suggesting that broader learning windows allow the rules to specialize more strongly.
Some rules learn sharply decaying recency curves, whereas others maintain relevance more gradually over longer time spans, leading to a diverse collection of temporal behaviors.}

{Figure~\ref{fig:window_g} presents the corresponding analysis for the frequency function.
For \texttt{ICEWS14} and \texttt{GDELT}, increasing~$\mathcal{W}$ leads to frequency curves that become both steeper and higher.
An interpretation is that with more historical context available, \ruc places greater weight on frequency and learns to differentiate reliable, frequently recurring patterns from noisy or isolated events.
At small~$\mathcal{W}$, by contrast, frequency becomes less informative, which is reflected in flatter curves.
At the extreme case of \texttt{ICEWS14} with $\mathcal{W}=3$, some rules even learn significantly negative~$\rho$, indicating situations in which additional occurrences reduce the score, whic is likely an effect of noise or event sparsity in very short windows. Such behavior disappears at larger windows and is less occurent for the other datasets.
\texttt{WIKI} shows a different pattern. Its frequency curves remain comparatively flat for all window sizes, and the distributions of $\rho$ and $\kappa$ show smaller variation.}

{Taken together, these observations show that \ruc does not converge to a single temporal pattern.
Instead, it learns a diverse set of curves, with substantial variation in parameters such as $\lambda$, $\rho$, and $\kappa$ across rules.
\texttt{GDELT} and \texttt{ICEWS14} exhibit similar parameter behaviors across window sizes, whereas \texttt{WIKI} behaves fundamentally differently due to its yearly resolution and high direct recurrence.}

{Importantly, the results provide no evidence that recency or frequency parameters are transferable across datasets with different temporal granularities.
Rather, the fitted parameters adapt strongly to the underlying time scale: coarser datasets learn sharper recency effects and flatter frequency responses, while finer-grained datasets support a wide spectrum of decay rates and frequency sensitivities.
The choice of window size interacts closely with temporal resolution, and the learned parameters remain dataset-specific rather than universal.}

\newpage

\subsection{{Rule Type Distributions}}\label{ap:rule_type}
\begin{figure*}
\vspace{-.5em}
    \centering
    \begin{minipage}[t]{\linewidth}
        \centering
        \includegraphics[trim={.4cm .5cm 0.3cm .18cm},clip,width=\linewidth]{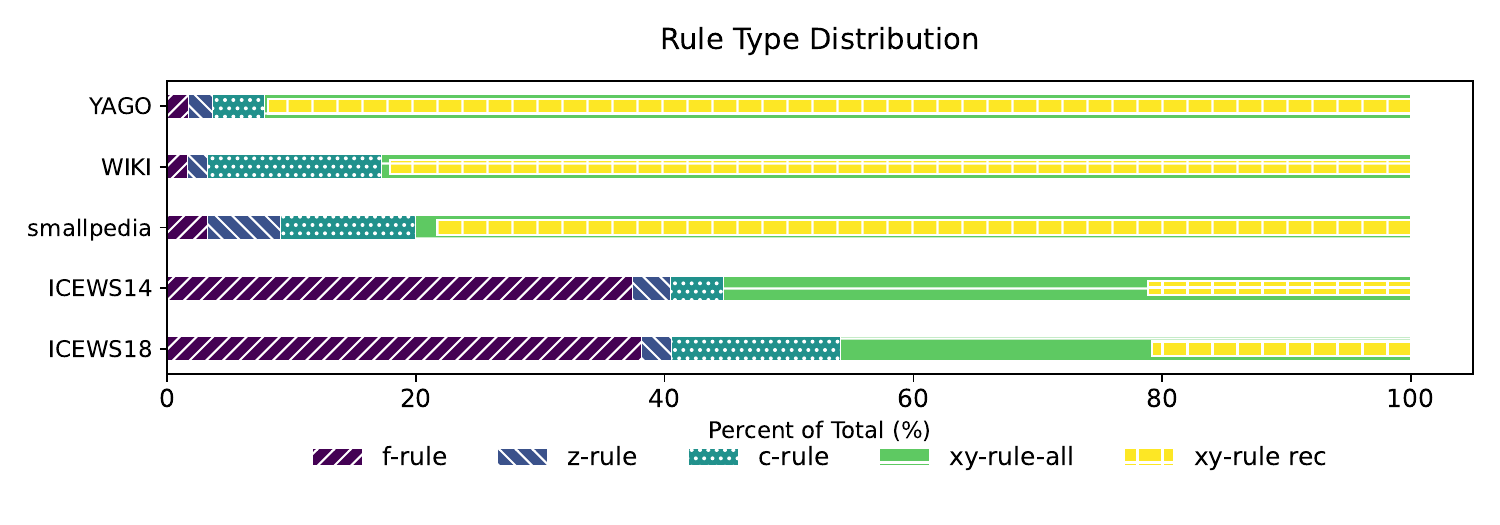} \\
        \includegraphics[trim={.4cm .5cm 0.3cm .18cm},clip,width=\linewidth]{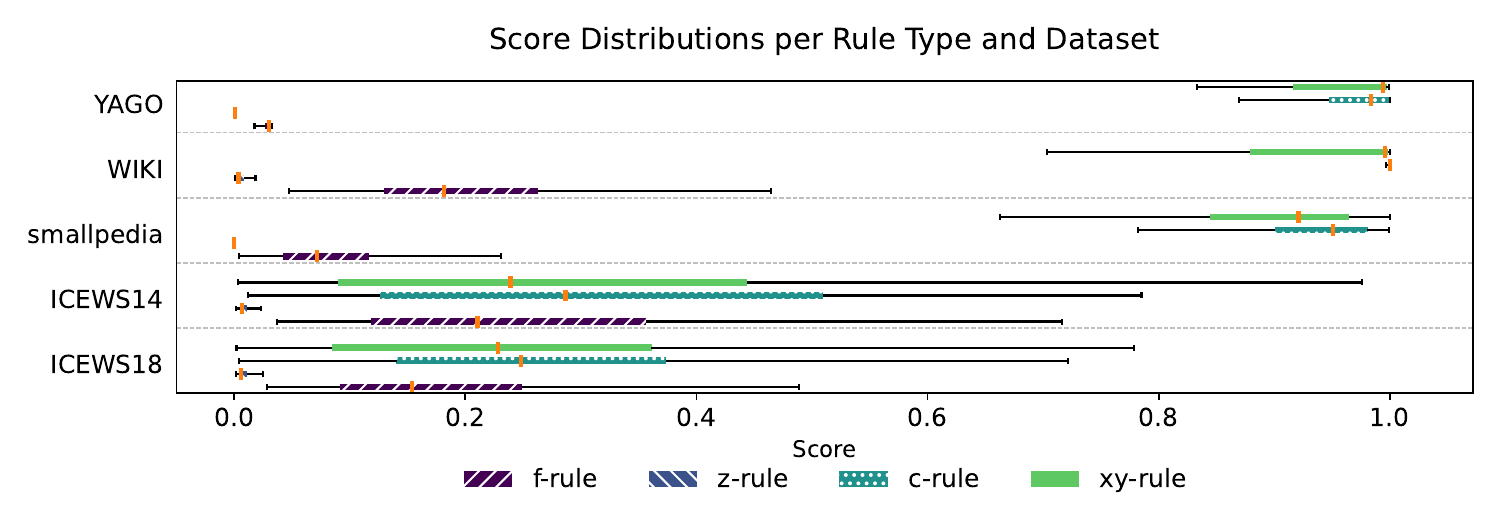}
        \vspace{-2em}
        \caption{{Distributions for the four rule types (top), and assigned scores for each rule type (bottom) for the highest predicted node and highest predicting rule for each test query.}
        }
        \label{fig:ruledist}
    \end{minipage}%
\end{figure*}
{Figure~\ref{fig:ruledist} shows the distributions of rule types (top) and the corresponding scores (bottom) for the highest-predicted node and its top-predicting rule across all test queries in the small and medium sized datasets.}

{The figure illustrates that the contributions of rule types vary significantly across datasets. For datasets based on Wikidata (\texttt{WIKI}, \texttt{smallpedia}) and YAGO (\texttt{YAGO}), which exhibit a high degree of recurrency~\cite{gastinger2024baselines}, xy-recurrency rules have the highest contributions. This is consistent with the ablation results in Table~2(a) (main paper), where \ruc on \texttt{WIKI} achieves its best MRR using only recurrency rules.
In contrast, for \texttt{ICEWS14} and \texttt{ICEWS18}, recurrency rules account for only around $20\%$ of the contribution. Large contributions come also from non-recurrent xy-rules and f-rules, again reflecting the trends observed in Table~2(a) (main paper). Across all datasets, z-rules contribute minimally ($<5\%$).}

{The score distributions in Figure~\ref{fig:ruledist} (bottom) show that xy- and c-rules achieve higher average scores for Wikidata- and YAGO-based datasets, whereas their average scores are substantially lower in \texttt{ICEWS14} and \texttt{ICEWS18}. The z-rules consistently have low median scores, as they primarily capture general distributions. The c-rules show the highest median scores for most datasets, which aligns with their specialized nature and ability to provide targeted, high-confidence predictions.}

{Overall, these distributions confirm our intuition: xy-rules contribute the most across datasets, but other rule types also play an important role in achieving high performance.}

\subsection{Runtime}\label{ap:runtimes}
Table~\ref{tab:runtime} reports runtimes in seconds for \ruc on all datasets. We report the total runtime (including dataset loading and all necessary steps), as well as the runtimes of the individual components described in the main paper, i.e., creating the Examples $E$ for each rule, learning the parameters for each rule, rule application and aggregation, and evaluation.

All runtimes were measured on an AMD EPYC 9474F 48-Core Processor running AlmaLinux 9.5. \ruc does not require GPU acceleration and thus no GPUs were used in these experiments. Parallelization during rule application was implemented via Ray using up to 20 parallel workers.

For reference, Table~\ref{tab:tkgtesttime} reports runtimes from~\cite{gastinger2024tgb} on the datasets included in their study. In addition, Table~\ref{tab:cogtesttime} reports runtimes for CognTKE on the datasets where we evaluated it, and Table~\ref{tab:tempvalidtesttime} and Table~\ref{tab:dimnettesttime} report runtimes for TempValid and DiMNet, respectively. Since these results were obtained on different hardware\footnote{Experiments in Table~\ref{tab:tkgtesttime} were conducted on Nvidia A100, V100, V100SXM2, and RTX8000 GPUs with 4 CPU nodes (from AMD Rome, Milan, or Intel Skylake) per experiment, using up to 1056 GB of RAM, and experiments in Table~\ref{tab:cogtesttime}, Table~\ref{tab:tempvalidtesttime}, and Table~\ref{tab:dimnettesttime} were conducted on an NVIDIA RTX A6000 GPU with 48~GB VRAM with up to 16 CPU nodes, using up to 500~GB of RAM}, they are not directly comparable, but they provide a useful point of reference.

Table~\ref{tab:tkgtesttime} shows that \ruc achieves substantially lower total runtimes on \texttt{smallpedia} and \texttt{polecat} than the runtimes reported for related work, despite running without GPU acceleration. On \texttt{icews}, \ruc attains comparable runtimes, while on \texttt{wikidata} it is the only method aside from the simple EdgeBank heuristic that completes successfully.
Table~\ref{tab:cogtesttime} further reports CognTKE runtimes on the datasets where we evaluated it. While these results were obtained on different hardware and are therefore not directly comparable, they provide an indication that \ruc is highly efficient in practice, achieving low runtimes using only CPU resources.

\begin{table*}
\caption{\ruc runtimes for all datasets in seconds.}
\footnotesize
\centering
\resizebox{\textwidth}{!}{
\begin{tabular}{l|r|r|r|r|r|r|r|r|r}
\toprule
& {\texttt{GDELT}} & {\texttt{YAGO}} & {\texttt{WIKI}} & {\texttt{ICEWS14}} & {\texttt{ICEWS18}} & {\texttt{smallp.}} & {\texttt{polecat}} & {\texttt{icews}} & {\texttt{wikidata}} \\
\midrule
\# rules      &  134,574     & 8,140 & 350  & 53,553   & 1,010,307 & 1,221            & 1,024         &    167,670        & 12,050        \\
\midrule
creation of examples {[}s{]} & 12,244 & 9   & 10  & 111 & 242   & 7   & 1,464  & 129,139 & 280   \\
parameter learning {[}s{]}   & 5,670  & 11  & 0.4 & 366 & 7,743  & 1   & 594   & 2,221   & 18    \\
rule application {[}s{]}     & 3,704  & 125 & 106 & 279 & 12,593 & 194 & 10,957 & 137,574 & 39,911 \\
evaluation {[}s{]}           & 253   & 45  & 18  & 64  & 510   & 69  & 620   & 6335   & 2,637 \\
total time {[}s{]}           & 21,926 & 191 & 143 & 825 & 21,149 & 280 & 13,681 & 276,269 & 43,029 \\
\bottomrule
\end{tabular}
}
\label{tab:runtime}
\end{table*}

\begin{table*}
\caption{Inference time as well as total train and validation times as reported in~\cite{gastinger2024tgb} in seconds. }
\footnotesize
\centering
\footnotesize
  \begin{tabular}{l | rr | rr | rr | rr}
  \toprule
  \multirow{2}{*}{Method} & \multicolumn{2}{c|}{{\texttt{smallpedia}}} &  \multicolumn{2}{c|}{{\texttt{polecat}}}  & \multicolumn{2}{c|}{{\texttt{icews}}} & \multicolumn{2}{c}{{\texttt{wikidata}}} \\
    & Test  & Total                  & Test  & Total                    & Test  & Total            & Test  & Total   \\
  \midrule
    $\text{EdgeBank}_{\text{tw}}$               & 2,935 & 5,810 & 46,629 & 94,475 & 311,278 & 600,929 & 5,445 & 8,875\\
     $\text{RecurrencyBaseline}$  & 310 & 9,895& 3,392 & 80,378   & 3,928 & 148,710 & - & -  \\
    RE-GCN      &       {165}  & {3,895}  &    {1,766} & {45,877}  &{6,848}  & {114,370}  & - & -  \\ 
    CEN            &         {331}  & {14,493}  &   {2,726}  & {77,953}  &{8,999}   &{202,477}  & - & -  \\ 
    TLogic     &         {331}  & {803} & {75,654}  & {138,636}  &{60,413}   &{128,391}   & - & -  \\ 
  \bottomrule
  \end{tabular}
\label{tab:tkgtesttime}
\end{table*}

\begin{table*}
\caption{Test time as well as total train and validation times for CognTKE in seconds. }
\footnotesize
\centering
\footnotesize
\resizebox{0.8\textwidth}{!}{
  \begin{tabular}{l | rr | rr |rr | rr | rr |rr |rr}
  \toprule
 & \multicolumn{2}{c|}{\texttt{GDELT}} &   \multicolumn{2}{c|}{\texttt{YAGO}} & \multicolumn{2}{c|}{\texttt{ICEWS14}}  & \multicolumn{2}{c|}{\texttt{smallp.}} &  \multicolumn{2}{c|}{{\texttt{polecat}}}  & \multicolumn{2}{c|}{{\texttt{icews}}} & \multicolumn{2}{c}{{\texttt{wikidata}}}\\
    & Test  & Total                  & Test  & Total         & Test  & Total  & Test  & Total      & Test  & Total      & Test  & Total      & Test  & Total              \\
  \midrule 
    CognTKE             & OM & OM  & 78 & 8,287 &  36 & 6,080 & 711 & 63,300 & 7,548 & 417,600 & OM & OM & OM & OM \\
  \bottomrule
  \end{tabular}
}
\label{tab:cogtesttime}
\end{table*}

\begin{table*}
\caption{Runtimes for TempValid in seconds.}
\footnotesize
\centering
\footnotesize
\resizebox{0.9\textwidth}{!}{
  \begin{tabular}{l | r | r |r |r | r | r| r  |r |r}
  \toprule
 & \multicolumn{1}{c|}{\texttt{GDELT}} &   \multicolumn{1}{c|}{\texttt{YAGO}} &    \multicolumn{1}{c|}{\texttt{WIKI}} &\multicolumn{1}{c|}{\texttt{ICEWS14}} &\multicolumn{1}{c|}{\texttt{ICEWS18}}  & \multicolumn{1}{c|}{\texttt{smallp.}} &  \multicolumn{1}{c|}{{\texttt{polecat}}}  & \multicolumn{1}{c|}{{\texttt{icews}}} & \multicolumn{1}{c}{{\texttt{wikidata}}}\\
  \midrule 
    \# rules           &   100,187 &   46   &  574  &  28,661 & 42,531 & 3,362  & 2,614  & 101,244   & 6,544 \\
\midrule
rule learning {[}s{]}  &    8,364 &  11     &  92  & 224  & 743 & 3,362  & 711  & 80,117   & 363,813  \\ 
feature generation {[}s{]}  &   OM &   18,651    & OT   & 22,702   & OT & OM  & error  & error   & error \\
tempvalid training and testing {[}s{]} &   OM &  41    & OT   & 4,275   & OT & OM  & error  & error   & error \\
total time {[}s{]}&   OM &  18,703     &  OT  & 27,201   & OT & OM  & error  & error   & error \\
  \bottomrule
  \end{tabular}
  }
\label{tab:tempvalidtesttime}
\end{table*}

\begin{table*}
\caption{Runtimes for DiMNet in seconds.} 
\footnotesize
\centering
\footnotesize
\resizebox{0.7\textwidth}{!}{
  \begin{tabular}{l | r | r |r |r | r | r| r  |r |r}
  \toprule
 & \multicolumn{1}{c|}{\texttt{GDELT}} &   \multicolumn{1}{c|}{\texttt{YAGO}} &    \multicolumn{1}{c|}{\texttt{WIKI}} &\multicolumn{1}{c|}{\texttt{ICEWS14}} &\multicolumn{1}{c|}{\texttt{ICEWS18}}  & \multicolumn{1}{c|}{\texttt{smallp.}} &  \multicolumn{1}{c|}{{\texttt{polecat}}}  & \multicolumn{1}{c|}{{\texttt{icews}}} & \multicolumn{1}{c}{{\texttt{wikidata}}}\\
    & Total                  & Total         & Total  & Total      & Total      & Total      & Total & Total  & Total              \\
  \midrule 
    DiMNet           & OM   &  914    & 2,062   & 1,300  & 7,033  & 2,045 & OM  & OM   & OM \\

  \bottomrule
  \end{tabular}
}
\label{tab:dimnettesttime}
\end{table*}

\subsection{{Hits@1 and Hits@3 Results}}\label{ap:hits}
{Tables~\ref{tab:_h1_1} and~\ref{tab:results_h1_2} report the results including Hits@1 and Hits@3 values.}

{The Hits@1 results do not provide a different impression from the MRR results presented in Table~1 (main paper). Whenever \ruc ranks first, second, or third in MRR among the compared models, it has the same position with respect to the Hits@1 score. The only exceptions are \texttt{YAGO}, where \ruc moves from first to second place for Hits@1, and \texttt{ICEWS18}, where it moves from third to second place.
The ranking of the Hits@3 values is also consistently within the top three positions among related work.}

{For the TGB2.0 datasets in Table~\ref{tab:results_h1_2}, no Hits@1 and Hits@3 results were reported for related work. We thus compare our results only against CognTKE. On one, CognTKE achieves higher Hits@1 and Hits@3 scores, on another \ruc achieves the highest results, and on two datasets, CognTKE did not produce results due to OM errors.}

\begin{table*}
\caption{{Model comparison including Hits@1 and Hits@3 on five datasets. OM means Out Of Memory (40~GB GPU or 1056~GB RAM), OT means Out Of Time (7 days), - means that no results have been reported. Best results are shown in bold and underlined font, second-best bold, third-best underlined.} }
\footnotesize
\centering
\resizebox{\textwidth}{!}{
\begin{tabular}{l|rrrr|rrrr|rrrr|rrrr|rrrr}
\toprule
& \multicolumn{4}{c|}{\texttt{GDELT}} & \multicolumn{4}{c|}{\texttt{YAGO}} & \multicolumn{4}{c|}{\texttt{WIKI}} & \multicolumn{4}{c|}{\texttt{ICEWS14}} & \multicolumn{4}{c}{\texttt{ICEWS18}}\\
\midrule
                          & MRR & H1 & H3  & H10 & MRR & H1 & H3  & H10 & MRR & H1 & H3  & H10 & MRR & H1 & H3  & H10 & MRR & H1 & H3  & H10   \\
\midrule
TRKG    & 21.5    &  13.7&  24.0  & 37.3       & 71.5      & 65.7 & 77.3    & 79.2      & 73.4       & 71.2 &  75.6  & 76.2    & 27.3       & 16.5 & 31.1   & 50.8      & 16.7       &  8.3& 18.2    & 35.4      \\
xERTE   & 18.9        &  12.7 & 21.1  & 32.0       & 87.3     & 84.2 & 90.3     & 91.2              & 74.5      & 70.3 &  78.6    & 80.1      & 40.9       &  33.0 &  45.5  & 57.1              & 29.2       & 20.9 & 33.5    & 46.3         \\
TANGO   & 19.2       &  12.2&  20.4  & 32.8       & 62.4      & 59.0 &  64.7   & 67.8               & 50.1      & 48.3 &  51.4   & 52.8      & 36.8      & 27.3 & 40.8    & 55.1      
        & 28.4    & 19.1 & 31.9  & 46.3         \\
Timetraveler    & 20.2       & \pc{14.1} & 22.2   & 31.2       & 87.7     & 84.6 & 90.9     
        & 91.2      & 78.7       & 75.2 & 82.0   & 83.1      & 40.8     & 31.9 &  45.4    
        & 57.6      & 29.1      & 21.3 &  32.5    & 43.9           \\
TiRGN   & \pc{21.7} & 13.6    & \pc{23.3} &  \pc{37.6}  & 88.0     & 84.3 & 91.4     & 92.9      & 81.7     &  77.8&   85.1         & \pb{87.1} & {44.0} &  {33.8}  & {49.0} &  \pb{63.8} & \pb{33.7}    & \pc{23.2} &  \pb{38.0}  & \pb{54.2}   \\
TempValid & OM & OM &OM & OM & 46.7 & 43.3 & 50.0 & 50.4 & OT &OT& OT&  OT & \pb{45.5} & \pc{35.4} & \pb{50.8} & \pa{64.5} & OT &OT &OT &  OT  \\
DiMNet & 21.1 & 13.3 & 22.5 & 36.5 & 72.4 & 67.1 & 75.0 & 82.7 & 71.2 & 66.4 & 74.0 &  79.7 & 41.4 & 31.4 & 46.3 &  60.8 & 33.9 & 23.2 & 38.2 &  55.4 \\
CognTKE & OM & OM &  OM & OM        & \pc{90.6} &   \pc{88.1}  &  \pa{92.9} & \pa{93.2} & \pa{83.2} & \pa{80.0}   &  \pa{86.0}        &  \pa{87.3} & \pa{46.1} &  \pa{36.5}   & 
\pa{51.1} & \pa{64.5} & \pa{35.2}  &   \pa{25.2}  & \pa{39.9} & \pa{54.7}     \\
TLogic  & 19.8         &  12.2 &  21.7 & 35.6       & 76.5      & 74.0& 78.9    & 79.2              & \pc{82.3}     &  \pc{78.6} & \pa{86.0} & 87.0      & 42.5       & 33.2 & 47.6    & 60.3             & 29.6        & 20.4  & 33.6   & 48.1         \\
RE-GCN  & 19.8       & 12.5 &  21.0  & 33.9       & 82.2 & 78.7 & 84.2  & 88.5      & 78.7      &  74.8 & 81.7 & 84.7      & 42.1  & 31.4 &  47.3  & {62.7} & 32.6 & 22.4 & 36.8    & \pc{52.6}   \\
CEN     & 20.4 & 13.0 & 21.8  & 35.0       & 82.7   &  78.8 &  85.2   & 89.4      & 79.3       & 75.5 &  82.4 & 84.9      & 41.8 & 31.9 &  46.6   & 60.9      & 31.5      & 21.7 & 35.4      & 50.7          \\
$\text{EdgeBank}_{\text{tw}}$    & 1.9       &  0.1 &    0.9  & 3.5        & 61.7        & 44.1 
        &    68.5 
        & 61.7      & 58.5     &  39.4&  67.3    & 84.4      & 13.5      & 3.2 &  14.1   & 34.2      & 7.2        &   1.5   & 6.3 & 17.9   \\
Rec.B ($\psi_\Delta\xi$)         & \pa{24.5} &  \pa{16.4}  &  \pa{26.8} &  \pb{39.8}  & \pa{90.9} &  \pa{89.0}   &  \pc{92.8}
        & \pc{93.0} & 81.4         & 76.9 & \pc{85.7} & \pb{87.1} & 37.4        & 29.9 &  41.2 & 51.5      & 28.7        & 20.8 & 32.3   & 43.6         \\
\midrule
\ruc    & \pb{23.8} &  \pb{15.4}  & \pb{26.3} &  \pa{40.3}  & \pa{90.9} &  \pb{88.8}   & \pa{92.9} & \pa{93.2} & \pb{82.7}     &    \pb{79.4}      & \pc{85.7} & 86.6      & \pb{45.0} &  \pb{36.0}  & \pc{49.8} &  62.0      & \pc{32.8}     & \pb{23.5} & \pc{36.9} & 51.0       \\
\bottomrule
\end{tabular}
}
\label{tab:_h1_1}

\end{table*}

\begin{table*}
\caption{{Model comparison including Hits@1 and Hits@3 on the TGB2.0 benchmark datasets. OM means Out Of Memory (40~GB GPU or 1056~GB RAM), OT means Out Of Time (7 days), - means that no results have been reported. Best results are shown in bold and underlined font, second-best bold, third-best underlined.} }
\centering
\resizebox{.9\textwidth}{!}{
\begin{tabular}{l|rrrr|rrrr|rrrr|rrrr}
\toprule
& \multicolumn{4}{c|}{\texttt{smallp}} & \multicolumn{4}{c|}{\texttt{polecat}} & \multicolumn{4}{c|}{\texttt{icews}} & \multicolumn{4}{c}{\texttt{wikidata}} \\
\midrule
                          & MRR & H1 & H3  & H10 & MRR & H1 & H3  & H10 & MRR & H1 & H3  & H10 & MRR & H1 & H3  & H10   \\
\midrule
TempValid &  OM & OM& OM & OM & error & error & error & error& error& error & error & error& error &   error & error &error \\
DiMNet &   54.3 & \pb{48.6} & \pc{57.4} &  64.7 & OM & OM & OM & OM & OM & OM& OM & OM & OM & OM & OM &OM\\
CognTKE          & 53.4      &  \pc{44.0} & \pb{60.2} & 70.1      & \pa{28.7} & \pa{20.1} & \pa{32.7} &  \pa{45.5} & OM  &OM  &    OM    & OM       & OM   & OM &     OM   & OM       \\
TLogic         & 59.5     & - & -    & \pc{70.7} & \pc{22.8} & - & -  &  \pc{37.8} & 18.6   & - & -      & 30.1      & OT       & OT& OT & OT       \\
RE-GCN                       & 59.4     & - & -    & 68.7      & 17.5   &- & -      & 29.2      & 18.2      & - & -   & \pb{33.1} & OM     &  OM &    OM  & OM       \\
CEN            & \pb{61.2} & - & -   & 70.5      & 18.4    & - & -    & 32.3      & \pc{18.7} &  - & -  & \pa{33.4} & OM      & OM  &  OM   & OM       \\
$\text{EdgeBank}_{\text{tw}}$   & 35.3     &- & -  & 56.6      & 5.6   &- & -      & 11.9      & 2.0     &- & -    & 5.8       & \pb{53.5} & - & - &  \pb{59.6} \\
Rec.B ($\psi_\Delta\xi$)         & \pc{60.5} & - &- & \pb{71.6} & 19.8  &- & -     & 31.7      & \pb{21.1} & - & -& \pc{32.4} & OT     & OT & OT   & OT       \\
\midrule
\ruc     & \pa{64.4} & \pa{59.5} & \pa{68.7} & \pa{71.7} & \pb{25.6} & \pb{17.8} & \pb{29.1} & \pb{40.8} & \pa{21.4} &\pa{15.7}  & \pa{24.1} &  32.1      & \pa{60.9}& \pa{59.8}  & \pa{61.6}  & \pa{62.8} \\
\bottomrule
\end{tabular}
}
\label{tab:results_h1_2}
\vskip -0.15in
\end{table*}

\subsection{Variance Across Repetitions}\label{ap:rep}
Table~\ref{tab:variance} reports test results across five repetitions for two selected datasets. We observe that the variance is less than $0.002$ for all test metrics and datasets, which is small considering that the MRR and Hits@10 range from $0\%$ to $100\%$. 
Compared to embedding-based methods, which may introduce randomness through factors such as random initialization or noise injection, our approach has fewer potential sources of variability: (i) sampling candidates for the examples of $c$-rules, (ii) computing the parameters that optimize the temporal confidence functions, and (iii) assigning random order to tied ranks during evaluation.

\begin{table*}
\caption{Test results on \texttt{WIKI} and \texttt{ICEWS14} (5 runs, mean, variance).}
\footnotesize
\centering
\begin{tabular}{l|r|r|r|r}
\toprule
         & \multicolumn{2}{c|}{\texttt{WIKI}} & \multicolumn{2}{c}{\texttt{ICEWS14}} \\
         & MRR         & H10        &     MRR         & H10         \\
 \midrule
         & 82.6809     & 86.5592    & 44.9916      & 62.0201      \\
         & 82.6615     & 86.5568    & 44.9828      & 62.0404      \\
         & 82.5704     & 86.5647    & 45.0014      & 62.1083      \\
         & 82.6181     & 86.5624    & 44.9864      & 62.0336      \\
         & 82.5992     & 86.5584    & 44.9685      & 62.0201      \\
\midrule
Mean     & 82.6260     & 86.5603    & 44.9862      & 62.0445      \\
Variance & 0.002038    & 0.000010   & 0.000146     & 0.001348     \\
\bottomrule
\end{tabular}
\label{tab:variance}
\end{table*}

\subsection{Significance Tests}\label{ap:sig}
We conduct a Chi-squared test to assess whether the observed differences in Hits@10 scores between methods are statistically significant. Since Hits@10 is a binary metric (a prediction is either in the top 10 or not), we treat the predictions as categorical outcomes: \textit{hit} and \textit{miss}. For a dataset, we have $2 \times N$ samples (where $N$ is the number of test quadruples), accounting for inverse relations.

For example, if a model achieves a Hits@10 score of 0.6 on a test set with 100 quadruples, this corresponds to $0.6 \times 200 = 120$ \textit{hit} predictions and 80 \textit{miss} ones.

We compare \ruc with the best-performing state of the art method regarding Hits@10 under the following null hypothesis:

\begin{quote}
\textbf{H\textsubscript{0}}: Prediction correctness (i.e., \textit{hit} vs. \textit{miss}) is independent of the method used.
\end{quote}

For examples, for the \texttt{GDELT} dataset, the Chi-squared test indicates a significant association between prediction correctness and method ($\chi^2 = 31.8$, $\mathit{df} = 1$, $p < 0.001$), suggesting that the improvement in Hits@10 by \ruc over the baseline is statistically significant.
Based on $p < 0.001$, in total, for the 4 datasets, where \ruc has higher Hits@10 scores than the others, for 2 of them the Chi-squared tests suggests a statistically significant improvement. For the 5 datasets, where \ruc has lower Hits@10 scores than the other methods, for 4 of them the Chi-squared test suggests a statistically significant difference.

\begin{table*}[]
\caption{Chi-squared test for all datasets. We test for $(df=1$, $p<0.001)$. and compare \ruc with the best performing state of the art method regarding Hits@10.}
\footnotesize
\centering
\resizebox{1.\textwidth}{!}{
\begin{tabular}{l|r|r|r|r|r}
\toprule
Dataset    & \ruc H10 & Best Baseline H10 & Best Baseline    & $\chi^2$ & Significant? \\
\midrule
\multicolumn{6}{l}{Group 1: \ruc best method}     \\
\midrule
\texttt{GDELT}      & 40.3                   & 39.8             & Rec. B.          & 31.78             & yes                              \\
\texttt{YAGO}       & 93.2                   & 93.2              & CognTKE          & 0             & no                               \\
\texttt{smallpedia} & 71.7                   & 71.6             & Rec. B.          & 0.40           & no                               \\

\texttt{wikidata}   & 62.8                   & 59.6             & Edgebank         & 6204.43           & yes                              \\
\midrule
\midrule
\multicolumn{6}{l}{Group 2: \ruc not best method}      \\
\midrule
\texttt{WIKI}       & 86.6                   & 87.3            & CognTKE & 27.25              & yes                              \\
\texttt{ICEWS14}    & 62.0                    & 64.5             & CognTKE           & 19.81              & no                               \\
\texttt{ICEWS18}    & 51.0                    & 54.7             & CognTKE           & 272.19         & yes                              \\
\texttt{icews}      & 32.1                   & 33.4             & CEN              & 1784.58              & yes                              \\
\texttt{polecat}    & 40.8                   & 45.4             & CognTKE           & 2297.88           & yes                              \\
\bottomrule
\end{tabular}
}
\end{table*}

\subsection{Memory Limits}\label{ap:memory}
The upper memory limits set in SLURM for each dataset (using 20 parallel processes) are as follows:
\begin{itemize}
\setlength{\itemsep}{0em}  
\setlength{\itemindent}{0pt}
\setlength{\leftmargini}{0.8em} 
    \item {Small datasets} (\texttt{WIKI}, \texttt{ICEWS14}, \texttt{YAGO}, \texttt{tkgl-smallpedia}): {20 GB}
    \item \texttt{polecat}: {160 GB}
    \item \texttt{ICEWS18}: {300 GB}
    \item \texttt{gdelt}: {350 GB}
    \item \texttt{wikidata}: {200 GB}
    \item \texttt{tkgl-icews}: {500 GB}
\end{itemize}
Please note that these are upper limits, not actual memory usage.  
Additionally, reducing the number of parallel processes during the application phase will decrease memory consumption at the cost of increased runtime.

\section{Details on the explanation tool}\label{ap:explanation}
In the following, we provide additional details on the explanation tool introduced in Section~6 in the main paper.
The code for the explanation tool is also published in the provided repository, along with notebooks demonstrating its use.

\paragraph{Pre-Analysis Features}
Before generating explanations, the tool helps gain an overview of datasets and models. It provides (i) {dataset statistics} like relation distributions, and number of triples over time, computes (ii) a {fine-grained evaluation} with metrics like MRR per relation and timestep, and conducts (iii) a {ranking comparison} that identifies quadruples where two methods differ significantly. 

For example, the ranking comparison showed that on the dataset \texttt{ICEWS14}, \ruc consistently outperforms RE-GCN on the relation \emph{Consult}, suggesting that simple temporal rules capture this relation comparatively well.  

\paragraph{Input and Workflow} 
The tool takes as input (i) a rule set (learned or user-defined), (ii) an optional configuration (e.g., aggregation function, window size), and (iii) queries to explain. 
It then applies rules with \ruc, records applied rules and scores, conducts an evaluation and generates visual explanations.

For instance, selecting all \emph{Consult} queries where \ruc outperforms RE-GCN enables targeted investigation. 
This allows researchers to inspect which rule patterns explain correct forecasts and where neural models may miss such dependencies.

\paragraph{Output}
For each query, the tool produces a structured explanation that includes the predicted node and score, the rules that contributed (with their parameters), visualizations of temporal confidence functions, and the frequency/recency of supporting quadruples.
In addition, the tool outputs evaluation scores (MRR, Hits@k) for the quadruples of interest, allowing fine-grained comparison of settings.

\section{{Examples for Learned Recency Functions}}
\begin{figure}
    \centering
    \includegraphics[width=.5\linewidth]{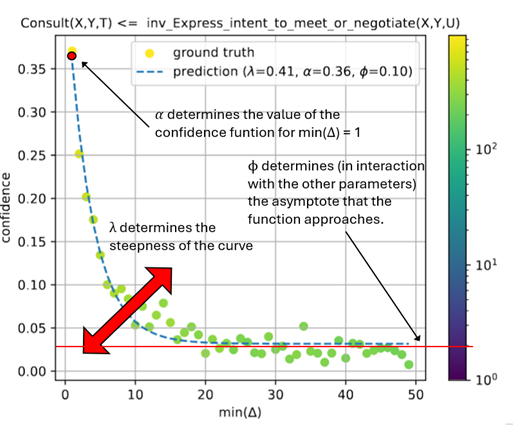}
    \caption{{Examples $E_r$ aggregated per $\min(\Delta)$ (points) and predicted confidence curves (blue lines) $f$ for a rule $r$. Colors indicate the number of samples in $E_r$ per point. The text explains the role of each parameter $\lambda, \alpha, \phi$.}}
    \label{fig:param_explain}
\end{figure}

\begin{figure*}
    \centering
       \begin{subfigure}[t]{0.45\linewidth}
        \includegraphics[width=\linewidth]{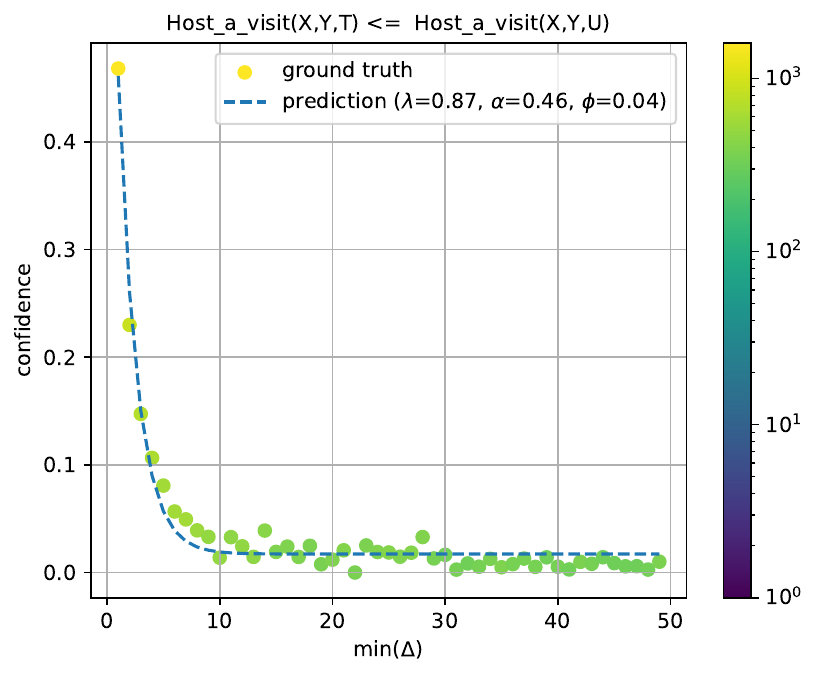}
    \end{subfigure}  
    \hspace{0.1em} 
    \begin{subfigure}[t]{0.45\linewidth}
        \includegraphics[width=\linewidth]{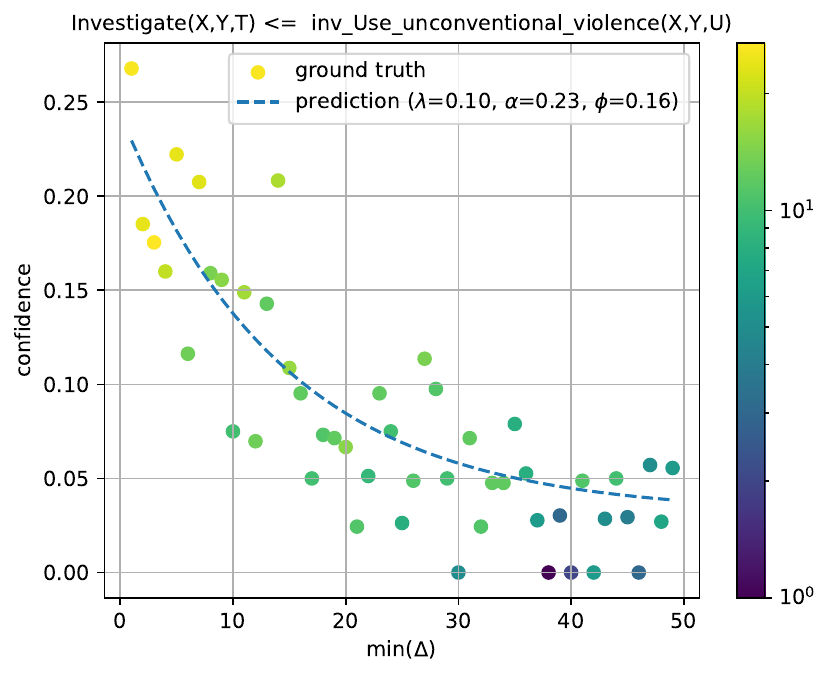}
    \end{subfigure}  
    \hfill
    \begin{subfigure}[t]{0.45\linewidth}
        \includegraphics[width=\linewidth]{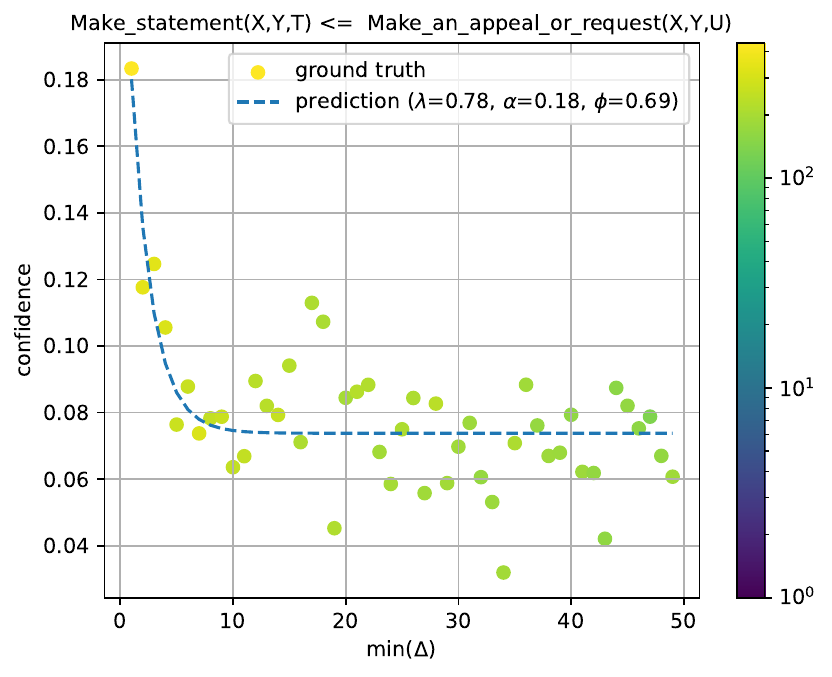}
    \end{subfigure}     
    \caption{
    {Examples $E_r$ aggregated per $\min(\Delta)$ (points) and predicted confidence curves (blue lines) $f$ for three rules. Colors indicate the number of samples in $E_r$ per point.}
    }
    \label{fig:example_recency}
    \vspace{-16pt}
\end{figure*}

{As explained in Section~4.2 in the main paper, 
every parameter in \ruc has a fixed and interpretable role.
Figure~\ref{fig:param_explain} illustrates the role of each parameter $\lambda, \alpha, \phi$ in the recency function $f$ for an individual rule. Figure~\ref{fig:example_recency} provides three concrete examples of rules that have learned different parameters, along with the corresponding sets of examples $E_r$ that were used to learn these curves. Together, the figures highlight that different rules indeed need different parameters to describe their behaviour.}

{For example, the first rule exhibits a steep decay, with its confidence approaching zero as $\min(\Delta)$ increases.
In contrast, the second rule requires a relatively shallow decay. 
The third rule does not converge to zero and is supported by positive examples at higher $\min(\Delta)$. }

{These observations align with our Ablation Study in Section~5.2, Table~2(c) in the main paper, 
which demonstrates that learning rule-specific parameters improves performance. In summary, the study confirms that different rules indeed require different confidence-function shapes, and modeling them individually is beneficial.}

\section{Rule Length Study for TLogic}\label{app:tlogic_rule}
Table~\ref{tab:rule} shows test MRRs of TLogic with body length restricted to~1 (TLogic$_{\texttt{one}}$), TLogic with its original body lengths (TLogic$_{\texttt{orig}}$)\footnote{TLogic$_{\texttt{orig}}$ 
was tested on rules up to length~3 for \texttt{ICEWS14/18} and \texttt{YAGO}, and up to length~2 for \texttt{GDELT/WIKI}. We omit other datasets where TLogic$_{\texttt{orig}}$ was tested only on length-1 rules.}, and CountTRuCoLa.
Multi-hop rules yield noticeable gains on \texttt{ICEWS14/18}, while the remaining datasets even show marginal performance decrease. This indicates that restricting rule bodys to length~1 does not substantially reduce performance on many standard datasets. Consequently, extending CountTRuCoLa with multi-hop rules may yield additional gains mainly on datasets such as \texttt{ICEWS14/18}. However, this would come at the cost of increased runtime and reduced simplicity.

\begin{table}[]
\footnotesize
    \centering
        \caption{Rule Body length study, test MRRs.}
    \resizebox{.7\textwidth}{!}{
    \begin{tabular}{l|r|r|r|r|r }
    \toprule
     Dataset & \texttt{ICEWS14} & \texttt{ICEWS18} & \texttt{YAGO} & \texttt{WIKI} & \texttt{GDELT} \\
     \midrule
     TLogic$_{\texttt{one}}$  & 40.7 & 27.5 & 76.4 & 82.4 & 20.0 \\
     TLogic$_{\texttt{orig}}$  & 42.5 & 29.6 & 76.5 & 82.3 & 19.8   \\
      CountTRuCoLa &  45.0 &32.8 &  90.9 &  82.7 & 23.8\\
      \bottomrule
    \end{tabular}}

    \label{tab:rule}
\vskip -0.2in

\end{table}

\end{document}